\documentclass[a4paper,10pt]{article}
\usepackage{amsmath,amsfonts,amssymb,amsthm}
\usepackage{graphicx,subfigure,booktabs,multirow}
\usepackage{mathrsfs,enumerate,color}
\usepackage{algorithm,algpseudocode}
\usepackage[textwidth=14cm]{geometry}
\usepackage{multirow}
\usepackage{epstopdf}
\usepackage{booktabs}
\usepackage{bm}

\usepackage[colorlinks,linkcolor=red,anchorcolor=blue,urlcolor=red,citecolor=blue]{hyperref}

\graphicspath{{Figures/}}
\newtheorem{theorem}{Theorem}[section]
\newtheorem{lemma}[theorem]{Lemma}
\newtheorem{proposition}[theorem]{Proposition}

\newtheorem{remark}[theorem]{Remark}
\newtheorem{definition}{Definition}[section]

\newcommand{\R}{\mathbb{R}}
\newcommand{\C}{\mathbb{C}}

\newcommand{\PP}{\mathbb{P}}
\newcommand{\E}{\mathbb{E}}
\newcommand{\<}{\langle}
\renewcommand{\>}{\rangle}

\newcommand{\vct}[1]{\boldsymbol{#1}}

\newcommand{\tub}[1]{\mathring{\vct{#1}}}
\newcommand{\tc}[1]{\vec{\vct{#1}}}

\newcommand{\rank}{\operatorname{rank}}

\newcommand{\eijk}{\tc{e}_{ik} \diamond_{\bf \Phi} \ddot{\bm{e}}_k \diamond_{\bf \Phi} \tc{e}^{H}_{jk}}

\newcommand{\PT}{{\cal P}_T}

\newcommand{\PO}{{\cal P}_{\Omega}}

\newcommand{\OpId}{\mathcal{I}}

\title{Tensor Completion by Multi-Rank via
Unitary Transformation\thanks{The research of G.-J. Song
was supported in part by the National Natural Science Foundation of China
under Grant 12171369 and Key NSF of Shandong Province under Grant ZR2020KA008.
The research of M. K. Ng
was supported in part by the HKRGC GRF 12300218, 12300519, 17201020, 17300021, and N-HKU769-21.
The research of X. Zhang was supported in part
by the National Natural Science Foundation of China
under Grant Nos. 12171189 and 11801206.
}}
\author{   Guang-Jing Song\thanks{G.-J. Song is with the School of Mathematics and Information Sciences, Weifang University,
Weifang 261061, China (e-mail: sgjshu@163.com).} \and
  Michael K. Ng\thanks{M. K. Ng is with the Department of Mathematics, The University of
  Hong Kong, Hong Kong (e-mail: mng@maths.hku.hk).}
\and
       Xiongjun Zhang\thanks{The Corresponding Author. X. Zhang is with the School of Mathematics and Statistics
and Hubei Key Laboratory of Mathematical Sciences,
Central China Normal University, Wuhan 430079, China (e-mail: xjzhang@mail.ccnu.edu.cn).}}

\begin{document}

 \maketitle

\noindent \textbf{Abstract:}
One of the key problems in tensor completion is the
number of uniformly random sample entries required for recovery guarantee. The main aim
of this paper is to study $n_1 \times n_2 \times n_3$ third-order tensor completion based on transformed tensor
singular value decomposition, and provide a bound on the number of required sample entries.
Our approach is to make use of the multi-rank of the underlying tensor instead of its
tubal rank in the bound. In numerical experiments on synthetic and imaging data sets,
we demonstrate the effectiveness of our proposed bound for the number of sample entries.
Moreover, our theoretical results are valid to any unitary transformation applied to $n_3$-dimension
under transformed tensor singular value decomposition.

\vspace{2mm} \noindent \textbf{Keywords:} tensor completion,
transformed tensor singular value decomposition, sampling sizes, transformed tensor nuclear norm

\vspace{2mm} \noindent \textbf{AMS Subject Classifications 2010:} 15A69, 15A83, 90C25

\section{Introduction}

The problem of recovering an unknown low-rank tensor from a small
fraction of its entries is known as the tensor completion problem,
and comes up in a wide range of  applications,
e.g., image processing \cite{imbiriba2018low, ng2017adaptive, zhang2021low},
computer vision \cite{liu2013, zhang2019corrected},
and machine learning \cite{romera2013multilinear, signoretto2014learning}.
The goal of low-rank tensor
completion is to recover a tensor with the lowest rank
based on observable entries of a given tensor.
Given a third-order tensor $\mathcal{M}\in\mathbb{C}^{n_1\times n_2\times n_3}$, the low rank tensor completion problem
 can be expressed as follows:
\begin{equation} \label{n1}
\begin{split}
\min_{\mathcal{Z}}& ~ \rank(\mathcal{Z})\\
\text{s.t.}&~ \mathcal{P}_{\Omega}(\mathcal{Z})=\mathcal{P}_{\Omega}(\mathcal{M}),
\end{split}
\end{equation}
where $\textup{rank}(\mathcal{Z})$ denotes the rank of the tensor $\mathcal{Z}$,
$\Omega$ is a subset of $\{1,\ldots, n_1\}\times\{1,\dots,n_2\}\times\{1,\dots,n_3\}$,
 and $\mathcal{P}_\Omega$ is the projection operator
such that the entries in $\Omega$ are given while the remaining entries are missing, i.e.,
\begin{align*}\label{p1}
(\mathcal{P}_\Omega(\mathcal{Z}))_{ijk}= \left\{
\begin{array}{ll}
\mathcal{Z}_{ijk}, & \mbox{if}\ (i,j,k)\in\Omega, \\
0, &\mbox{otherwise}.
\end{array}
\right.
\end{align*}
In particular, if $n_{3}=1$,
the tensor completion problem in (\ref{n1}) reduces to
the well-known matrix completion problem,
which has received a considerable amount of attention in the past decades, see, e.g.,
\cite{candes2010matrix,candes2009exact,candes2010the,gross2010,recht2010guaranteed}
and references therein.
In the matrix completion problem, a given incoherent $n\times n$ matrix could be recovered with high probability
if the uniformly random sample size is of order  $O(r n \log (n)),$
where $r$ is the rank of the given matrix.
This bound has been shown to be optimal, see \cite{candes2010the} for detailed discussions.

The main aim of this paper is to study  the incoherence conditions
of low-rank tensors based on the transformed tensor singular value decomposition (SVD)
and provide a  lower bound on the number of random sample entries required for exact tensor recovery.
Different kinds of tensor ranks setting in model \eqref{n1} lead to different convex relaxation models
and different sample sizes required to exactly recover the original tensor.
When the rank in the model \eqref{n1} is chosen as the Tucker rank  \cite{Tucker66},
Liu et al. \cite{liu2013}
proposed to use the sum of the nuclear norms (SNN) of unfolding matrices
of a tensor to recover a low Tucker rank tensor. Within the SNN framework,
Tomioka et al. \cite{tomioka2011statistical}
proved that a given $d$-order tensor $\mathcal{X}\in\mathbb{R}^{n\times\cdots\times n}$
with Tucker rank $(r,\ldots,r)$ can be exactly recovered with high probability
if the Gaussian measurements size is of order $O(rn^{d-1})$.
Afterwards, Mu et al. \cite{mu2014square} showed that $O(rn^{d-1})$ Gaussian measurements
are necessary for a reliable recovery by the SNN method.
In fact, the degree of freedoms of a tensor
$\mathcal{X}\in \R^{n  \times\cdots\times n}$ with
Tucker rank $(r_{1},\ldots,r_{d})$ is $\prod_{i=1}^dr_i+\sum_{i=1}^{d}(r_{i}n -r_{i}^{2})$,
which is much smaller than $O(rn^{d-1})$.
Recently, Yuan et al. \cite{yuan2016tensor,yuan2017incoherent}
showed that  an  $n\times n\times n$ tensor with Tucker rank $(r,r,r)$ can be exactly recovered with high probability
by  $O((r^{\frac{1}{2}}n^{\frac{3}{2}}+r^2n)\log^2(n))$ entries,
which have a great improvement compared with the
number of sampled sizes required in \cite{mu2014square} when $n$ is relatively large.
Later, based on a gradient descent algorithm designed on some product smooth manifolds,
 Xia et al. \cite{Xia2019On} showed that an  $n\times n\times n$ tensor with multi-linear rank
 $(r,r,r)$ can be  reconstructed with high probability by  $O(r^{\frac{7}{2}}n^{\frac{3}{2}}\log^{\frac{3}{2}}(n)+r^7n\log^6 (n))$  entries.   When the rank in the model \eqref{n1} is chosen as the CANDECOMP/PARAFAC (CP) rank
 \cite{Kolda2009},
Mu et.al \cite{mu2014square} introduced a square deal method
which only uses an individual  nuclear norm of a balanced matrix instead of
using a combination of  all $d$ nuclear norms of unfolding matrices of  the tensor. Moreover,
they showed that $O(r^{\lfloor\frac{d}{2}\rfloor}n^{\lceil \frac{d}{2}\rceil })$ samples
are sufficient to recover a CP rank $r$ tensor with high probability.

Besides, some tensor estimation and recovery problems for the observations
with  Gaussian measurements were proposed
and  studied in the literature
  \cite{ahmed2020tensor, cai2020provable, luo2021low, rauhut2017low, tong2021scaling}.
For example, Ahmed et al. \cite{ahmed2020tensor} proposed
and studied the tensor regression problem by using low-rank and sparse Tucker decomposition,
where a tensor variant of projected gradient descent was proposed
 and the sample complexity of this algorithm
 for a $d$-order tensor $n_1\times \cdots\times n_d$ with Tucker rank $(r_1,\ldots, r_d)$
  is $O(\bar{r}^d+\bar{s}\bar{r}d\log^2(3\bar{n}d))$
  under the restricted isometry property for sub-Gaussian linear maps.
  Here $\bar{r}=\max\{r_1,\ldots,r_d\}$ and $\bar{n}=\max\{n_1,\ldots, n_d\}$
  and $\bar{s}=\max\{s_1,\ldots, s_d\}$,
  where $s_i$ is the upper bound of the number of nonzero entries of each column of the $i$-th factor matrix in the Tucker decomposition.
Moreover,  Cai et al. \cite{cai2020provable} showed that
  the Riemannian gradient algorithm  can reconstruct
  a $d$-order tensor of size $n\times \cdots \times n$ and Tucker rank $(r, \ldots, r)$ with high
probability from only $O(nr^2 + r^{d+1})$ measurements under the tensor restricted isometry property for  Gaussian measurements,
where  one step of iterative hard
thresholding was used for the initialization.

The tubal rank of a third-order tensor was first proposed by Kilmer et al. \cite{kilmer,Kilmer2011},
which is based on tensor-tensor product
(t-product). Within the associated algebraic framework of t-product,
the tensor SVD was proposed and studied similarly to matrix SVD  \cite{Kilmer2011}.
For tensor tubal rank minimization, Zhang et al. \cite{Zhang2017}
proved that the tensor tubal nuclear norm (TNN) can be used as
a convex relaxation of the tensor tubal rank.
 Then they showed that an $n\times n\times n$ tensor with tubal rank $r$
can be exactly recovered by $O(rn^{2}\log(n^2))$ uniformly sampled entries.
However, the TNN is not the convex envelope of the tubal rank of a tensor,
which may lead to more sample entries needed to exactly recover the original tensor.
In Table \ref{table1}, we summarize existing results and our contribution for the $n\times n\times n$ tensor completion problem.
It is interesting  that there are other
factors that affect sample sizes requirement such as sampling methods and incoherence conditions.
For detailed discussions,  the interested readers are referred to
\cite{barak2016noisy,jain2014provable,krishnamurthy2013low,montanari2018spectral}.

\begin{table}[h!]
\small
  \caption{Sampling sizes and sampling methods for  third-order tensor completion.} \label{table1}
 \begin{center}
   \begin{tabular}{|c|c|c|c|}
  \hline
  Rank & Sampling  & Incoherent  and Other &  Requirement  \\
  Assumption& Method &Conditions&Sampling Sizes\\\hline
    CP rank $r$ \cite{mu2014square}& Gaussian& N/A &$O(rn^{2})$  \\\hline
 \multirow{2}{*}{CP rank $r$ \cite{jain2014provable}}&Uniformly & Incoherent condition of &  \multirow{2}{*}{$O(n^{\frac{3}{2}}r^{5}\log^4(n))$}\\
      & Random&symmetric tensor &   \\\hline
    Tucker rank \cite{yuan2017incoherent}&Uniformly& Matrix incoherent condition &  \multirow{2}{*}{$O(rn^{\frac{3}{2}}+r^2n)\log^2(n)$} \\
    $(r,r,r)$& Random &on model-$n$ unfolding &  \\\hline
    Tucker rank \cite{huang2014provable}&  \multirow{2}{*}{Random}  & Matrix incoherent condition &  \multirow{2}{*}{$O(rn^{2}\log^2(n))$}\\
    $(r,r,r)$&  &on model-$n$ unfolding &   \\\hline
     Tucker rank \cite{mu2014square}&  \multirow{2}{*}{Gaussian}  &  \multirow{2}{*}{N/A} &  \multirow{2}{*}{$O(rn^{2})$}  \\
     $(r,r,r)$ & & & \\\hline
     Tucker rank \cite{Xia2019On}&Uniformly& Matrix incoherent condition & \multirow{2}{*}{$O(r^{\frac{7}{2}}n^{\frac{3}{2}}\log^{\frac{3}{2}}(n)+r^7n\log^6 (n))$} \\
     $(r,r,r)$& Random &on model-$n$ unfolding &  \\\hline
      \multirow{2}{*}{Tubal rank $r$ \cite{Zhang2017}}& Uniformly & \multirow{2}{*}{Tensor incoherent condition} &  \multirow{2}{*}{$O(rn^{2}\log(n^2))$}  \\
      & Random  & &  \\\hline
    multi-rank $(r_{1},\ldots,r_{n})$ & Uniformly  & \multirow{2}{*}{Tensor incoherent condition} &  \multirow{2}{*}{$O(\sum_{i=1}^{n} r_{i}n\log(n^2))$}  \\
    (in this paper)& Random  & &   \\
  \hline
\end{tabular}
\end{center}
\end{table}

In this paper, we mainly study the $n_1 \times n_2 \times n_3$
third-order tensor completion problem based on transformed tensor
SVD and transformed tensor nuclear norm (TTNN) \cite{song2019robust}. We show that
such low-rank tensors can be exactly recovered  with high probability when the number of
randomly observed entries is of order $O( \sum_{i=1}^{n_3}r_i \max \{ n_1, n_2 \} \log ( \max \{ n_1, n_2 \} n_3))$,
where $r_i$ is the $i$-th element of the transformed multi-rank of a tensor.

The  rest of this paper is  organized as follows.
In Section \ref{Sect2}, the transformed tensor
SVD related to an arbitrary unitary transformation is reviewed.
In Section \ref{mainresults}, we provide the bound on the
number of sample entries for tensor completion via  any
unitary transformation. In Section \ref{Sect4},
several synthetic data and imaging data sets
are performed to demonstrate
that our theoretical result is valid and the performance of the proposed method
is better than the existing methods in terms of sample sizes requirement.
Some concluding remarks are given in Section \ref{Sect5}.
Finally, the proofs of auxiliary lemmas supporting our main
theorem are provided in the appendix.

\section{Transformed Tensor Singular Value Decomposition}\label{Sect2}
First, some notations used throughout this paper are introduced.
$\mathbb{N}_{\geq 0}^{n}$ denotes the nonnegative $n$-dimensional  integers space.
 We use ${\bf \Phi}$ to denote an arbitrary unitary matrix, i.e.,
${\bf \Phi} {\bf \Phi}^H = {\bf \Phi}^H {\bf \Phi} =  I$,
where ${\bf \Phi}^H$ is the conjugate transpose of ${\bf \Phi}$ and $I$ is the identity matrix whose dimension
should be clear from the context.
Tensors are represented by capital Euler script letters,
e.g., $\mathcal{A}$.
 A tube of a third-order tensor is defined by fixing the first two indices and varying the third \cite{Kilmer2011}.

 Let $\mathcal{A}\in\mathbb{C}^{n_1\times n_2\times n_3}$ be a  third-order tensor.
$\mathcal{A}_{ijk}$ denotes the $(i,j,k)$-th entry of $\mathcal{A}$.
We use $\hat{\mathcal{A}}_{{\bf \Phi}}$ to denote a
third-order tensor obtained as follows:
\begin{equation*}
\textrm{vec}\left(\hat{\mathcal{A}}_{{\bf \Phi}}(i,j,:)\right)={{\bf
\Phi}} \left(\textrm{vec}(\mathcal{A}(i,j,:))\right),
\end{equation*}
where $\textrm{vec}(\cdot)$ is the vectorization operator from
$\mathbb{C}^{1\times1\times n_3}$ to $\mathbb{C}^{n_3}$  and $\mathcal{A}(i,j,:)$ denotes the $(i,j)$-th tube of $\mathcal{A}$.
For simplicity, we denote $\hat{\mathcal{A}}_{{\bf
\Phi}}={\bf \Phi}[\mathcal{A}]$.
In the same fashion, one can also compute $\mathcal{A}$
from $\hat{\mathcal{A}}_{{\bf \Phi}}$,
i.e., ${\cal A} = {\bf\Phi}^H[\hat{\mathcal{A}}_{{\bf \Phi}}]$.
A block diagonal matrix can be derived by the frontal slices of ${\cal A}$ using the ``blockdiag''  operator:
$$
\textrm{blockdiag}(\mathcal{\mathcal{A}}):=
\left(
                                  \begin{array}{cccc}
                                    \mathcal{A}^{(1)} &  &  &  \\
                                     & \mathcal{A}^{(2)}&  & \\
                                     & & \ddots &  \\
                                     &  &  &\mathcal{A}^{(n_{3})}\\
                                  \end{array}
                                  \right),
$$
where $\mathcal{A}^{(i)}$  is the $i$-th frontal
slice of ${\cal A},~i=1,\ldots,n_{3}$.
Conversely,  the block diagonal matrix $\textrm{blockdiag}(\mathcal{\mathcal{A}})$ can be converted
into a tensor via the following ``fold'' operator:
$$
\textrm{fold}(\textrm{blockdiag}(\mathcal{\mathcal{A}})) := \mathcal{\mathcal{A}}.
$$
After introducing the tensor notation and terminology,
we give the basic definitions about the tensor product,
the conjugate transpose of a tensor, the identity tensor and the unitary tensor with respect to the unitary transformation matrix ${\bf \Phi}$, respectively.

\begin{definition}\cite[Definition 1]{song2019robust}
\label{def1} The ${\bf \Phi}$-product of $\mathcal{A} \in \C^{n_1
\times n_2 \times n_3}$ and $\mathcal{B} \in \C^{n_2 \times n_4
\times n_3}$ is a tensor $\mathcal{C} \in \C^{n_1 \times n_4 \times
n_3}$, which is given by
\begin{equation*}
\mathcal{C} =\mathcal{A} \diamond_{\bf \Phi} \mathcal{B}={\bf
\Phi}^{H}\left[ \emph{fold} \left(
 \emph{blockdiag}(\mathcal{\hat{A}}_{\bf \Phi}) \cdot
\emph{blockdiag}(\mathcal{\hat{B}}_{\bf \Phi})\right) \right].
\end{equation*}
\end{definition}

\begin{remark}
Kernfeld et al. \cite{kernfel} defined
the tensor product between two tensors by using frontal slices
products in the transformed domain based on an arbitrary
invertible linear transformation.
In this paper, we mainly focus on the tensor product
based on unitary transformations.
Moreover, the relation between  ${\bf \Phi}$-product
and tensor product by using fast Fourier transform (FFT) \cite{Kilmer2011} is shown in \cite{song2019robust}.
\end{remark}

\begin{definition}\cite[Definition 2]{song2019robust}
\label{def2.5} For any $\mathcal{A}\in
\C^{n_{1}\times n_{2}\times n_{3}}$, its conjugate transpose with respect to ${\bf \Phi}$,
denoted by $\mathcal{A}^{H}\in \C^{n_{2}\times n_{1}\times n_{3}}$,  is defined as
\begin{equation*}
\mathcal{A}^{H} = {\bf \Phi}^{H} \left [ \emph{fold} \left
(\emph{blockdiag}(\mathcal{\hat{A}}_{\bf \Phi})^{H}\right
)\right].
\end{equation*}
\end{definition}

\begin{definition} \cite[Proposition 4.1]{kernfel}\label{idd}
The identity tensor $\mathcal{I}_{\bf \Phi} \in \C^{n \times n
\times n_3}$ (with respect to ${\bf \Phi}$) is defined to be a
tensor such that ${\cal I}_{\bf \Phi} = {\bf \Phi}^H [ {\cal T}]$,
where each frontal
slice of ${\cal T} \in \R^{n \times n \times n_3}$ is the $n \times
n$ identity matrix.
\end{definition}

\begin{definition}  \cite[Definition 5.1]{kernfel}
A tensor $\mathcal{Q} \in \C^{n \times n \times n_3}$ is
unitary with respect to ${\bf \Phi}$-product
if it satisfies
\begin{equation*}
\mathcal{Q}^{H} \diamond_{{\bf \Phi}} \mathcal{Q} = \mathcal{Q}
\diamond_{\bf \Phi} \mathcal{Q}^{H} =\mathcal{I}_{\bf \Phi}.
\label{eq7}
\end{equation*}
\end{definition}

In addition, $\mathcal{A}$ is a diagonal tensor if and only if each frontal
slice $\mathcal{A}^{(i)}$ of $\mathcal{A}$ is a diagonal matrix. By above definitions, the
transformed tensor SVD with
respect to ${\bf \Phi}$ can be given as follows.

\begin{theorem} \cite[Theorem 5.1]{kernfel}\label{them1}
Suppose that $\mathcal{A}\in \C^{n_{1}\times n_{2}\times n_{3}}$.
Then $\mathcal{A}$ can be factorized as
\begin{equation} \label{equ12}
\mathcal{A}= \mathcal{U} \diamond_{\bf \Phi} \mathcal{S}
\diamond_{\bf \Phi} \mathcal{V}^{H},
\end{equation}
where $\mathcal{U}\in \C^{n_{1}\times n_{1}\times n_{3}},$
$\mathcal{V}\in \C^{n_{2}\times n_{2}\times n_{3}}$ are unitary
tensors with respect to ${\bf \Phi}$-product, and $\mathcal{S}\in
\C^{n_{1}\times n_{2}\times n_{3}}$ is a diagonal tensor.
\end{theorem}

Based on the transformed tensor SVD given in Theorem \ref{them1}, the  transformed multi-rank and tubal rank of a tensor can be defined as follows.

\begin{definition}\cite[Definition 6]{song2019robust}
(i) The  transformed multi-rank of  $\mathcal{A} \in
\C^{n_1 \times n_2 \times n_3}$, denoted by
$\rank_{t}(\mathcal{A})$, is a vector $\textup{\bf r} \in \mathbb{N}_{\geq 0}^{n_3}$
with its $i$-th entry being the rank of the $i$-th frontal slice of
$\hat{\mathcal{A}}_{\bf \Phi}$, i.e.,
\begin{equation*}
\rank_{t}(\mathcal{A}_{\bf \Phi})=\textup{\bf r}~~ \textup{with}~~r_i = \rank(\hat{\mathcal{A}}_{\bf \Phi}^{(i)}),~~i=1,\ldots,n_{3}.
\end{equation*}

(ii) The transformed
tubal rank of $\mathcal{A} \in
\C^{n_1 \times n_2 \times n_3}$, denoted by $\rank_{tt}(\mathcal{A})$, is defined as the
number of nonzero singular tubes of $\mathcal{S}$, where
$\mathcal{S}$ comes from the transformed tensor SVD of $\mathcal{A}$,
i.e.,
\begin{equation}
\rank_{tt}(\mathcal{A}) = \#\{i: \mathcal{S}(i, i, :) \neq \vct{0}\}
= \max_{i} r_i. \label{eq9}
\end{equation}
\end{definition}

\begin{remark}
 For computational improvement,
 we will use the skinny transformed tensor SVD throughout this paper unless otherwise stated, which is defined as follows:
The skinny transformed tensor SVD of $\mathcal{A}\in \C^{n_{1}\times n_{2}\times n_{3}}$ with $\emph{rank}_{tt}(\mathcal{A})=r$ is given as $\mathcal{A}= \mathcal{U} \diamond_{\bf \Phi} \mathcal{S}
\diamond_{\bf \Phi} \mathcal{V}^{H},$ where $\mathcal{U}\in \C^{n_{1}\times r\times n_{3}}$ and
$\mathcal{V}\in \C^{n_{2}\times r\times n_{3}}$ satisfying $\mathcal{U}^H\diamond_{\bf \Phi}\mathcal{U}=\mathcal{I}_{\bf \Phi}, \mathcal{V}^H\diamond_{\bf \Phi}\mathcal{V}=\mathcal{I}_{\bf \Phi}$, and $\mathcal{S}\in
\C^{r\times r\times n_{3}}$ is a diagonal tensor. Here $\mathcal{I}_{\bf \Phi}\in\mathbb{C}^{r\times r\times n_3}$ is the identity tensor.
\end{remark}

The inner product  of two
tensors $\mathcal{A},\mathcal{B}\in\mathbb{C}^{n_1\times n_2\times n_3}$
 related to the unitary transformation ${\bf \Phi}$ is defined as
\begin{equation} \label{tprod}
\langle\mathcal{A}, \mathcal{B} \rangle = \sum_{i=1}^{n_{3}}\langle
\mathcal{A}^{(i)}, \mathcal{B}^{(i)}  \rangle =\langle
\overline{\mathcal{A}}_{\bf \Phi}, \overline{\mathcal{B}}_{\bf \Phi}
\rangle,
\end{equation}
where $\langle
\mathcal{A}^{(i)}, \mathcal{B}^{(i)}  \rangle$ is the usual inner product of two matrices
 and $\overline{\mathcal{A}}_{\bf \Phi}=\textrm{blockdiag}(\mathcal{A}_{\bf \Phi})$.
The following fact will be used throughout the paper:
For any tensor $\mathcal{A}\in \C^{n_{1}\times n_{2}\times n_3}$
and $\mathcal{B}\in \C^{n_{2}\times n_{4}\times n_3},$ one can get that
$\mathcal{A}\diamond_{\bf \Phi}\mathcal{B}=\mathcal{C}\Leftrightarrow
\overline{\mathcal{A}}_{\bf \Phi}\cdot\overline{\mathcal{B}}_{\bf \Phi}=\overline{\mathcal{C}}_{\bf \Phi}.$

 The tensor spectral norm of an arbitrary tensor
$\mathcal{A}\in \C^{n_1 \times n_2 \times n_3}$ related to ${\bf
\Phi}$, denoted by $\|\mathcal{A}\|$, can be defined as
$\|\mathcal{A}\| = \|\overline{\mathcal{A}}_{\bf \Phi}\|$ \cite{song2019robust},
i.e., the spectral norm of its block
diagonal matrix $\overline{\mathcal{A}}_{\bf \Phi}$ in the transformed
domain.  Suppose that ${\tt L}$
is a tensor operator,  its operator norm is defined as
$\|{\tt L}\|_{\textup{op}} = \sup_{\|\mathcal{A}\|_F \leq 1} \|{\tt
L}(\mathcal{A})\|_F,$
where the tensor Frobenius norm of $\mathcal{A}$ is defined as $\|\mathcal{A}\|_{F} =
\sqrt{\sum_{i,j,k}|\mathcal{A}_{ijk}|^2}$.
Specifically, if the  operator norm can be
represented as a tensor $\mathcal{L}$ via ${\bf \Phi}$-product with
$\mathcal{A}$, we have $\| {\tt L}\|_{\textup{op}} = \|\mathcal
{L}\|$. The tensor infinity norm and
the tensor $l_{\infty,2}$  and  are  defined as
\[
\begin{split}
&\|\mathcal{A}\|_{\infty} = \max_{i,j,k}|\mathcal{A}_{ijk}| ~ \textup{and} ~ \|\mathcal{A}\|_{\infty,2}=\max\left\{\max_{i}\sqrt{\sum_{b,k}|\mathcal{A}_{ibk}|^2},\max_{j}\sqrt{\sum_{a,k}|\mathcal{A}_{ajk}|^2} \right\}.
\end{split}
\]
 Moreover, the weighted tensor $l_{\infty,w}$
norm with respect to a weighted vector $w=(\alpha_1,\ldots,\alpha_{n_3})^H\in\mathbb{R}^{n_3}$ is defined as
\begin{equation}\label{norm1}
\|\mathcal{A}\|_{\infty, w}=\max\left\{\max_{i}\sqrt{\sum_{b,k}\alpha^2_{k}|\mathcal{A}_{ibk}|^2},\max_{j}\sqrt{\sum_{a,k} \alpha^2_{k}|\mathcal{A}_{ajk}|^2} \right\},
\end{equation}
where $\sum_{k=1}^{n_3}\alpha^2_{k}=1$.
The aim of this paper is to recover a low transformed
multi-rank tensor, which  motivates us to introduce the following definition
of TTNN.

\begin{definition}\cite[Definition 7]{song2019robust}
The transformed tensor nuclear norm of  $\mathcal{A} \in
\C^{n_1 \times n_2 \times n_3}$, denoted by
$\|\mathcal{A}\|_{\textup{TTNN}}$, is the sum of nuclear norms of
all frontal slices of $\hat{\mathcal{A}}_{\bf \Phi}$, i.e.,
$\|\mathcal{A}\|_{\textup{TTNN}} =
 \sum_{i=1}^{n_3}
\|{\hat{\mathcal{A}}}^{(i)}_{\bf \Phi}\|_{\ast}$. \label{def9}
\end{definition}

Recently, Song et al. \cite{song2019robust} showed that the
TTNN of a tensor is the convex envelope of the sum of the
entries of the transformed multi-rank of a tensor, which is stated in the following.
\begin{lemma}\cite[Lemma 1]{song2019robust}\label{lemm1}
For any tensor $\mathcal{A} \in \C^{n_1 \times n_2 \times n_3}$,
let $\rank_{sum}(\mathcal{A})=\sum_{i=1}^{n_3}
\rank(\hat{\mathcal{A}}^{(i)}_{\bf \Phi})$.
Then $\|\mathcal{A}\|_{\textup{TTNN}}$ is
the convex envelope of $\rank_{sum}(\mathcal{A})$ on
the set $\{ \mathcal{A} \ | \ \| \mathcal{A}\| \leq 1\}$.
\end{lemma}

 Lemma \ref{lemm1} shows that the
TTNN is the tightest convex relaxation of the sum of the entries of the transformed
multi-rank of the tensor over a
unit ball of the tensor spectral norm.
That is why the TTNN is effective in
studying the tensor recovery theory with transformed multi-rank minimization.

Next we introduce two kinds of tensor basis which will be exploited to
derive our main result.

\begin{definition} \label{defn}
(i) The column basis, denoted as
$\tc{e}_{ik}$, is a tensor of size $n_1 \times 1 \times n_3$ with the
$(i,1,k)$-th element equaling to $1$ and the others  equaling to 0.

(ii)  Denote $\tub{e}_{k}$ as a tensor of
size $1 \times 1 \times n_3$ with the $(1,1,k)$-th element equaling to $1$ and
 the remaining elements equaling to 0,   $({\bf \Phi}[\tub{e}_{k}])_{j}$
  as  the $(1,1,j)$-th  element of ${\bf \Phi}[\tub{e}_{k}]$, $j=1,\ldots,n_{3}$.

(iii) The transformed tube
basis,  denoted as $\ddot{\textbf{e}}_k$,
is a tensor of
size $1 \times 1 \times n_3$ with the $(1,1,j)$-th element of $\ddot{\textbf{e}}_k$
equaling to $(({\bf \Phi}[\tub{e}_{k}])_{j})^{-1},$ if $({\bf \Phi}[\tub{e}_{k}])_{j}\neq0,$ and $0$, otherwise, $j=1,\ldots, n_3$.
\end{definition}

\begin{remark}
The transformed tube basis $\ddot{\textbf{e}}_k$ is determined by
the unitary transformation matrix and the original tube $\tub{e}_{k}$,
whose detailed formulation
can be obtained for a given unitary transformation ${\bf \Phi}$.
\end{remark}

\section{Main Results} \label{mainresults}

In this section, we consider the third-order tensor completion problem,
which aims to recover a low transformed multi-rank tensor  under some limited observations. Mathematically, the problem can be described as follows:
\begin{equation*} \label{nj1}
\begin{split}
\min_{\mathcal{Z}}& ~  \sum_{i=1}^{n_3}\rank(\hat{\mathcal{Z}}_{\bf \Phi}^{(i)})  \\
\text{s.t.}&~ \mathcal{P}_{\Omega}(\mathcal{Z})=\mathcal{P}_{\Omega}(\mathcal{M}),
\end{split}
\end{equation*}
where $\rank(\hat{\mathcal{Z}}_{\bf \Phi}^{(i)})$ is
the rank of $\hat{\mathcal{Z}}_{\bf \Phi}^{(i)}$, i.e.,
the $i$-th element of the transformed multi-rank of $\mathcal{Z}, i=1,\ldots, n_3$,
$\Omega$ and $\mathcal{P}_{\Omega}(\mathcal{Z})$ are defined in model \eqref{n1}.
Note that the rank minimization problem is NP-hard
and  the TTNN is the convex envelope of the
sum of the entries of the transformed multi-rank of a tensor \cite[Lemma 1]{song2019robust}.
Therefore, we propose to utilize the TTNN as a convex
relaxation of the sum of the entries of the  transformed multi-rank of a tensor.
More precisely speaking, the convex
relaxation model is given by
\begin{equation}\label{ObjFT}
\begin{split}
\min_{\mathcal{Z}} & \
\|\mathcal{Z}\|_{\textup{TTNN}} \\
 \textup{s.t.} & \
\mathcal{P}_\Omega(\mathcal{Z}) = \mathcal{P}_\Omega(\mathcal{M}).
\end{split}
\end{equation}
In the following, we need to introduce the tensor incoherence conditions between the underlying tensor $\mathcal{Z}$ and the column basis given in Definition \ref{defn}.
\begin{definition} Let $\mathcal{Z} \in \C^{n_{1}\times n_{2}\times n_{3}}$ with $\rank_{t}(\mathcal{Z})= \textup{\bf r}$, where $\textup{\bf r}=(r_1,\ldots, r_{n_3})$.
Assume  its skinny
transformed tensor SVD is $\mathcal{Z} = \mathcal{U} \diamond_{\bf \Phi} \mathcal{S} \diamond_{\bf \Phi}
\mathcal{V}^{H}$. Then $\mathcal{Z}$ is said to satisfy the
tensor incoherence conditions,
if there exists a  parameter $\mu\geq 1$, such that
\begin{align}
\max_{i=1, \ldots, n_1}\max_{k=1, \ldots, n_3} \|\mathcal{U}^{H} \diamond_{\bf \Phi} \tc{e}_{ik}\|_F \leq \sqrt{\frac{\mu \sum_{i=1}^{n_3} {r_{i}} }{n_1n_{3}}}, \label{eq16}\\
\max_{j=1, \ldots, n_2} \max_{k=1, \ldots, n_3}\|\mathcal{V}^{H} \diamond_{\bf \Phi} \tc{e}_{jk}\|_F\leq
\sqrt{\frac{\mu \sum_{i=1}^{n_3}{r_{i}}}{n_2n_{3}}}, \label{eq17}
\end{align}\label{def13}
where  $\tc{e}_{ik}\in\mathbb{R}^{n_1\times 1\times n_3}$ and $\tc{e}_{jk}\in\mathbb{R}^{n_2\times 1\times n_3}$ are the column
basis.
\end{definition}

 Denote by $T$ the linear space of tensors
\begin{equation}
T = \big\{\mathcal{U}\diamond_{\bf \Phi}\mathcal{Y}^{H} + \mathcal{W}\diamond_{\bf \Phi}
\mathcal{V}^{H} ~|~ \mathcal{Y} \in \C^{n_2 \times r \times
n_3},~\mathcal{W} \in \C^{n_1 \times r \times n_3} \big\},
\label{e1}
\end{equation}
and by $T^{\perp}$ its orthogonal complement,
where
 $\mathcal{U}\in \C^{n_{1}\times r\times n_{3}}$ and
$\mathcal{V}\in \C^{n_{2}\times r\times n_{3}}$ are column unitary tensors,
respectively, i.e., $\mathcal{U}^H\diamond_{\bf \Phi}\mathcal{U}=\mathcal{I}_{\bf \Phi}, \mathcal{V}^H\diamond_{\bf \Phi}\mathcal{V}=\mathcal{I}_{\bf \Phi}$.
In the light of \cite[Proposition B.1]{zhang2019corrected}, for any $\mathcal{Z}\in\mathbb{C}^{n_1\times n_2\times n_3}$, the
orthogonal projections onto $T$ and its complementary are given as follows:
\[
\begin{split}
&\mathcal{P}_{T}(\mathcal{Z}) = \mathcal{U}\diamond_{\bf \Phi} \mathcal{U}^{H}\diamond_{\bf \Phi}
\mathcal{Z} +\mathcal{Z}\diamond_{\bf \Phi} \mathcal{V}\diamond_{\bf \Phi} \mathcal{V}^{H} -
\mathcal{U}\diamond_{\bf \Phi} \mathcal{U}^{H}\diamond_{\bf \Phi} \mathcal{Z}\diamond_{\bf \Phi}
\mathcal{V}\diamond_{\bf \Phi} \mathcal{V}^{H}, \\
&\mathcal{P}_{T^{\perp}}(\mathcal{Z}) =  (\mathcal{I}_{\bf \Phi} - \mathcal{U}\diamond_{\bf \Phi}
\mathcal{U}^{H})\diamond_{\bf \Phi} \mathcal{Z}\diamond_{\bf \Phi} (\mathcal{I}_{\bf \Phi} - \mathcal{V}\diamond_{\bf \Phi}
\mathcal{V}^{H}).
\end{split}
\]

Denote $n_{(1)}=\max\{n_{1},n_{2}\},n_{(2)}=\min\{n_{1},n_{2}\}.$
We can improve the low bound on the  number of
sampling sizes  for tensor completion
 by using transformed multi-rank instead of using tubal rank, which is stated in the following theorem.

\begin{theorem}\label{Theorem1}
Suppose that $\mathcal{Z} \in \C^{n_1 \times n_2 \times n_3}$  with $\rank_{t}(\mathcal{Z})= \textup{\bf r}$ and its  skinny
transformed tensor SVD is $\mathcal{Z} = \mathcal{U} \diamond_{\bf \Phi} \mathcal{S} \diamond_{\bf \Phi}
\mathcal{V}^{H}$,  where $\textup{\bf r}=(r_1,\ldots, r_{n_3})$, $\mathcal{U}\in \C^{n_{1}\times r\times n_{3}},$ $\mathcal{S}\in \C^{r\times r\times n_{3}}$ and
$\mathcal{V}\in \C^{n_{2}\times r\times n_{3}}$ with $\rank_{tt}(\mathcal{Z})=r$.
Suppose that $\mathcal{Z}$
satisfies the tensor incoherence conditions \eqref{eq16}-\eqref{eq17} and the observation set $\Omega$ with $|\Omega|=m$ is
uniformly distributed among all sets of cardinality, then there
exist universal constants $c_0, c_1, c_2 > 0$ such that if
\begin{align}\label{new1}
m\geq c_0 \mu\sum_{i=1}^{n_{3}}{r_{i}}n_{(1)}\log(n_{(1)}n_3),
\end{align}
$\mathcal{Z}$ is the unique minimizer to $\eqref{ObjFT}$ with
probability at least $1- c_1(n_{(1)}n_3)^{-c_2} $.
\end{theorem}

 Next we compare the number of sample sizes requirement for exact recovery in \cite{Zhang2017}.
Note that the tensor incoherence conditions in \cite{Zhang2017} are given by
\begin{align*}
&\max_{i=1, \dots, n_1} \|\mathcal{U}^{H} \diamond_{\bf \Phi} \tc{e}_{i1}\|_F \leq \sqrt{\frac{\mu_{old} r}{n_1}},\\
& \max_{j=1, \dots, n_2} \|\mathcal{V}^{H} \diamond_{\bf \Phi}
\tc{e}_{j1}\|_F\leq \sqrt{\frac{\mu_{old} r}{n_2}},
\end{align*}
where $r$ is the tubal rank of the underlying tensor,
$\mu_{old}>0$ is a parameter, ${\bf \Phi}$ is FFT,
 $\tc{e}_{i1}$ and $\tc{e}_{j1}$ are defined in Definition \ref{defn}.
When the number of samples is larger than or equal to  $\widetilde{c}rn_{(1)}n_3\log(n_{(1)}n_3)$, the underlying tensor
can be recovered exactly \cite{Zhang2017}, where $\widetilde{c}>0$ is a given constant.
Neglecting the constants, we know that the bound of sample sizes
requirement in Theorem \ref{Theorem1} is smaller than that of \cite{Zhang2017}
since $\sum_{i=1}^{n_3}r_i$ is smaller than $rn_3$ in general.
Especially,
when the the vector of the transformed multi-rank ${\bf r}$ is sparse,
the number of sample sizes requirement  given in \eqref{new1} is much smaller than that in \cite{Zhang2017} for exact recovery.
 Moreover, the exact recovery theory in Theorem \ref{Theorem1}
not only holds for FFT but
also for any unitary transformation, which is very
meaningful in practical applications.
Based on Theorem \ref{Theorem1}, for a given data tensor,
we can choose a  suitable unitary transformation such that
 the sum of elements of
the transformed multi-rank of the underlying tensor is small \cite{song2019robust, zhang2021low},
which can guarantee to derive better results than that by using FFT directly.

To facilitate our proof of the main theorem, we will consider
the  independent and identically distributed \textit{(i.i.d.)
Bernoulli-Rademacher model}. More precisely, we assume $\Omega =
\{(i, j, k) \ | \ \delta_{ijk} = 1\}$, where $\delta_{ijk}$ are
i.i.d. Bernoulli variables taking value one with probability $\rho = \frac{m}{n_1n_2n_3}$
and zero with probability $1-\rho$. Such a Bernoulli sampling is
denoted by $\Omega \sim \textup{Ber}(\rho)$ for short. As a proxy
for uniform sampling, the probability of failure under Bernoulli
sampling  closely approximates the
probability of failure under uniform sampling.

Recall the definitions of tensor Frobenius norm and the tensor
incoherence conditions given in (\ref{eq16})-(\ref{eq17}), we can get the following result easily.

\begin{proposition} \label{pro1}
Let $\mathcal{Z} \in \C^{n_1 \times n_2 \times n_3}$ be an arbitrary
tensor  with $\rank_{t}(\mathcal{Z})= \textup{\bf r}$, and $T$ be given as (\ref{e1}). Suppose that the
tensor incoherence conditions (\ref{eq16})-(\ref{eq17}) are
satisfied, then
$$\|\mathcal{P}_{T}( \tc{e}_{ik}\diamond_{\bf \Phi}
\ddot{\textbf{e}}_k \diamond_{\bf \Phi} \tc{e}_{jk}^H)\|^{2}_{\textrm{F}}\leq \frac{2\mu\sum_{i=1}^{n_3} {r_{i}}}{n_{(2)}n_{3}}.$$
\end{proposition}

Proposition \ref{pro1} plays an important role in  the proofs of Lemmas
\ref{le1}, \ref{lemma2} and \ref{le3}, which can be found in the Appendix.

\begin{lemma} \label{le1}
Suppose that $\Omega \sim \textup{Ber}(\rho),$   where $\Omega$ with $|\Omega|=m$ is a set of indices sampled
independently and uniformly without replacement, $\rho=\frac{m}{n_{1}n_{2}n_{3}}$ and $T$ is given as
(\ref{e1}). Then with high probability,
\begin{align} \label{11}
\|\rho^{-1}\mathcal{P}_{T}\mathcal{P}_{\Omega}\mathcal{P}_{T}-\mathcal{P}_{T}\|_\textup{op}\leq
\epsilon,
 \end{align}
  provided that $m\geq C_0\epsilon^{-2} \mu \sum_{i=1}^{n_3}{r_{i}}n_{(1)}
\log(n_{(1)}n_3) $ for
some numerical constant $C_0
> 0$.
\end{lemma}

\begin{lemma}\label{lemma2}
Suppose that $\mathcal{Z} \in \C^{n_1 \times n_2 \times n_3}$ is a
tensor with $\rank_{t}(\mathcal{Z})= \textup{\bf r}$,  $\Omega, \rho$ and $m$ are defined in Lemma \ref{le1}.
Then for all $c>1$ and $C_{0}>0$,
\begin{equation*}
\|(\rho^{-1}\mathcal{P}_{\Omega}-\OpId)\mathcal{Z}\| \leq
c\left(\frac{\log (n_{(1)}n_{3})}{\rho}\|\mathcal{Z}\|_{\infty}+\sqrt{\frac{\log (n_{(1)}n_{3})}{\rho}}\|\mathcal{Z}\|_{\infty,w}\right)
\label{eq25}
\end{equation*}
holds with high probability
provided that
$m\geq C_0 \epsilon^{-2} \mu \sum_{i=1}^{n_3}{r_{i}}n_{(1)}\log(n_{(1)}n_3),$
 where $\OpId$ denotes the identity operator.
\end{lemma}
\begin{lemma}\label{le3}
Suppose that $\mathcal{Z} \in \C^{n_1 \times n_2 \times n_3}$ is a
tensor with $\rank_{t}(\mathcal{Z})= \textup{\bf r}$, $\Omega, \rho$ and $m$ are defined in Lemma \ref{le1}. Then for some sufficiently large $C_{0}$,
\begin{equation}
\|(\rho^{-1}\mathcal{P}_{T}\mathcal{P}_{\Omega}-\mathcal{P}_{T})\mathcal{Z}\|_{\infty,w}\leq\frac{1}{2}\sqrt{\frac{n_{(1)}n_{3}}{\mu\sum_{i=1}^{n_{3}}r_{i}}}\|\mathcal{Z}\|_{\infty}
+\frac{1}{2}\|\mathcal{Z}\|_{\infty,w}
\end{equation}
holds with high probability provided that  $m\geq C_0 \epsilon^{-2} \mu \sum_{i=1}^{n_3}{r_{i}}n_{(1)}\log(n_{(1)}n_3).$
\end{lemma}

The proofs of the three lemmas are left to the appendix.
Lemma \ref{lemma2}  establishes an upper bound of the tensor
spectral norm of $(\rho^{-1}\mathcal{P}_{\Omega}-\OpId)\mathcal{Z}$
in terms of $\|\mathcal{Z}\|_{\infty}$ and $\|\mathcal{Z}\|_{\infty,w},$
which is tighter than  the existing result given by $\|\mathcal{Z}\|_{\infty}$
and $\|\mathcal{Z}\|_{\infty,2}$. The weights in $\|\mathcal{Z}\|_{\infty,w}$
are determined by the unitary transformation,  which
guarantee the upper bound of $\|(\rho^{-1}\mathcal{P}_{\Omega}-\OpId)\mathcal{Z}\|$ to be smaller.
Lemma \ref{le3} shows that  the $l_{{\infty,w}}$ norm
of $(\rho^{-1}\mathcal{P}_{T}\mathcal{P}_{\Omega}-\mathcal{P}_{T})\mathcal{Z}$ can be
 dominated by $\|\mathcal{Z}\|_{\infty,w}$ and $\|\mathcal{Z}\|_{\infty}$.
Moreover, Lemmas C1 and C2 in  \cite{Zhang2017} can be seen as special cases of Lemmas \ref{lemma2} and \ref{le3}, respectively, if the transformation is chosen as FFT.
\begin{lemma}\cite[Lemma 4.1]{Zhang2017}\label{4.7}
Suppose that $\|\rho^{-1}\PT\PO\PT - \PT\|_\textup{op}\leq \frac{1}{2}.$ Then for any $\mathcal{Z}\in\mathbb{C}^{n_1\times n_2\times n_3 }$ such that $\mathcal{P}_{\Omega}(\mathcal{Z})=0,$  the following inequality
\begin{equation*}
\frac{1}{2}\|\mathcal{P}_{T^{\bot}}(\mathcal{Z})\|_{\emph{TTNN}}>\frac{1}{4n_{(1)}n_{3}}\|\mathcal{P}_{T}(\mathcal{Z})\|_{F}
\end{equation*}
holds with high probability.
\end{lemma}
With the tools in hand we can list the proof of Theorem \ref{Theorem1} in  detail.

\vspace{2mm}
\noindent
{\bf Proof of Theorem \ref{Theorem1}}.
 The high level road map of the proof is a standard one just as shown in \cite{candes2011robust}: by convex analysis, to show $\mathcal{Z}$ is the unique optimal solution to the problem $\eqref{ObjFT}$, it is sufficient to find a dual certificate $\mathcal{Y}$ satisfying several subgradient type conditions.  In our case, we need to find a tensor $\mathcal{Y}=\mathcal{P}_{\Omega}(\mathcal{Y})$
such that
\begin{align}
& \|\mathcal{P}_{T}(\mathcal{Y})-\mathcal{U}\diamond_{\bf \Phi}\mathcal{V}^{H}\|_{F}\leq \frac{1}{4n_{(1)}n_{3}^{2}}, \label{cd1}\\
& \|\mathcal{P}_{T^{\bot}}(\mathcal{Y})\|\leq\frac{1}{2}.\label{cd2}
\end{align}
It follows  from \eqref{cd2} that we need to estimate the
tensor spectral norm $\|\mathcal{P}_{T^{\bot}}(\mathcal{Y})\|$.
Similar to matrix cases \cite{candes2011robust}, we can use the tensor infinite norm to establish the upper bound of $\|\mathcal{P}_{T^{\bot}}(\mathcal{Y})\|$. However, if $\|\mathcal{Y}\|_{\infty}$ is applied,  then it will ultimately  link to
$\|\mathcal{U}\diamond_{\bf \Phi}\mathcal{V}^H\|_{\infty}$
and lead to the joint incoherence condition.
In order to avoid applying the joint incoherence condition,
$\|\mathcal{Y}\|_{\infty,2}$ is used in \cite{chen2015incoherence} and \cite{Zhang2017}
to compute the upper bound of $\|\mathcal{Y}\|$ for the matrix and tensor cases, respectively.
 It follows from  \cite{chen2015incoherence} and \cite{Zhang2017}
that a lower upper bound can be derived by  using $\|\mathcal{Y}\|_{\infty,2}$.
However, the upper bound derived by $\|\mathcal{Y}\|_{\infty,2}$
can be relaxed further for an arbitrary unitary transformation.
Here, we derive a new upper bound of $\|\mathcal{P}_{T^{\bot}}(\mathcal{Y})\|$
by using the $l_{\infty,w}$ norm $\|\mathcal{Y}\|_{\infty,w}$
as defined in \eqref{norm1}. Note that $\|\mathcal{Y}\|_{\infty,w}$
is not larger than $\|\mathcal{Y}\|_{\infty,2}$
for any tensor $\mathcal{Y},$ which leads to a tighter upper bound of  $\|\mathcal{Y}\|$.

We now return to the proof Theorem \ref{Theorem1} in detail.
We apply the Golfing Scheme method introduced by Gross
\cite{gross2010} and modified by Cand\`{e}s et al.
\cite{candes2011robust} to construct a dual tensor $\mathcal{Y}$
supported by $\Omega^c$ iteratively. Similar to the proof of
\cite[Theorem 3.1]{Zhang2017}, we consider the set $\Omega^c\sim
\textup{Ber}(1-\rho)$ as a union of sets of support $\Omega_j$, i.e.,
$\Omega^c = \bigcup_{j=1}^p \Omega_j$, where $\Omega_j \sim
\textup{Ber}(q),$
which implies $q \geq C_0\rho/\log(n_{(1)}n_3)$.
Hence we have
$ \rho =  (1 - q)^{p},$
 where $p = \lfloor 5\log(n_{(1)}n_3) + 1\rfloor$. Denote
\begin{equation}\label{s1}
\mathcal{Y}= \sum_{j=1}^p
\frac{1}{q}\mathcal{P}_{\Omega_j}(\mathcal{Z}_{j-1}), ~\text{with}~\mathcal{Z}_{j} = \Big(\mathcal{P}_{T} - \frac{1}{q}\mathcal{P}_{T}
\mathcal{P}_{\Omega_j}\mathcal{P}_{T}\Big)\mathcal{Z}_{j-1},~ \mathcal{Z}_0 =
\mathcal{P}_{T}(\mathcal{U} \diamond_{\bf \Phi} \mathcal{V}^{H} ).
\end{equation}

In the following we will show that $\mathcal{Y}$ defined in \eqref{s1} satisfies the conditions \eqref{cd1} and \eqref{cd2}.

For \eqref{cd1}. Set $\mathcal{D}_{k}:=\mathcal{U}\diamond_{\bf \Phi}\mathcal{V}^{H}-\mathcal{P}_{T}(\mathcal{Z}_{k})$
for $k=0,\ldots,p$. By the definition
of $\mathcal{Z}_{k}$, we have $\mathcal{D}_{0}=\mathcal{U}\diamond_{\bf \Phi}\mathcal{V}^{H}$ and
\begin{equation}\label{eqn1}
\mathcal{D}_{k}=(\mathcal{P}_{T}-\mathcal{P}_{T}\mathcal{P}_{\Omega}\mathcal{P}_{T})\mathcal{D}_{k-1}, \ k=1,\ldots, p.
\end{equation}
 Note that
$\Omega_{k}$ is independent of $\mathcal{D}_{k-1}$ and $q\geq c_{0}\mu\sum r_{i}\log(n_{(1)}n_{3})/(n_{(1)}n_{3}).$
For each $k,$  replacing $\Omega$ by $\Omega_{k},$ then by Lemma \ref{le1}, we have
\begin{equation*}
\|\mathcal{D}_{k}\|_{{F}}\leq \|\mathcal{P}_{T}-\mathcal{P}_{T}\mathcal{P}_{\Omega_{k}}\mathcal{P}_{T}\|\|\mathcal{D}_{k-1}\|_{{F}}\leq \frac{1}{2}\|\mathcal{D}_{k-1}\|_{F}.
\end{equation*}
As a consequence, one can obtain that
\begin{align*}
\|\mathcal{P}_{T}(\mathcal{Y})-\mathcal{U}\diamond_{\bf \Phi} \mathcal{V}^{H}\|_{{F}}=\|\mathcal{D}_{p}\|_{{F}}\leq \left(\frac{1}{2}\right) ^{p}\|\mathcal{U}\diamond_{\bf \Phi} \mathcal{V}^{H}\|_{{F}}\leq\frac{1}{4(n_{(1)}n_{3})^{2}}\sqrt{r}
\leq \frac{1}{4n_{(1)}n_{3}^2}.
\end{align*}

For \eqref{cd2}. Note that $\mathcal{Y}=\sum_{k=1}^{p}\mathcal{P}_{\Omega_{k}}\mathcal{P}_{T}(\mathcal{D}_{k-1}),$ thus
\begin{align}\label{17py}
\|\mathcal{P}_{T^{\bot}}(\mathcal{Y})\|\leq \sum_{k=1}^{p}\|\mathcal{P}_{T^{\bot}}(\mathcal{P}_{\Omega_{k}}\mathcal{P}_{T}-\mathcal{P}_{T})(\mathcal{D}_{k-1})\|
\leq \sum_{k=1}^{p}\|(\mathcal{P}_{\Omega_{k}}-\mathcal{I})\mathcal{P}_{T}(\mathcal{D}_{k-1})\|.
\end{align}
Applying Lemma \ref{lemma2} with $\Omega$ replaced by $\Omega_{k}$ to each summand of (\ref{17py})  yields
\begin{align}
\|\mathcal{P}_{T^{\bot}}(\mathcal{Y})\|&\leq c \sum_{k=1}^{p}\Biggl(\frac{\log (n_{(1)}n_{3})}{q}\|\mathcal{D}_{k-1}\|_{\infty}+\sqrt{\frac{\log (n_{(1)}n_{3})}{q}}\|\mathcal{D}_{k-1}\|_{\infty,w}\Biggl)\nonumber\\
&\leq \frac{c}{\sqrt{c_{0}}} \sum_{k=1}^{p}\Biggl(\frac{ n_{(1)}n_{3}}{\mu\sum r_{i}}\|\mathcal{D}_{k-1}\|_{\infty}+\sqrt{\frac{ n_{(1)}n_{3}}{\mu\sum r_{i}}}\|\mathcal{D}_{k-1}\|_{\infty,w}\Biggl).\label{eq4}
\end{align}
Using \eqref{eqn1}, and applying Lemma \ref{le1} with $\Omega$ replaced by $\Omega_{k},$ we can get
\begin{align*}
\|\mathcal{D}_{k-1}\|_{\infty}=\|(\mathcal{P}_{T^{\bot}}(\mathcal{P}_{\Omega_{k-1}}\mathcal{P}_{T}-\mathcal{P}_{T})\cdots \mathcal{P}_{T^{\bot}}(\mathcal{P}_{\Omega_{1}}\mathcal{P}_{T}-\mathcal{P}_{T}))\mathcal{D}_{0}\|_{\infty}\leq
\frac{1}{2^{k}}\|\mathcal{U}\diamond_{\bf \Phi} \mathcal{V}^{H}\|_{\infty}.
\end{align*}
It follows from Lemma \ref{le3} that
\begin{align*}
\|\mathcal{D}_{k-1}\|_{\infty,w}=\|(\mathcal{P}_{T}-\mathcal{P}_{T}\mathcal{P}_{\Omega_{k-1}}\mathcal{P}_{T})\mathcal{D}_{k-2}\|_{\infty,w}
\leq\frac{1}{2}\sqrt{\frac{n_{(1)}n_{3}}{\mu\sum r_{i}}}\|\mathcal{D}_{k-2}\|_{\infty}+\frac{1}{2}\|D_{k-2}\|_{\infty,w}
\end{align*}
holds with high probability. Moreover, it follows from \eqref{eqn1} that
\begin{align*}
\|\mathcal{D}_{k-1}\|_{\infty,w}\leq\frac{ k}{2^{k-1}}\sqrt{\frac{n_{(1)}n_{3}}{\mu\sum r_{i}}}\|\mathcal{U}\diamond_{\bf \Phi} \mathcal{V}^{H}\|_{\infty}+\frac{1}{2^{k-1}}\|\mathcal{U}\diamond_{\bf \Phi} \mathcal{V}^{H}\|_{\infty,w}.
\end{align*}
Taking them back to \eqref{eq4} yields
\begin{align}
&~\|\mathcal{P}_{T^{\bot}}(\mathcal{Y})\|\nonumber\\
\leq &~\frac{c}{\sqrt{c_{0}}}\frac{n_{(1)}n_{3}}{\mu\sum r_{i}}\|\mathcal{U}\diamond_{\bf \Phi} \mathcal{V}^{H}\|_{\infty}\sum_{k=1}^{p}(k+1)\left(\frac{1}{2}\right)^{k-1} \\
&~~ +\frac{c}{\sqrt{c_{0}}}\sqrt{\frac{n_{(1)}n_{3}}{\mu\sum r_{i}}}\|\mathcal{U}\diamond_{\bf \Phi} \mathcal{V}^{H}\|_{\infty,w}\sum_{k=1}^{p}\left(\frac{1}{2}\right)^{k-1}\nonumber\\
\leq &~\frac{6c}{\sqrt{c_{0}}}\frac{n_{(1)}n_{3}}{\mu\sum r_{i}}\|\mathcal{U}\diamond_{\bf \Phi} \mathcal{V}^{H}\|_{\infty}+ \frac{2c}{\sqrt{c_{0}}}\sqrt{\frac{n_{(1)}n_{3}}{\mu\sum r_{i}}}\|\mathcal{U}\diamond_{\bf \Phi}\mathcal{V}^{H}\|_{\infty,w}.\label{e7}
\end{align}
By the incoherence conditions given in \eqref{eq16}-\eqref{eq17}, we can get
\begin{align}
&\|\mathcal{U}\diamond_{\bf \Phi} \mathcal{V}^{H}\|_{\infty}\leq \max_{i,j,k}\|\mathcal{U}\diamond_{\bf \Phi}\tc{e}_{ik}\|_{F}\|\mathcal{V}^{H}\diamond_{\bf \Phi}\tc{e}_{ik}\|_{{F}}\leq \frac{\mu\sum r_{i}}{n_{(1)}n_{3}},\label{e10}\\
&\|\mathcal{U}\diamond_{\bf \Phi} \mathcal{V}^{H}\|_{\infty,w}\leq \max\left\{\max_{i,k}\| \mathcal{U}\diamond_{\bf \Phi} \mathcal{V}^{H}\diamond_{\bf \Phi}\tc{e}_{ik}\|_{{F}},\max_{j,k}\| \tc{e}_{jk}^{H}\diamond_{\bf \Phi}\mathcal{U}\diamond_{\bf \Phi} \mathcal{V}^{H}\|_{{F}}\right\}\leq \sqrt{\frac{\mu\sum r_{i}}{n_{(1)}n_{3}}}.\label{e11}
\end{align}
 Plugging \eqref{e10} and \eqref{e11} into \eqref{e7}, we obtain that
\begin{align*}
\|\mathcal{P}_{T^{\bot}}(\mathcal{Y})\|\leq\frac{6c}{\sqrt{c_{0}}}+\frac{2c}{\sqrt{c_{0}}}\leq \frac{1}{2}
\end{align*}
provided $c_{0}$ is sufficiently large.

 Moreover, for any tensor $\mathcal{W}\in \{\mathcal{W} \in \C^{n_{1}\times
n_{1}\times n_{3}}|\mathcal{P}_{\Omega}(\mathcal{W})=0\},$ denote the skinny
transformed tensor SVD of $\mathcal{P}_{T^\bot}(\mathcal{W})$ by
$$
 \mathcal{P}_{T^\bot}(\mathcal{W}) =
\mathcal{U}_\perp \diamond_{\bf \Phi}\mathcal{S}_\perp \diamond_{\bf \Phi}
\mathcal{V}^{H}_\perp.
$$
Since
$\mathcal{\overline{U}}_{\bf \Phi}^{H}\cdot(\mathcal{\overline{U}}_{\perp})_{\bf \Phi}=\mathbf{0}$ and
$\mathcal{\overline{V}}_{\bf \Phi}^{H}\cdot(\mathcal{\overline{V}}_{\perp})_{\bf \Phi}=\mathbf{0},$ we
have
$$
\|\mathcal{U} \diamond_{\bf \Phi} \mathcal{V}^{H} + \mathcal{U}_\perp
\diamond_{\bf \Phi} \mathcal{V}^{H}_\perp\| = \|\mathcal{\overline{U}}_{\bf \Phi}\cdot
\mathcal{\overline{V}}_{\bf \Phi}^{H} + (\mathcal{\overline{U}}_\perp)_{\bf \Phi}\cdot
((\mathcal{\overline{V}}_\perp)_{\bf \Phi})^{H}\|=1.
$$
Thus, we get that
\begin{align}
 \|\mathcal{Z} + \mathcal{W}\|_{\textup{TTNN}}
 \geq & ~ \langle \mathcal{U}
\diamond_{\bf \Phi} \mathcal{V}^{H}
+ \mathcal{U}_\perp \diamond_{\bf \Phi} \mathcal{V}^{H}, \mathcal{Z} + \mathcal{W}\rangle \nonumber \\
 = &~ \langle\mathcal{U} \diamond_{\bf \Phi}
\mathcal{V}^{H},\mathcal{Z}\rangle + \langle\mathcal{U}_\perp
\diamond_{\bf \Phi} \mathcal{V}^{H}_\perp,
\mathcal{P}_{T^{\bot}}(\mathcal{W})\rangle +
\langle\mathcal{U} \diamond_{\bf \Phi} \mathcal{V}^{H}, \mathcal{W}\rangle \nonumber\\
 = &~ \|\mathcal{Z}\|_{\textup{TTNN}} +
\|\mathcal{P}_{T^{\bot}}(\mathcal{W})\|_{\textup{TTNN}} +\langle \mathcal{U}
\diamond_{\bf \Phi} \mathcal{V}^{H}, \mathcal{W}\rangle \nonumber \\
\geq &~
\|\mathcal{Z}\|_{\textup{TTNN}}+\mathcal{P}_{T^{\bot}}(\mathcal{W})\|_{\textup{TTNN}}
 -|\langle\mathcal{Y}-\mathcal{U}\diamond_{\bf \Phi} \mathcal{V}^{H}, \mathcal{W}\rangle
 - \langle\mathcal{Y}, \mathcal{W}\rangle| \nonumber\\
\geq &  ~ \|\mathcal{Z}\|_{\textup{TTNN}}+
\|\mathcal{P}_{T^{\bot}}(\mathcal{W})\|_{\textup{TTNN}}- \|\mathcal{P}_{T^{\bot}}(\mathcal{Y})\|
\|\mathcal{P}_{T^{\bot}}(\mathcal{W})\|_{\textup{TTNN}}
 \nonumber\\
&~-\|\mathcal{P}_{T}(\mathcal{Y}) - \mathcal{U}
\diamond_{\bf \Phi} \mathcal{V}^{H}\|_{{F}} \|\mathcal{P}_{T}(\mathcal{W})\|_{{F}}  \nonumber\\
 \geq  &~ \|\mathcal{Z}\|_{\textup{TTNN}}+
\frac{1}{2}\|\mathcal{P}_{T^{\bot}}(\mathcal{W})\|_{\textup{TTNN}}  -
\frac{1}{4n_{(1)}n_3} \|\mathcal{P}_T(\mathcal{W})\|_{{F}}. \label{eq21}
\end{align}
 Thus, it follows from Lemma \ref{4.7} that $ \|\mathcal{Z} + \mathcal{W}\|_{\textup{TTNN}}>\|\mathcal{Z}\|_{\textup{TTNN}}$ holds
for any $\mathcal{W}$ with $\mathcal{P}_{\Omega}(\mathcal{W}) = 0.$
As a consequence, $\mathcal{Z}$  is the unique minimizer to $\eqref{ObjFT}.$
This completes the proof. \qed

In the next section, we demonstrate that the theoretical results can be obtained under
valid incoherence conditions and the tensor completion performance of the proposed
method is better than that of other testing methods.

\section{Experimental Results}\label{Sect4}

In this section, numerical examples are presented to demonstrate the effectiveness of the proposed model.
All numerical experiments are obtained from a desktop computer running on 64-bit Windows
Operating System having 8 cores with Intel(R) Core(TM) i7-6700 CPU at 3.40GHz and 20 GB memory.

Firstly, we employ an alternating direction method of multipliers
(ADMM) \cite{fazel2013hankel, Glowinski1976Approximations} to solve problem (\ref{ObjFT}).
Let $\mathcal{Z}=\mathcal{Y}$. Then problem (\ref{ObjFT}) can be rewritten as
\begin{equation}\label{ObjLg}
\begin{split}
\min_{\mathcal{Z}} & \
\|\mathcal{Z}\|_{\textup{TTNN}} \\
 \textup{s.t.} & \ \mathcal{Z}=\mathcal{Y}, \
\mathcal{P}_\Omega(\mathcal{Y}) = \mathcal{P}_\Omega(\mathcal{M}).
\end{split}
\end{equation}
The augmented Lagrangian function associated with (\ref{ObjLg}) is defined as
$$
L(\mathcal{Z},\mathcal{Y},\mathcal{X}):=\|\mathcal{Z}\|_{\textup{TTNN}}
-\langle \mathcal{X},\mathcal{Z}-\mathcal{Y}\rangle+\frac{\beta}{2}\|\mathcal{Z}-\mathcal{Y}\|_F^2,
$$
where $\mathcal{X}\in \C^{n_{1}\times n_{2}\times n_3}$ is the Lagrangian multiplier and $\beta>0$ is the penalty parameter.
The ADMM iteration system is given as follows:
\begin{eqnarray}\label{u1211}
   &&\mathcal{Z}^{k+1}=\arg\min_{\mathcal{Z}}\Big\{L(\mathcal{Z},\mathcal{Y}^k,\mathcal{X}^k)\Big\},\\\label{w1}
   &&\mathcal{Y}^{k+1}=\arg\min_{\mathcal{Y}}\Big\{L(\mathcal{Z}^{k+1},\mathcal{Y},\mathcal{X}^k): \mathcal{P}_\Omega(\mathcal{Y}) = \mathcal{P}_\Omega(\mathcal{M})\Big\},\\ \label{u1}
   &&\mathcal{X}^{k+1}=\mathcal{X}^k-\gamma\beta\left(\mathcal{Z}^{k+1}-\mathcal{Y}^{k+1}\right),
\end{eqnarray}
where $\gamma\in(0,\frac{1+\sqrt{5}}{2})$ is the dual steplength.
It follows from \cite[Theorem 3]{song2019robust} that
the optimal solution with respect to $\mathcal{Z}$ in (\ref{u1211}) is given by
\begin{equation}\label{XS}
\mathcal{Z}^{k+1}=\mathcal{U} \diamond_{\bf \Phi} \mathcal{S}_{\beta}
\diamond_{\bf \Phi} \mathcal{V}^{H},
\end{equation}
where $\mathcal{Y}^k+\frac{1}{\beta}\mathcal{X}^k=\mathcal{U} \diamond_{\bf \Phi} \mathcal{S}
\diamond_{\bf \Phi} \mathcal{V}^{H}$,
$\mathcal{S}_{\beta}=\mathbf{\Phi}^{H}[\hat{\mathcal{S}}_{\beta}]$,
and $\hat{\mathcal{S}}_{\beta}=\max\{\hat{\mathcal{S}}_{\bf \Phi}-\frac{1}{\beta},0\}$.

The optimal solution with respect to $\mathcal{Y}$ in (\ref{w1}) is given by
\begin{equation}\label{YS}
\mathcal{Y}^{k+1}=\mathcal{P}_{\overline{\Omega}}\left(\mathcal{Z}^{k+1}-\frac{1}{\beta}\mathcal{X}^k\right)+\mathcal{P}_\Omega(\mathcal{M}),
\end{equation}
where $\overline{\Omega}$ denotes the complementary set of $\Omega$
on $\{1,\ldots,n_1\}\times\{1,\ldots,n_2\}\times\{1,\ldots,n_3\}$.
The detailed description of ADMM for solving (\ref{ObjLg}) is given in Algorithm \ref{alg1}.
\begin{algorithm}[h]
\caption{Alternating direction method of multipliers for solving
(\ref{ObjLg})} \label{alg1}
\textbf{Step 0.} Let $\tau\in(0,(1+\sqrt{5})/2), \beta>0$ be given constants.
Given $\mathcal{Y}^0, \mathcal{X}^0$. For $k=0,1,2,\ldots,$ perform the following steps: \\
\textbf{Step 1.} Compute $\mathcal{Z}^{k+1}$ by (\ref{XS}). \\
\textbf{Step 2.} Compute $\mathcal{Y}^{k+1}$ via (\ref{YS}). \\
\textbf{Step 3.} Compute $\mathcal{X}^{k+1}$ by (\ref{u1}).
\end{algorithm}

The convergence of a two-block ADMM for solving convex optimization problems has been established in \cite[Theorem B.1]{fazel2013hankel}
and the convergence of  Algorithm \ref{alg1} can be derived from this theorem easily.
We omit the details here for the sake of brevity.

The Karush-Kuhn-Tucker (KKT) conditions associated with problem (\ref{ObjLg}) are given as follows:
\begin{equation}\label{KKT}
\left\{\begin{array}{lll}
0\in\partial \|\mathcal{Z}\|_{\textup{TTNN}}-\mathcal{X}, \\
\mathcal{Z} =\mathcal{Y}, \
\mathcal{P}_\Omega(\mathcal{Y}) = \mathcal{P}_\Omega(\mathcal{M}),
\end{array}\right.
\end{equation}
where $\partial \|\mathcal{Z}\|_{\textup{TTNN}}$ denotes the subdifferential of TTNN at $\mathcal{Z}$.
Based on the KKT conditions in (\ref{KKT}), we adopt the following relative residual to
measure the accuracy of a computed solution in the numerical experiments:
$$
\eta:=\max\{\eta_x,\eta_y\},
$$
where
$$
\eta_x=\frac{\|\mathcal{Z}-\mbox{Prox}_{\|\cdot\|_{\textup{TTNN}}}(\mathcal{X}+\mathcal{Z})\|_F}{1+\|\mathcal{Z}\|_F+\|\mathcal{X}\|_F}, \ \
\eta_y=\frac{\|\mathcal{Z}-\mathcal{Y}\|_F}{1+\|\mathcal{Z}\|_F+\|\mathcal{Y}\|_F}.
$$
Here $\mbox{Prox}_f(y):=\arg\min_x\{f(x)+\frac{1}{2}\|x-y\|^2\}$.
In the practical implementation,
Algorithm \ref{alg1} will be terminated
if $\eta\leq 10^{-3}$ or the maximum number of iterations exceeds $600$.
We set $\gamma=1.618$ for the convergence of ADMM \cite{fazel2013hankel} in all experiments.
Since the penalty parameter $\beta$ is not too sensitive to the recovered results,
we set $\beta=0.05$ in the following experiments.
The relative error (Rel) is defined by
$$
\mbox{Rel}:=\frac{\|\mathcal{Z}_{est}-\mathcal{Z}\|_F}{\|\mathcal{Z}\|_F},
$$
where $\mathcal{Z}_{est}$ is the estimated tensor and $\mathcal{Z}$ is the ground-truth tensor.
To evaluate the performance of the proposed method for real-world tensors,
the peak signal-to-noise ratio (PSNR) is used to
measure the quality of the estimated tensor, which is defined as follows:
$$
\mbox{PSNR}:=10\log_{10}
\frac{n_1n_2n_3({\mathcal{Z}}_{\max}-{\mathcal{Z}}_{\min})^2}{\|\mathcal{Z}_{est}-{\mathcal{Z}}\|_F^2},
$$
where ${\mathcal{Z}}_{\max}$ and ${\mathcal{Z}}_{\min}$ denote
the maximum and minimum entries of $\mathcal{Z}$, respectively.
The structural similarity (SSIM) index \cite{wang2004image} is used to
measure the quality of the recovered images:
$$
\mbox{SSIM}:=\frac{(2\mu_x\mu_y+c_1)(2\sigma_{xy}+c_2)}{(\mu_x^2+\mu_y^2+c_1)(\sigma_x^2+\sigma_y^2+c_2)},
$$
where $\mu_x, \sigma_x$ are the mean intensities and standard deviation of the original image, respectively,
$\mu_y, \sigma_y$ denote the  mean intensities and standard deviation of the recovered images, respectively,
$\sigma_{xy}$ denotes the covariance of the original and recovered images,
and $c_1,c_2>0$ are constants.
For the real-world tensor data, the SSIM is used to denote the average SSIM values of all images.

\subsection{Transformations of tensor SVD}\label{trans3}

In this subsection, we use three kinds of transformations in the ${\bf \Phi}$-product and transformed tensor SVD.
The first two transformations are FFT (t-SVD (FFT)) and
 discrete cosine transform (t-SVD (DCT)).
The third one is based on given data to construct a
unitary transform matrix \cite{qiu2021nonlocal, song2019robust, zhang2021low}.
Note that we unfold $\mathcal{Z}$ into a matrix $Z$ along the third-dimension (called
t-SVD (data)) and take the SVD of the unfolding matrix $Z=U {\Sigma}{V}^H$.
Suppose that $\text{rank}(Z)=r$.
It is interesting to observe that $U^H$ is the optimal transformation to obtain a low rank approximation of $Z$:
$$
\min_{\mathbf{\Phi},B}\ \|\mathbf{\Phi} Z-B\|_F^2 \quad \mbox{s.t.} \quad \text{rank}(B)=r, \ \mathbf{\Phi}^H\mathbf{\Phi} = \mathbf{\Phi}\mathbf{\Phi}^H=I.
$$
It has been demonstrated that the chosen unitary transformation $U^H$ is very effective
for the tensor completion problems in the literature, e.g., \cite{qiu2021nonlocal, song2019robust, zhang2021low}.
In practice, the estimator of $\mathcal{Z}$
obtained by t-SVD (DCT) can be used to generate $\mathbf{\Phi}$ for tensor completion.

 Now we give the computational cost of TTNN based on the three transformations for any $n_1 \times n_2 \times n_3$ tensor,
which is the main cost of Algorithm \ref{alg1}. Suppose that $n_2 \leq n_1$.
The computational cost of TTNN is given as follows:
\begin{itemize}
\item The application of FFT or DCT to a tube ($n_3$-vector) is of $O(n_3 \log(n_3))$ operations.
There are $n_1n_2$ tubes in an $n_1 \times n_2 \times  n_3$ tensor.  In the transformed
tensor SVD based on FFT or DCT, we need to compute $n_3$ $n_1$-by-$n_2$ SVDs in the transformed domain
and then the cost is $O(n_1n_2^2n_3)$ for these matrices.
Hence, the total cost of TTNN based on FFT or DCT is of
$O(n_1n_2n_3 \log(n_3)+n_1n_2^2n_3)$ operations.

\item The application
of a unitary transformation ($n_3$-by-$n_3$) to an $n_3$-vector is of $O(n_3^2)$ operations.
And there are still $n_3$ $n_1$-by-$n_2$ SVDs to be calculated in the transformed domain.
Therefore, the total cost of computing the TTNN based on the given data
is of $O(n_1n_2n_3^2+n_1n_2^2n_3)$ operations.
\end{itemize}

\subsection{Recovery Results}

In this subsection, we show the recovery results to demonstrate
the performance of our analysis for synthetic data and real imaging data sets.

\subsubsection{Synthetic Data}

For the synthetic data,
the random tensors are generated as follows:
$\mathcal{Z}=\mathcal{A} \diamond_{\bf \Phi} \mathcal{B}\in\mathbb{C}^{n_1\times n_2\times n_3}$
with different transformed multi-rank $\mathbf{r}$,
where $\hat{\mathcal{A}}_{{\bf\Phi}}^{(i)}$ and $\hat{\mathcal{B}}_{{\bf\Phi}}^{(i)}$
are generated by MATLAB commands $\mbox{randn}(n_1,r_i)$ and $\mbox{randn}(n_2,r_i)$,
and $r_i$ is the $i$-th element of the transformed multi-rank $\mathbf{r}$.
Here ${\bf \Phi}$ denotes FFT, DCT, and an orthogonal matrix generated
by the SVD of the unfolding matrix of ${\cal Z}$ along the third-dimension, see
Section \ref{trans3}.

\begin{figure}[!t]
	\centering
	\subfigure[t-SVD (FFT)]{
		\begin{minipage}[b]{0.99\textwidth}
			\centerline{\scriptsize } \vspace{1.5pt}
			\includegraphics[width=5.5in,height=1.8in]{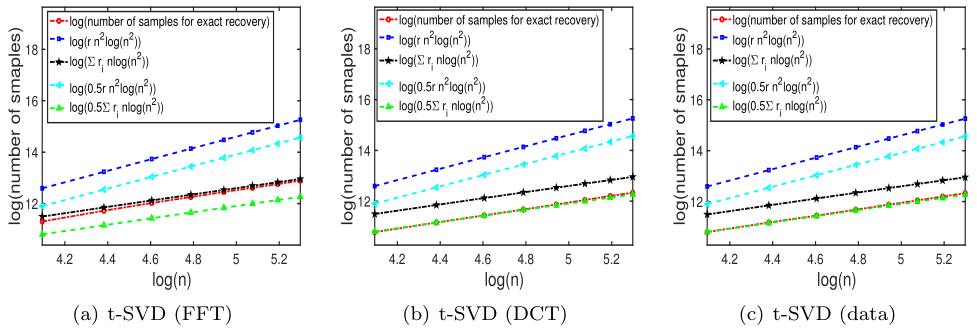}
		\end{minipage}
	}
	\caption{\small  The number of samples of exact recovery, the sample sizes required by the right hand sides of (\ref{con1}) and (\ref{con2})
	for different values of $n$ with $\sum_{i=1}^{n}r_i=200$.}
	\label{fixsum200}

\vspace{2mm}
	\centering
	\subfigure{
		\begin{minipage}[b]{0.99\textwidth}
			\centerline{\scriptsize } \vspace{1.5pt}
			\includegraphics[width=5.5in,height=1.8in]{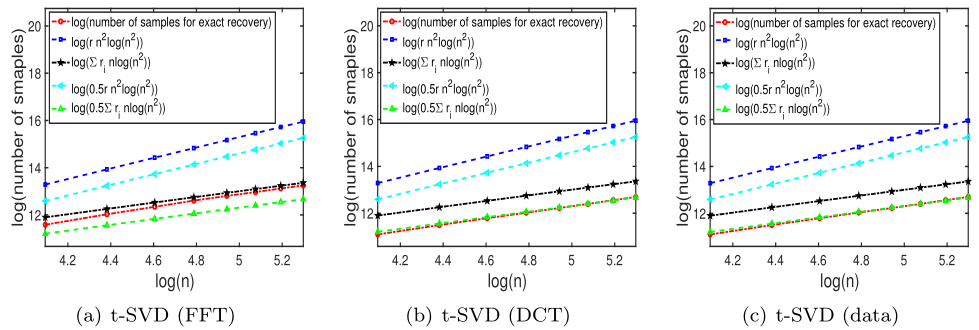}
		\end{minipage}
	}
	\caption{\small  The number of samples of exact recovery, the sample sizes required by the right hand sides of (\ref{con1}) and (\ref{con2})
	for different values of $n$ with $\sum_{i=1}^{n}r_i=300$.}\label{fixsum300}
\end{figure}

 In  Figures \ref{fixsum200} and \ref{fixsum300}, we show the actual number of sample
sizes for exact recovery and the theoretical bounds of sample sizes requirements in Theorem \ref{Theorem1}:
\begin{equation} \label{con1}
m \ge \ {\rm constant}_1 \ \sum_{i=1}^{n}r_i n \log(n^2)
\end{equation}
and
the results in \cite{Zhang2017}:
\begin{equation} \label{con2}
m \ge \ {\rm constant}_2 \ r n^2 \log(n^2)
\end{equation}
for the $n\times n\times n$
tensor with the fixed sum of the transformed multi-rank and fixed transformed
tubal rank by using
t-SVD (FFT), t-SVD (DCT), and t-SVD (data).
In the randomly generated tensor, we set
(i) $\max_{1 \le i \le n} r_i =10$
(the transformed tubal rank is 10)
and $\sum_{i=1}^nr_i=200$; and
(ii) $\max_{1 \le i \le n} r_i =20$
(the transformed tubal rank is 20)
and $\sum_{i=1}^nr_i=300$.

In Figures  \ref{fixsum200} and \ref{fixsum300}, we test different values of $n$
(from 60 to 200 with the increment size being 20).
The exact recovery means that five trials are tested
and all of the relative errors are less than or equal to $10^{-2}$ in the experiments.
We test constant$_1$, constant$_2$ = 0.5 or 1 in the right hand sides of
(\ref{con1}) and (\ref{con2}) respectively to check
how the theoretical bounds of sample sizes
match with the actual number of samples for different values
of $n$.
It can be seen from Figures \ref{fixsum200} and \ref{fixsum300} that
the curve $\sum_{i=1}^{n}r_in\log(n^2)$ based on the proposed
bound with constant$_1=1$ is close to the curve for the number of samples
required by using t-SVD (FFT), and the curves
$0.5 \sum_{i=1}^{n}r_i n\log(n^2)$ based on the proposed bound
with constant$_1=0.5$ is
close to the curves for the number of samples required
by using t-SVD (DCT) and t-SVD (data).
Note that the corresponding slope of these lines in Figures \ref{fixsum200} and  \ref{fixsum300} is equal to 1 derived by the $n$ term.
In contrast, the curves
constant$_2 r n^2 \log(n^2)$ based on
the results in \cite{Zhang2017} do not fit
the curves for the number of samples required
by using t-SVD (FFT), t-SVD (DCT) and t-SVD (data), see
the curves with constant$_2 = 0.5, 1$ in Figures \ref{fixsum200} and  \ref{fixsum300}.
The main reason is that the corresponding slope of the lines
in Figures \ref{fixsum200} and  \ref{fixsum300} is equal to 2 derived by the $n^2$ term.
According to Figures \ref{fixsum200} and  \ref{fixsum300}, we find that
the theoretical bounds of sample sizes requirements in Theorem \ref{Theorem1} match with
the actual number of sample
sizes for recovery.

\subsubsection{Hyperspectral Images}\label{HPDS}

\begin{figure}[!t]
\centering
\subfigure{
\begin{minipage}[b]{0.99\textwidth}
\centerline{\scriptsize } \vspace{1.5pt}
              \includegraphics[width=5.5in,height=2.8in]{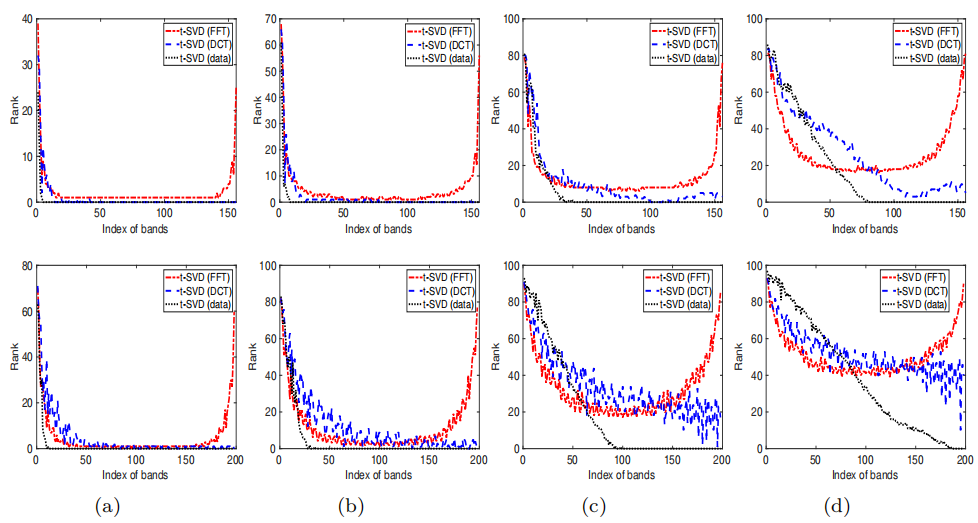}
\end{minipage}
}
       \caption{\small The distribution of transformed multi-ranks of the hyperspectral data sets with different truncations in each band.
       First row: Samson data set.
       Second row: Japser Ridge data set.
       (a) $\varpi=70\%$.
       (b) $\varpi=80\%$.
       (c) $\varpi=90\%$.
       (d) $\varpi=95\%$.}\label{rankhyp}
\end{figure}

In this subsection, two hyperspectral data sets
(Samson ($95\times95\times156$) and Japser Ridge ($100\times100\times198$) \cite{zhu2014spectral})
are used to
demonstrate
the required number of samples for tensor recovery by the proposed bound.
For Samson data, $n_1$ and $n_2$ are equal to 95 and $n_3$ is equal to 156.
For Japser Ridge data, $n_1$ and $n_2$ are equal to 100 and $n_3$ is equal to 198.
Here we compare our method with
the low-rank tensor completion method using the sum of nuclear norms of
unfolding matrices of a tensor (LRTC)\footnote{\footnotesize http://www.cs.rochester.edu/$\sim$jliu/} \cite{huang2014provable, liu2013},
tensor factorization method (TF)\footnote{\footnotesize https://homes.cs.washington.edu/$\sim$sewoong/papers.html} \cite{jain2014provable}, Square Deal \cite{mu2014square},
gradient descent algorithm on Grassmannians (GoG) \cite{Xia2019On}.
These testing hyperspectral data is normalized on $[0, 1]$.
Their theoretical estimation of samples required are presented in
Table \ref{table1}.

We remark that these hyperspectral images are not exactly low multi-rank tensors,
the multi-rank ($\sum_{i=1}^{n_3} r_i$) is not available.
A truncated tensor is used to compute the transformed multi-rank and tubal rank by using the threshold
 $\varpi$.
Here for a given tensor $\mathcal{Z}$ with transformed  tensor SVD in (\ref{equ12}) of Theorem \ref{them1}
and $\varpi$,
we determine the smallest value $k$ such that
$$
\frac{ \sum_{i=1}^{k} \varrho_i }
{ \sum_{i=1}^{n_{(2)} n_3} \varrho_i } \ge \varpi,
$$
where $\{ \varrho_i \}$ is the sorted value in ascending order of the numbers
$\{ (\hat{{\cal S}}_{\bf \Phi})_{jj\ell} \}_{1 \le j \le n_{(2)},1 \le \ell \le n_3}$
appearing in the diagonal tensor ${\cal S}$ in (\ref{equ12}). The ratio is used to determine
the significant numbers are kept in the truncated tensor based on
the threshold $\varpi$. Now we can define the transformed multi-rank
$\mathbf{r}(\varpi)$ of the truncated tensor as follows:
$$
\mathbf{r}(\varpi):=(r_1(\varpi),\ldots, r_{n_{3}}(\varpi)), \
\mbox{with}\ r_{\ell}(\varpi):=\# \{  (\hat{{\cal S}}_{\bf \Phi})_{jj\ell} \geq \varrho_k, 1 \le j \le n_{(2)} \},
\ \ell=1,\ldots, n_{3},
$$
Accordingly, the transformed tubal rank of the truncated tensor is defined as
$r(\varpi):=\max\{r_1(\varpi),\ldots, r_{n_{3}}(\varpi)\}$.
The distributions of the transformed  multi-ranks of the two hyperspectral
data sets with different $\varpi$ are shown in Figure \ref{rankhyp}.
It can be seen that the $\sum_{i=1}^{n_3} r_i(\varpi)$ obtained by t-SVD (data) is much smaller than
that obtained by t-SVD (FFT) and t-SVD (DCT) for different truncations $\varpi=70\%,  80\%,  90\%, 95\%$,
which implies that the  t-SVD (data)  needs lower number of samples for successful recovery
than t-SVD (FFT) and t-SVD (DCT).
The distributions of the transformed multi-ranks of the truncated tensor obtained by t-SVD (FFT) are symmetric
due to symmetry of FFT.

\begin{table}[!t]
\scriptsize
  \begin{center}
          \setlength{\abovecaptionskip}{-1pt}
       \setlength{\belowcaptionskip}{-1pt}
   \caption{\small The PSNR, SSIM values, and CPU time (in seconds) of different methods for the hyperspectral images.}\label{tab4}
  \medskip
   \mbox{\hspace{-0.5cm}}\begin{tabular}{|c| c| c   c  c c c c c | } \hline
                       &      &\multicolumn{7}{c|}{Samson}                  \\   \cline{3-9}
                       & const$_1$  &  20    &  30    & 40     & 50     & 60  & 70  & 80       \\
                       \cline{2-9}
											& sampling     &  \multirow{2}{*}{0.013}     & \multirow{2}{*}{0.019} & \multirow{2}{*}{0.026} & \multirow{2}{*}{0.032} & \multirow{2}{*}{0.039} & \multirow{2}{*}{0.045} & \multirow{2}{*}{0.052} \\
                       & rate    &       &  &  &  &  &  &  \\ \hline \hline
 \multirow{3}{*}{LRTC} & PSNR & 13.08  & 16.00  & 18.34  & 19.82  & 20.99  & 22.12  & 22.98  \\
                       & SSIM & 0.3254 & 0.5720 & 0.6223 & 0.6643 & 0.6964 & 0.7287 & 0.7548  \\
                       & CPU  & 26.44  & 29.81  & 29.50  & 27.13  & 25.50  & 26.35  & 28.66  \\ \hline
 \multirow{3}{*}{TF}   & PSNR & 4.05   & 7.45   & 20.26  & 29.78  & 32.86  & 35.54  & 36.39 \\
                       & SSIM & 0.3990 & 0.4852 & 0.7798 & 0.8627 & 0.8856 & 0.9334 & 0.9440  \\
                       & CPU  & 207.55 & 207.20 & 206.90 & 206.80 & 216.99 & 210.21 & 207.86   \\ \hline
 \multirow{3}{*}{Square Deal}
                       & PSNR & 16.35  & 18.55  & 20.38  & 22.15  & 23.77  & 25.57  & 26.77  \\
                       & SSIM & 0.4502 & 0.5727 & 0.6696 & 0.7269 & 0.7916 & 0.8310 & 0.8667 \\
                       & CPU  & 37.46  & 34.93  & 33.05  & 30.96  & 29.48  & 29.63  & 34.05  \\ \hline
 \multirow{3}{*}{GoG}  & PSNR & 16.72  & 26.02  & 27.13  & 30.01  & 30.45  & 30.87  & 31.43  \\
                       & SSIM & 0.4851 & 0.6978 & 0.7560 & 0.8372 & 0.8366 & 0.8398 & 0.8665  \\
                       & CPU  & 1.05e4 & 1.40e4 & 3.41e4 & 3.87e4 & 4.61e4 & 3.46e4 & 4.15e4  \\ \hline
 \multirow{3}{*}{t-SVD (FFT)}
                       & PSNR & 22.09  & 23.88  & 25.21  & 26.27  & 27.09  & 27.68  & 28.52  \\
                       & SSIM & 0.5743 & 0.6498 & 0.6996 & 0.7442 & 0.7689 & 0.7912 & 0.8107 \\
                       & CPU  & 30.24  & 26.30  & 25.11  & 24.07  & 22.45  & 23.62  & 26.58  \\ \hline
 \multirow{3}{*}{t-SVD (DCT)}
                       & PSNR & 26.35  & 28.72  & 30.42  & 31.92  & 33.03  & 33.96  & 35.04   \\
                       & SSIM & 0.7355 & 0.8098 & 0.8552 & 0.8875 & 0.9096 & 0.9215 & 0.9343 \\
                       & CPU  & 172.41 & 167.56 & 177.44 & 185.69 & 193.36 & 187.29 & 190.13  \\ \hline
 \multirow{3}{*}{t-SVD (data)}
       & PSNR&{\bf28.24}&{\bf31.37}&{\bf33.39}&{\bf35.25}&{\bf36.53} &{\bf 37.09}   & {\bf 38.62}  \\
                       & SSIM & 0.8324 & 0.8971 & 0.9311 & 0.9539 & 0.9643 & 0.9658 & 0.9746 \\
                       & CPU  & 241.48 & 258.86 & 270.17 & 282.82 & 284.68 & 263.78 & 291.26 \\ \hline\hline
                              &                    &\multicolumn{7}{c|}{Japser Ridge}  \\   \cline{3-9}
                       & const$_1$     &  20    &  30    & 40     & 50     & 60   & 70  & 80     \\
                       \cline{2-9}
					 & sampling     & \multirow{2}{*}{0.010} & \multirow{2}{*}{0.015} & \multirow{2}{*}{0.020}  & \multirow{2}{*}{0.025} & \multirow{2}{*}{0.030} & \multirow{2}{*}{0.035} & \multirow{2}{*}{0.040} \\
                       & rate    &       &  &  &  &  &  & \\ \hline \hline
 \multirow{3}{*}{LRTC} & PSNR & 12.48  & 14.80  & 16.78  & 18.17  & 18.97  & 19.61  & 20.15 \\
                       & SSIM & 0.3245 & 0.4104 & 0.4404 & 0.4611 & 0.4793 & 0.5011 & 0.5223 \\
                       & CPU  & 39.54  & 39.98  & 42.73  & 39.81  & 36.51  & 40.21  & 42.36 \\ \hline
 \multirow{3}{*}{TF}   & PSNR & 7.12   & 11.41  & 12.09  & 15.39  & 27.03  & 27.83  & 28.07  \\
                       & SSIM & 0.2076 & 0.3728 & 0.6120 & 0.6418 & 0.7246 & 0.7603 & 0.7693  \\
                       & CPU & 400.27 & 399.79 & 398.94 & 398.46 & 286.83  & 362.15 & 384.52 \\ \hline
 \multirow{3}{*}{Square Deal}
                       & PSNR & 15.38  & 17.50  & 20.47  & 22.48  & 24.25  & 26.13  & 27.12 \\
                       & SSIM & 0.3890 & 0.5168 & 0.6199 & 0.7021 & 0.7596 & 0.8046 & 0.8378 \\
                       & CPU  & 43.33  & 41.89  & 41.82  & 41.99  & 40.65  & 42.59  & 40.99 \\ \hline
 \multirow{3}{*}{GoG}  & PSNR & 15.32  & 22.38  & 23.54  & 24.35  & 25.44  & 25.90  & 26.35 \\
                       & SSIM & 0.2643 & 0.5430 & 0.5624 & 0.5835 & 0.6550 & 0.6808 & 0.7193 \\
                       & CPU  & 3.49e4 & 4.13e4 & 5.33e4 & 5.50e4 & 5.58e4 & 5.39e4 & 5.41e4 \\ \hline
 \multirow{3}{*}{t-SVD (FFT)}
                       & PSNR & 21.04  & 22.95  & 24.09  & 25.17  & 26.06  & 26.74  & 27.25 \\
                       & SSIM & 0.4877 & 0.5802 & 0.6353 & 0.6891 & 0.7270 & 0.7578 & 0.7756  \\
                       & CPU  & 50.54  & 42.01  & 37.62  & 35.38  & 34.36  & 42.69  & 35.99 \\ \hline
 \multirow{3}{*}{t-SVD (DCT)}
                       & PSNR & 21.93  & 23.40  & 24.48  & 25.49  & 26.36  & 27.06  & 27.65  \\
                       & SSIM & 0.5206 & 0.5945 & 0.6481 & 0.7015 & 0.7378 & 0.7647 & 0.7892 \\
                       & CPU  & 200.90 & 171.96 & 164.15 & 169.65 & 172.05 & 184.23 & 193.06 \\ \hline
 \multirow{3}{*}{t-SVD (data)}
                       & PSNR &{\bf23.54}&{\bf25.63}&{\bf27.47}&{\bf28.74}&{\bf30.09}&{\bf 30.78} &{\bf 31.33} \\
                       & SSIM & 0.6004 & 0.7041 & 0.7816 & 0.8323 & 0.8675 & 0.8777 & 0.8962  \\
                       & CPU  & 235.74 & 217.07 & 217.13 & 230.64 & 235.28 & 225.95 & 231.87 \\ \hline
    \end{tabular}
  \end{center}
\end{table}

\begin{figure}[htbp]
\centering
\subfigure[15th band]{
\begin{minipage}[b]{0.2\textwidth}
\centerline{\scriptsize } \vspace{1pt}
              \includegraphics[width=1.2in,height=0.76in]{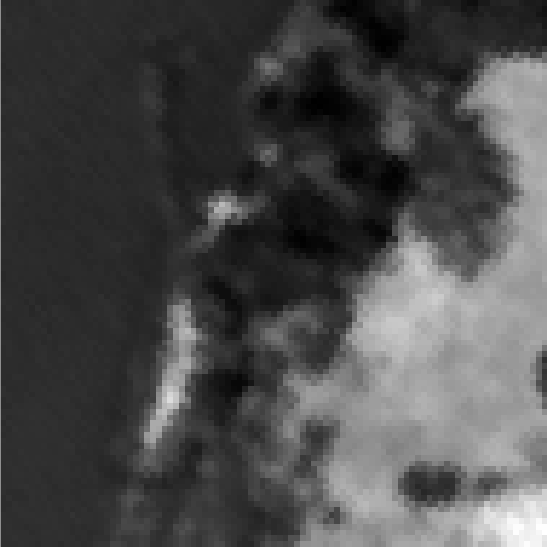}
\centerline{\scriptsize }\\ \vfill \vspace{-23pt}
              \includegraphics[width=1.2in,height=0.76in]{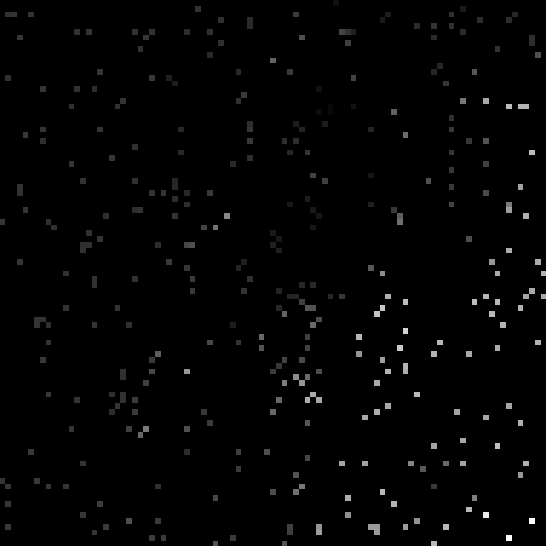}
\centerline{\scriptsize }\\ \vfill \vspace{-23pt}
              \includegraphics[width=1.2in,height=0.76in]{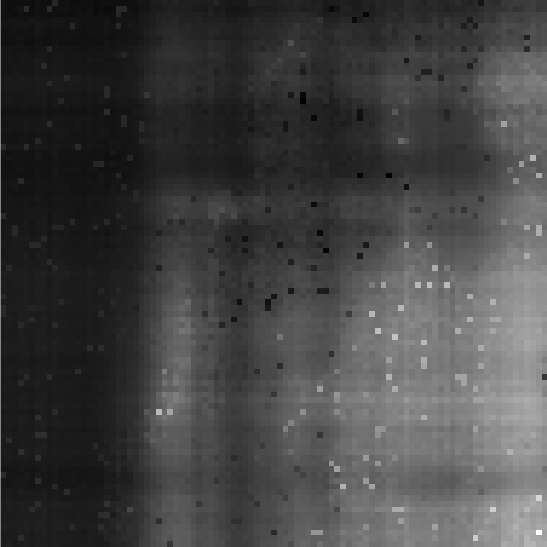}
\centerline{\scriptsize }\\ \vfill \vspace{-23pt}
              \includegraphics[width=1.2in,height=0.76in]{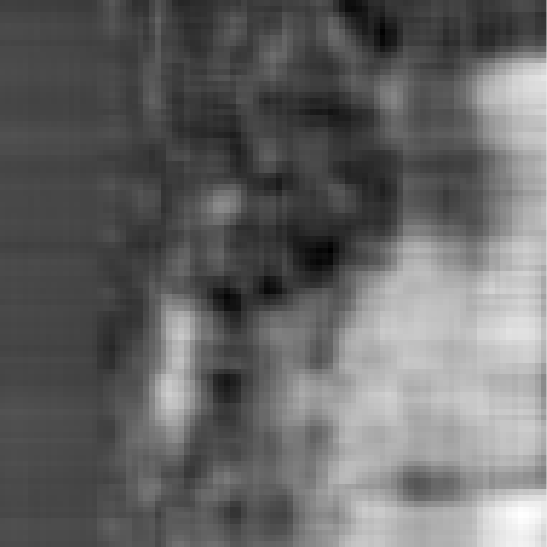}
\centerline{\scriptsize }\\ \vfill \vspace{-23pt}
\includegraphics[width=1.2in,height=0.76in]{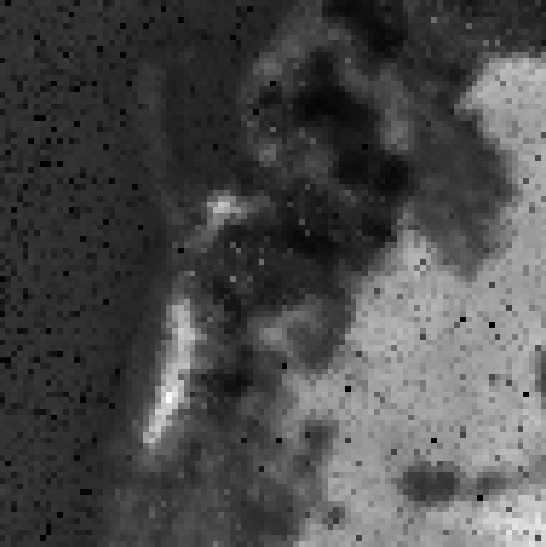}
\centerline{\scriptsize }\\ \vfill \vspace{-23pt}
\includegraphics[width=1.2in,height=0.76in]{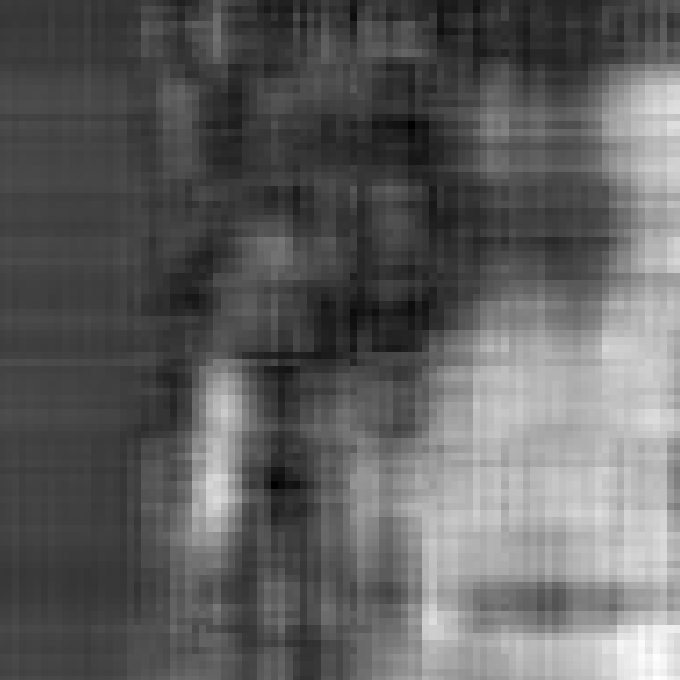}
\centerline{\scriptsize }\\ \vfill \vspace{-23pt}
              \includegraphics[width=1.2in,height=0.76in]{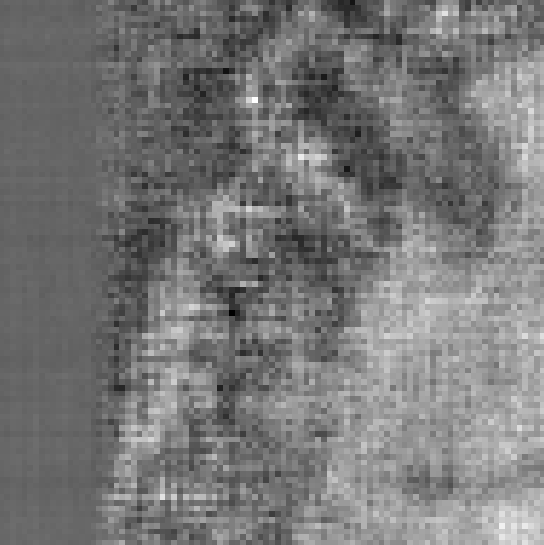}
\centerline{\scriptsize }\\ \vfill \vspace{-23pt}
              \includegraphics[width=1.2in,height=0.76in]{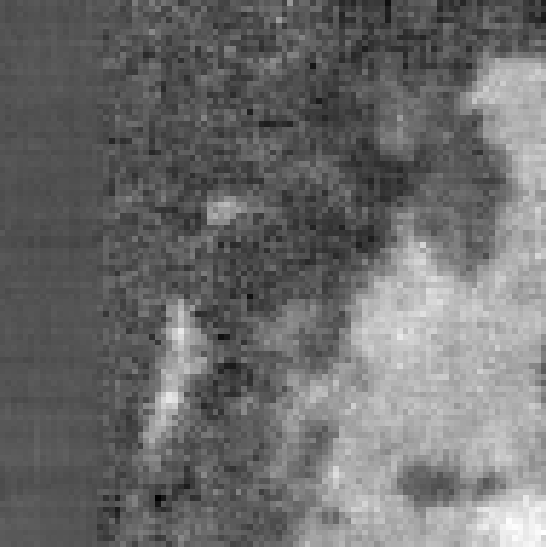}
\centerline{\scriptsize }\\ \vfill \vspace{-23pt}
              \includegraphics[width=1.2in,height=0.76in]{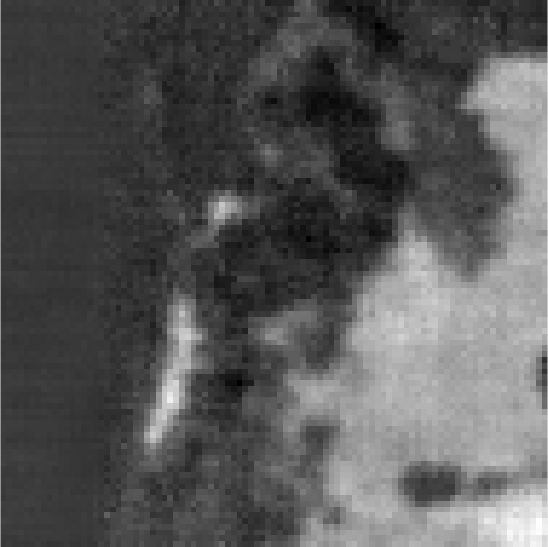}
\end{minipage}
}
\subfigure[60th band]{
\begin{minipage}[b]{0.2\textwidth}
\centerline{\scriptsize } \vspace{1pt}
              \includegraphics[width=1.2in,height=0.76in]{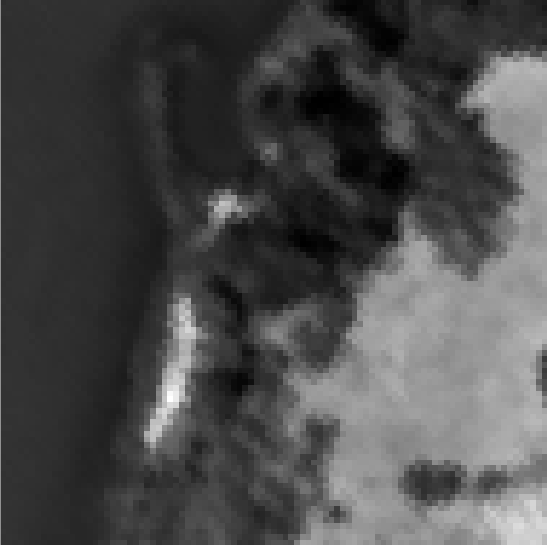}
\centerline{\scriptsize }\\ \vfill \vspace{-23pt}
              \includegraphics[width=1.2in,height=0.76in]{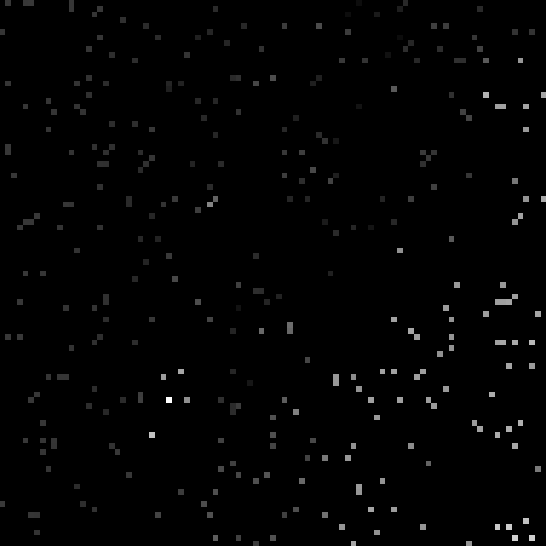}
\centerline{\scriptsize }\\ \vfill \vspace{-23pt}
              \includegraphics[width=1.2in,height=0.76in]{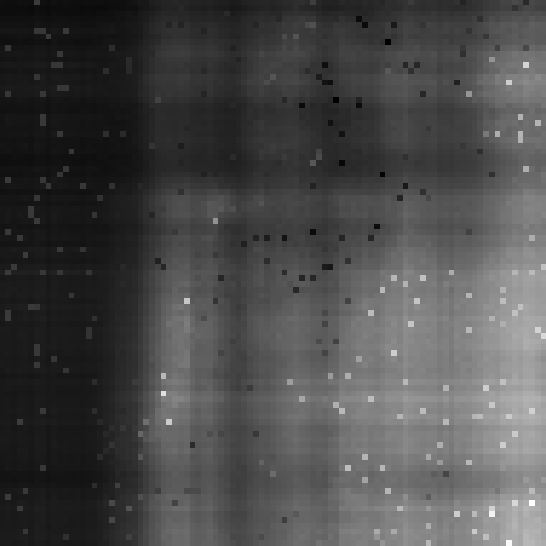}
\centerline{\scriptsize }\\ \vfill \vspace{-23pt}
              \includegraphics[width=1.2in,height=0.76in]{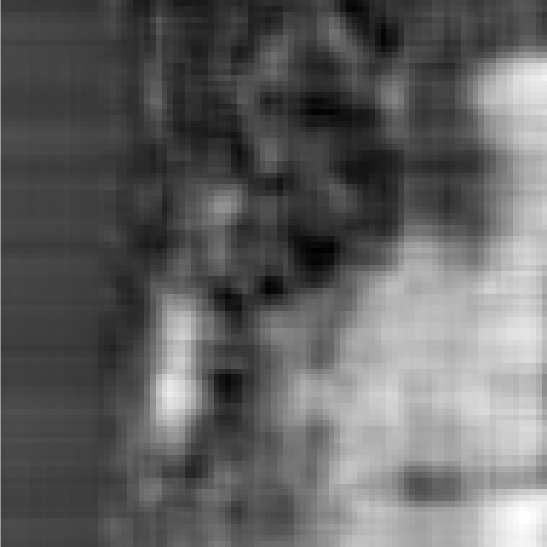}
\centerline{\scriptsize }\\ \vfill \vspace{-23pt}
\includegraphics[width=1.2in,height=0.76in]{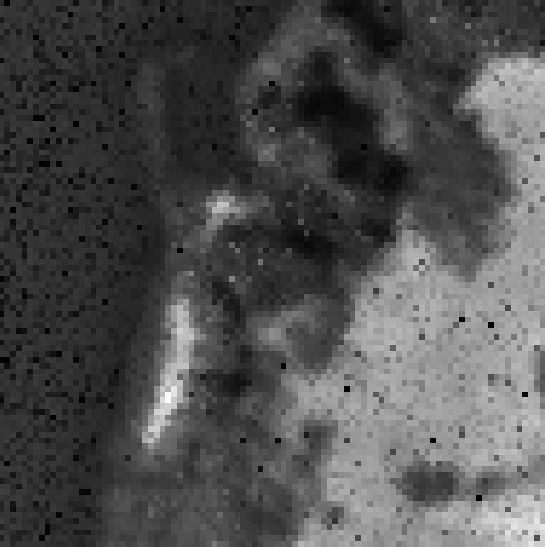}
\centerline{\scriptsize }\\ \vfill \vspace{-23pt}
\includegraphics[width=1.2in,height=0.76in]{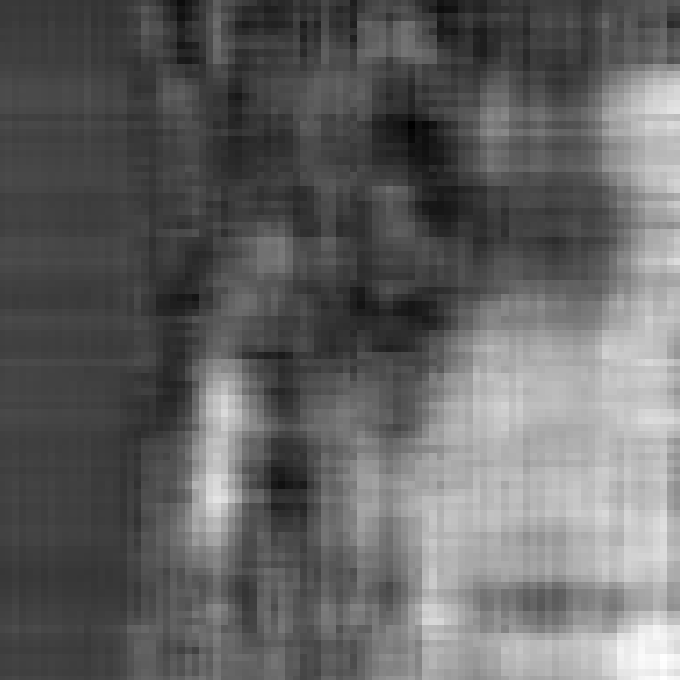}
\centerline{\scriptsize }\\ \vfill \vspace{-23pt}
              \includegraphics[width=1.2in,height=0.76in]{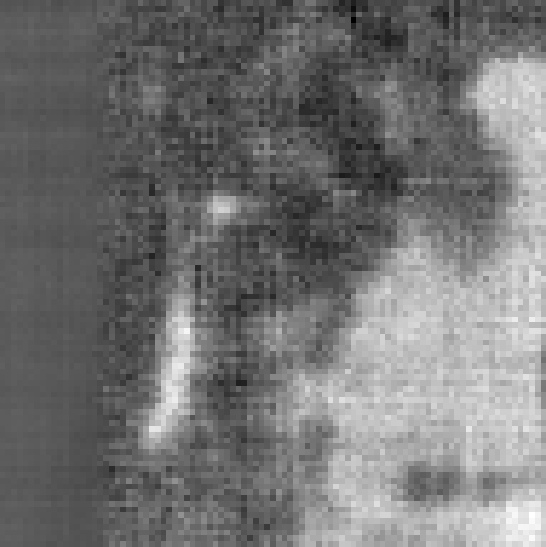}
\centerline{\scriptsize }\\ \vfill \vspace{-23pt}
              \includegraphics[width=1.2in,height=0.76in]{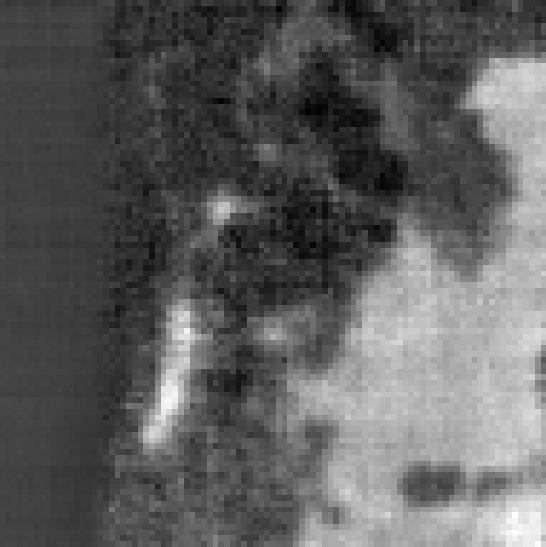}
\centerline{\scriptsize }\\ \vfill \vspace{-23pt}
              \includegraphics[width=1.2in,height=0.76in]{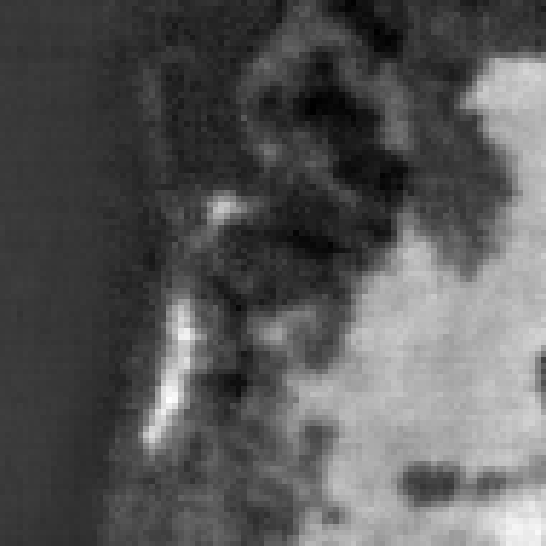}
\end{minipage}
}
\subfigure[90th band]{
\begin{minipage}[b]{0.2\textwidth}
\centerline{\scriptsize } \vspace{1pt}
              \includegraphics[width=1.2in,height=0.76in]{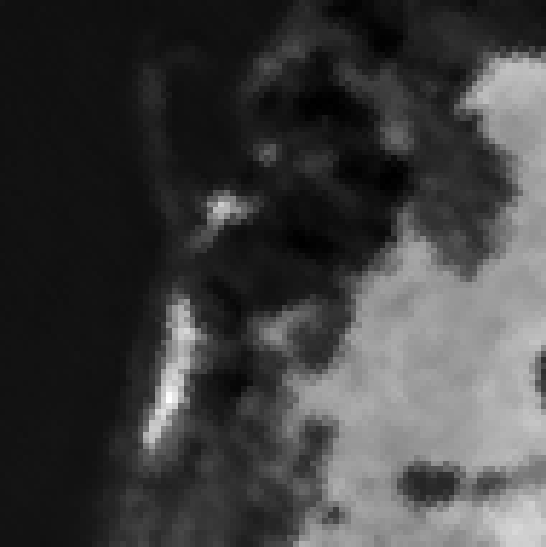}
\centerline{\scriptsize }\\ \vfill \vspace{-23pt}
              \includegraphics[width=1.2in,height=0.76in]{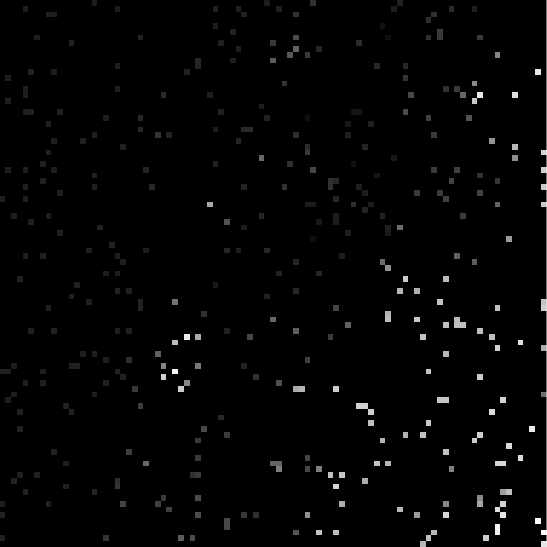}
\centerline{\scriptsize }\\ \vfill \vspace{-23pt}
              \includegraphics[width=1.2in,height=0.76in]{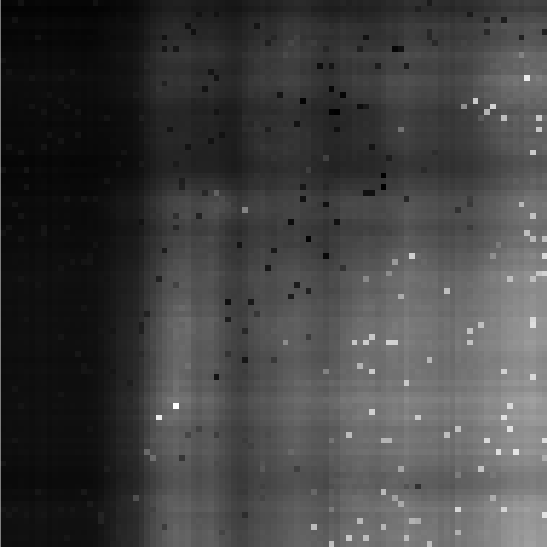}
\centerline{\scriptsize }\\ \vfill \vspace{-23pt}
              \includegraphics[width=1.2in,height=0.76in]{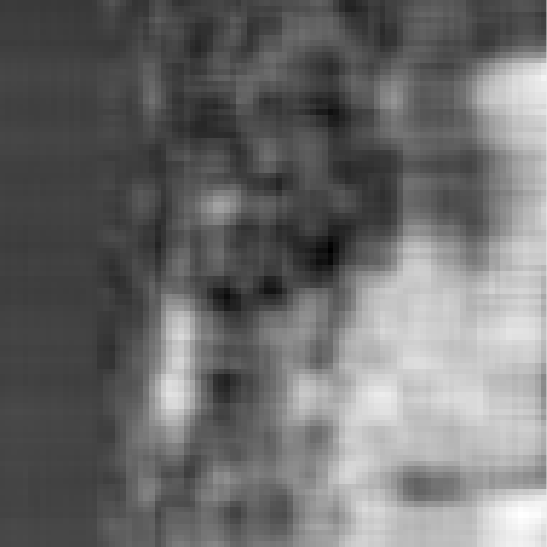}
\centerline{\scriptsize }\\ \vfill \vspace{-23pt}
\includegraphics[width=1.2in,height=0.76in]{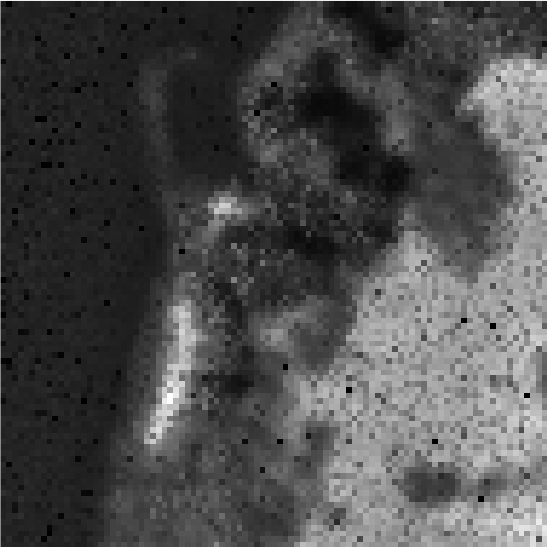}
\centerline{\scriptsize }\\ \vfill \vspace{-23pt}
\includegraphics[width=1.2in,height=0.76in]{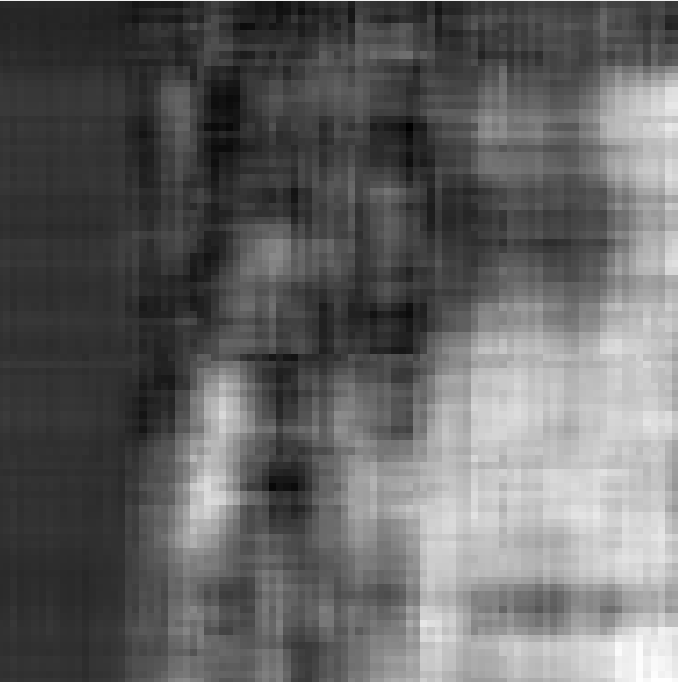}
\centerline{\scriptsize }\\ \vfill \vspace{-23pt}
              \includegraphics[width=1.2in,height=0.76in]{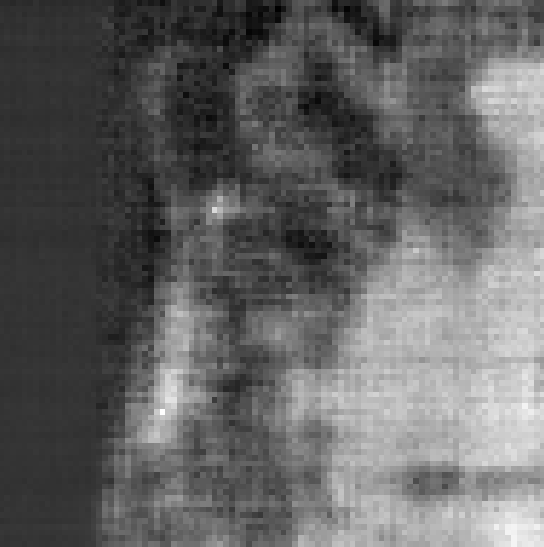}
\centerline{\scriptsize }\\ \vfill \vspace{-23pt}
              \includegraphics[width=1.2in,height=0.76in]{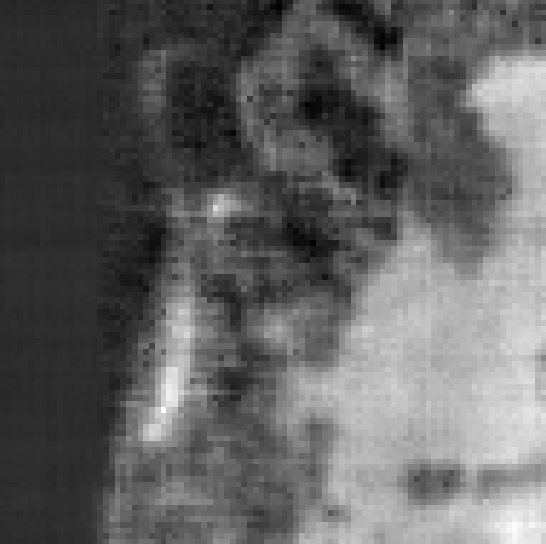}
\centerline{\scriptsize }\\ \vfill \vspace{-23pt}
              \includegraphics[width=1.2in,height=0.76in]{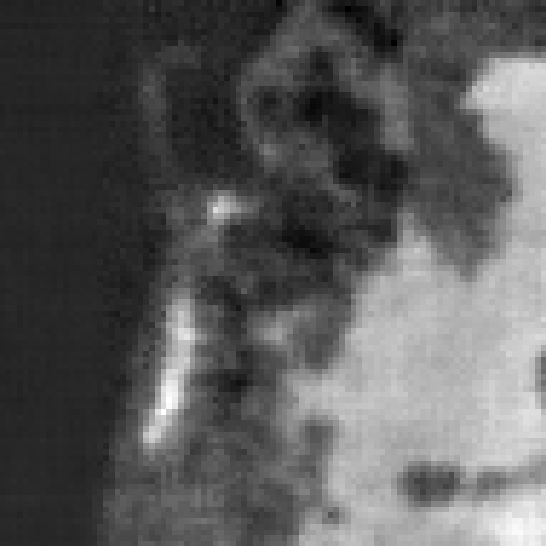}
\end{minipage}
}
\subfigure[130th band]{
\begin{minipage}[b]{0.2\textwidth}
\centerline{\scriptsize } \vspace{1pt}
              \includegraphics[width=1.2in,height=0.76in]{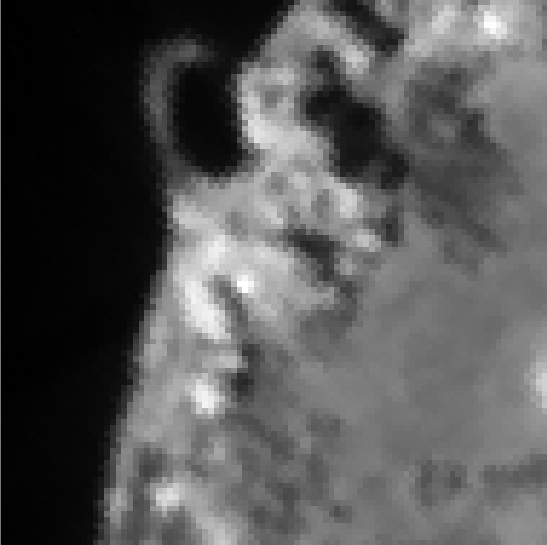}
\centerline{\scriptsize }\\ \vfill \vspace{-23pt}
              \includegraphics[width=1.2in,height=0.76in]{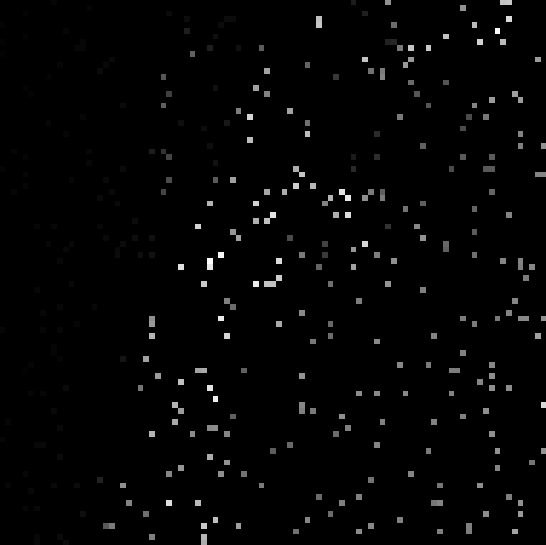}
\centerline{\scriptsize }\\ \vfill \vspace{-23pt}
              \includegraphics[width=1.2in,height=0.76in]{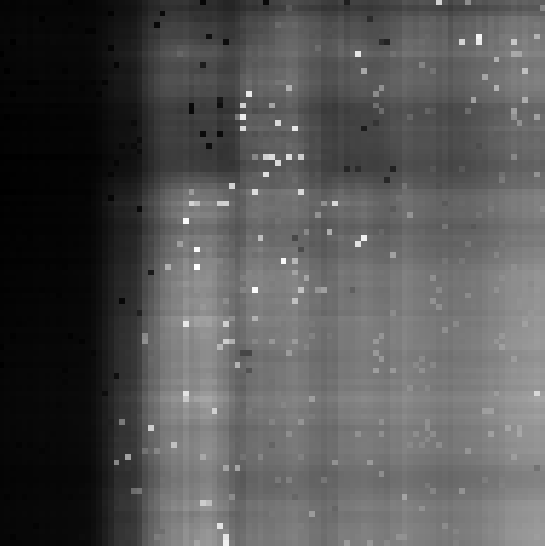}
\centerline{\scriptsize }\\ \vfill \vspace{-23pt}
              \includegraphics[width=1.2in,height=0.76in]{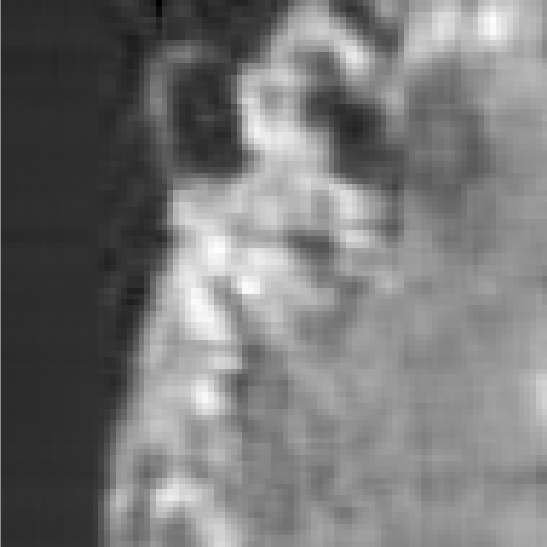}
\centerline{\scriptsize }\\ \vfill \vspace{-23pt}
\includegraphics[width=1.2in,height=0.76in]{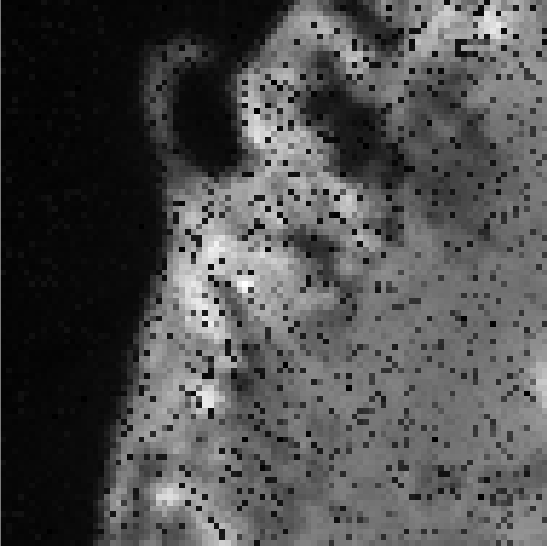}
\centerline{\scriptsize }\\ \vfill \vspace{-23pt}
\includegraphics[width=1.2in,height=0.76in]{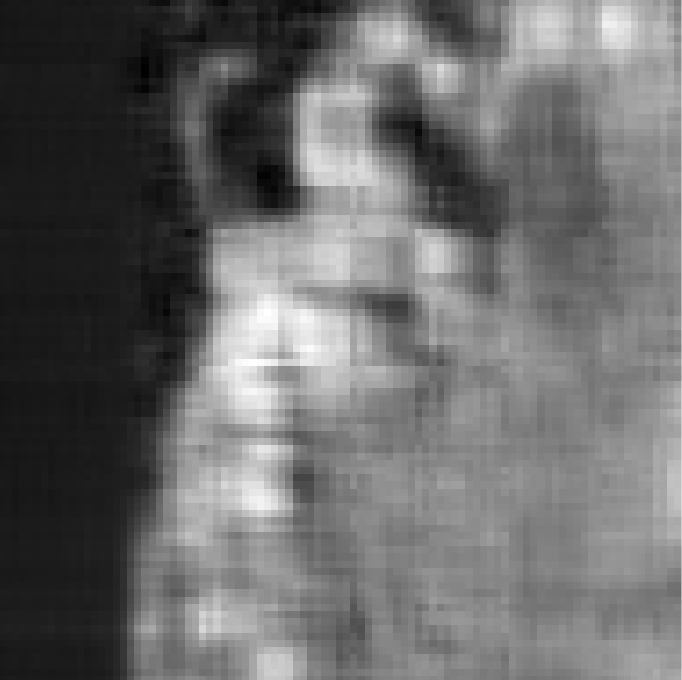}
\centerline{\scriptsize }\\ \vfill \vspace{-23pt}
              \includegraphics[width=1.2in,height=0.76in]{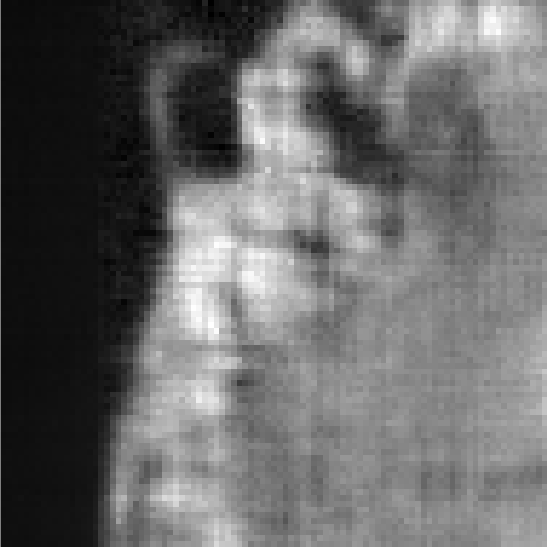}
\centerline{\scriptsize }\\ \vfill \vspace{-23pt}
              \includegraphics[width=1.2in,height=0.76in]{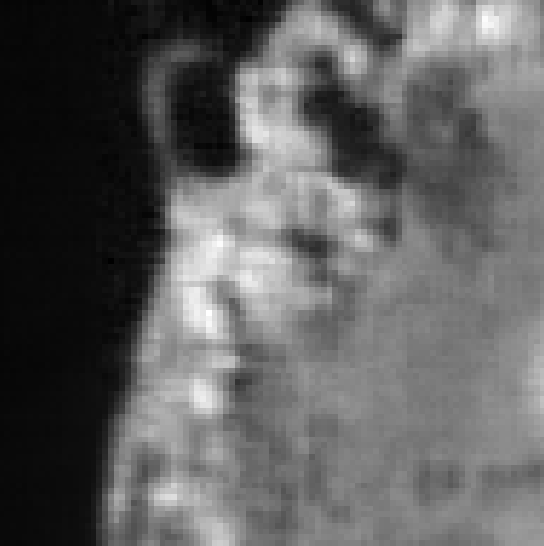}
\centerline{\scriptsize }\\ \vfill \vspace{-23pt}
              \includegraphics[width=1.2in,height=0.76in]{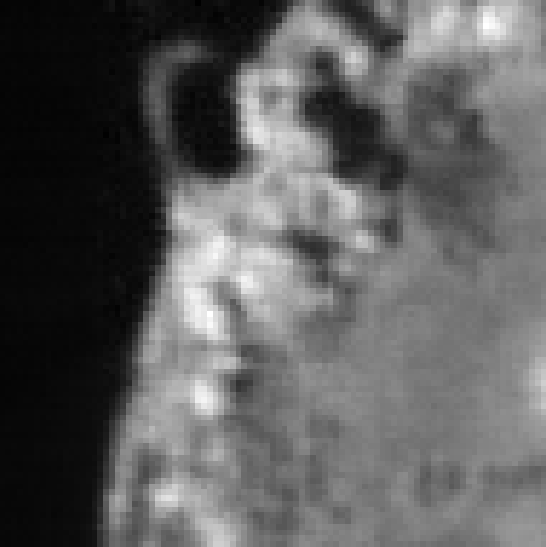}
\end{minipage}
}
       \caption{\small Recovery results by different methods for the Samson data with const$_1=60$.
       First row: Original images.
       Second row: Observed images.
       Third row: Recovered images  by LRTC.
       Fourth row: Recovered images by TF.
       Fifth row: Recovered images by Square Deal.
       Sixth row: Recovered images by GoG.
       Seventh row: Recovered images by t-SVD (FFT).
       Eighth row: Recovered images by t-SVD (DCT).
       Ninth row: Recovered images by t-SVD (data).}\label{HPVisc}
\end{figure}

\begin{figure}[!t]
\centering
\subfigure{
\begin{minipage}[b]{0.99\textwidth}
\centerline{\scriptsize } \vspace{1.5pt}
              \includegraphics[width=5.5in,height=2.8in]{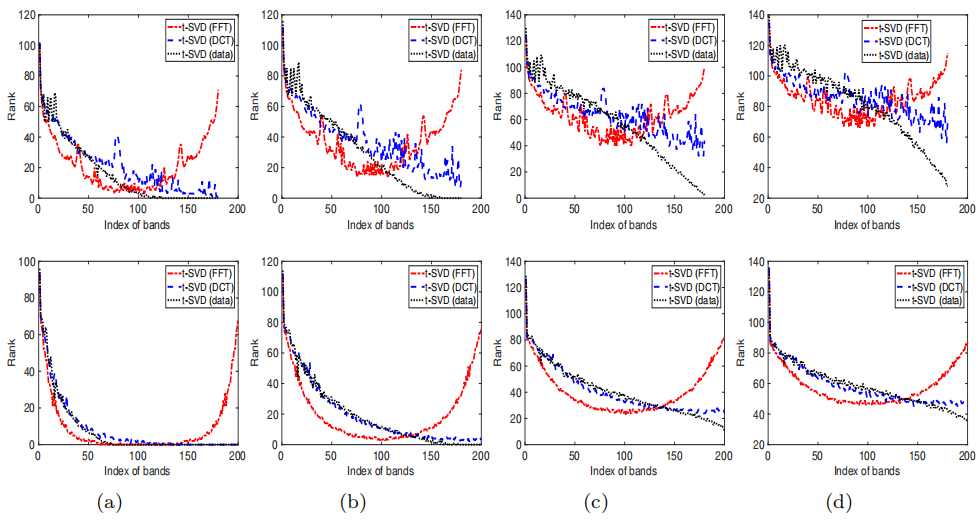}
\end{minipage}
}
       \caption{\small The distribution of transformed multi-ranks of the video data sets with different truncations in each frame.
       First row: Carphone  data set.
       Second row: Announcer data set.
       (a) $\varpi=70\%$.
       (b) $\varpi=80\%$.
       (c) $\varpi=90\%$.
       (d) $\varpi=95\%$.}\label{rankAK}
\end{figure}

In Table \ref{tab4}, we present the number of samples (const$_1 n_{(1)} \log (n_{(1)} n_3 )$) for several values
of const$_1$ and their corresponding PSNR and SSIM values
of the recovered tensors by different methods, where const$_1$ varies from
20 to 80 with step-size $10$.
Here we can regard const$_1$ as $c_0\mu\sum_{i=1}^{n_3}r_i(\varpi)$ in (\ref{new1}) of Theorem \ref{Theorem1}.
For $\varpi = 70\%$, the values $\sum_{i=1}^{n_3}r_i(\varpi)$ of Samson hyperspectral
image are $346$, $129$, $40$
for  t-SVD (FFT), t-SVD (DCT), and t-SVD (data), respectively.
Thus, the chosen const$_1$ is proportional to $\sum_{i=1}^{n_3}r_i(\varpi)$
for the testing tensor of hyperspectral image with approximately low transformed multi-rank.
We can see from the table that for
different values of const$_1$,
the PSNR and SSIM values obtained by t-SVD (data) and t-SVD (DCT)
are higher than those obtained by LRTC, TF, Square Deal, GoG, and t-SVD (FFT).
Moreover, when const$_1 =60$, the recovery performance of the t-SVD (FFT),
t-SVD (DCT), and t-SVD (data) is good enough in terms of PSNR and SSIM values,
which implies the number of sizes is enough for successful recovery by Theorem \ref{Theorem1}.
When const$_1$ is larger, e.g., const$_1 =70,80$, the improvements of PSNR values of the t-SVD (FFT), t-SVD (DCT), and t-SVD (data)
are smaller than those of other small const$_1$.
For Samson data, the performance of GoG is better than that of t-SVD (FFT) for const$_1 \ge 30$. But
the computational time required by GoG is significantly more than that required by the other methods.
For Japser Ridge data, the PSNR and SSIM values obtained by all three t-SVD methods
are almost higher than those obtained by LRTC, TF, Square Deal, and GoG. Moreover, the computational time
required by the t-SVD methods are quite efficient compared with the other methods.

Figure \ref{HPVisc} shows the visual comparisons of different bands obtained by LRTC, TF, Square Deal,
GoG, and three t-SVD methods for the Samson data, where const$_1=60$.
We can see that the t-SVD (data) and t-SVD (DCT) outperform LRTC, TF, Square Deal, and t-SVD (FFT) in terms of visual quality.
Moreover,  the images recovered by t-SVD (data) keep
more details than those recovered by LRTC, TF, Square Deal, t-SVD (FFT), and t-SVD (DCT).

\subsubsection{Video Data}

In this subsection, we test two video data sets (length $\times$ width $\times$ frames)
to show the performance of the proposed method,
where the testing videos include
Carphone ($144\times176\times180$) and
Announcer ($144\times176\times200$)\footnote{\footnotesize https://media.xiph.org/video/derf/},
and we just use the first channels of all frames
in the original data. Moreover, the first $180$ and $200$
frames for the two videos are chosen to improve the computational time.
The intensity range of the video images is scaled into $[0,1]$ in the experiments.

\begin{table}[!t]
\tiny
  \begin{center}
          \setlength{\abovecaptionskip}{-1pt}
       \setlength{\belowcaptionskip}{-1pt}
   \caption{\small The PSNR, SSIM values, and CPU time (in seconds) of different methods for the video data sets.}\label{tab5}
  \medskip
   \mbox{\hspace{-1.3cm}}\begin{tabular}{|c|c| c| c   c  c c  c  c  c  c    c  c c | } \hline
                  &    & const$_1$  &  20    &  30    & 40     & 50     & 60      &  70    &  80    & 90     & 100   & 110  & 120      \\
         \cline{3-14}
									&		& sampling     & \multirow{2}{*}{0.008}      & \multirow{2}{*}{0.012}  & \multirow{2}{*}{0.016} & \multirow{2}{*}{0.020} & \multirow{2}{*}{0.024} & \multirow{2}{*}{0.028} & \multirow{2}{*}{0.032}
                       & \multirow{2}{*}{0.036} & \multirow{2}{*}{0.040} & \multirow{2}{*}{0.044} & \multirow{2}{*}{0.048} \\
                   &    & rate    &       &  &  &  &  &  &
                       &  &  & &  \\ \hline \hline
 \multirow{21}{*}{Carphone}&
 \multirow{3}{*}{LRTC} & PSNR & 10.81  & 11.65  & 12.49  & 13.19  & 13.97   & 14.75  & 15.40  & 16.01  & 16.53  & 17.05  & 17.52 \\
               &       & SSIM & 0.3498 & 0.3571 & 0.3687 & 0.3870 & 0.4085  & 0.4319 & 0.4496 & 0.4701 & 0.4862 & 0.5038 & 0.5206 \\
               &       & CPU  & 90.78  & 85.06  & 85.92  & 85.12  & 89.76   & 92.28  & 91.70  & 90.47  & 92.91  & 92.06  & 95.21 \\ \cline{2-14}
& \multirow{3}{*}{TF}  & PSNR & 4.96   & 4.64   & 13.41  & 12.98  & 20.82   & 22.57  & 22.66  & 22.74  & 22.75  & 22.85  & 23.04 \\
               &       & SSIM & 0.2855 & 0.4507 & 0.5081 & 0.5499 & 0.5837  & 0.6461 & 0.6531 & 0.6607 & 0.6622 & 0.6705 & 0.6746 \\
               &       & CPU  & 780.55 & 779.63 & 771.43 & 782.11 & 762.25  & 770.57 & 771.03 & 765.42 & 760.32 & 782.45 & 795.41  \\ \cline{2-14}
& \multirow{3}{*}{Square Deal}
                       & PSNR & 10.04  & 12.99  & 13.55  & 15.00  & 16.98   & 17.15  & 18.10  & 19.15  & 20.42  & 20.47  & 21.17 \\
               &       & SSIM & 0.1796 & 0.2675 & 0.3304 & 0.3988 & 0.4497  & 0.4930 & 0.5348 & 0.5719 & 0.6113 & 0.6178 & 0.6396 \\
               &       & CPU  & 76.98  & 80.23  & 76.06  & 77.03  & 76.76   & 76.36  & 74.78  & 80.00  & 78.59  & 77.21  & 79.75 \\ \cline{2-14}
& \multirow{3}{*}{GoG} & PSNR & 18.59  & 20.18  & 20.77  & 20.46  & 21.22   & 21.35  & 21.48  & 22.11  & 22.47  & 22.86  & 23.15 \\
               &       & SSIM & 0.4010 & 0.4653 & 0.5229 & 0.5023 & 0.5449  & 0.5483 & 0.5547 & 0.5768 & 0.5874 & 0.5936 & 0.6058 \\
               &       & CPU  & 1.09e5 & 2.27e5 & 1.30e5 & 1.40e5 & 2.47e5  & 6.27e5 & 3.76e5 & 6.71e5 & 6.32e5 & 6.52e5 & 6.48e5 \\ \cline{2-14}
& \multirow{3}{*}{t-SVD (FFT)}
                       & PSNR & 9.79   & 11.55  & 15.06  & 18.53  & 22.04   & 22.59  & 23.07  & 23.38  & 23.72  & 24.01  & 24.24  \\
               &       & SSIM & 0.1814 & 0.2463 & 0.3759 & 0.4899 & 0.6121  & 0.6367 & 0.6566 & 0.6690 & 0.6816 & 0.6934 & 0.7017 \\
               &       & CPU  & 53.78  & 58.69  & 77.43  & 88.78  & 107.26  & 103.15 & 98.25  & 93.03  & 88.88  & 90.36  & 92.81 \\ \cline{2-14}
& \multirow{3}{*}{t-SVD (DCT)}
                       & PSNR & 20.24  & 21.18  & 21.93  & 22.42  & 22.91   & 23.24  & 23.60  & 23.87  & 24.15  & 24.44  & 24.65  \\
               &       & SSIM & 0.5230 & 0.5720 & 0.6044 & 0.6286 & 0.6506  & 0.6642 & 0.6772 & 0.6890 & 0.6484 & 0.7106 & 0.7185 \\
               &       & CPU  & 323.11 & 291.24 & 271.55 & 251.17 & 237.77  & 226.54 & 214.85 & 205.35 & 198.58 & 200.15 & 197.32  \\ \cline{2-14}
& \multirow{3}{*}{t-SVD (data)}
             & PSNR &{\bf20.39}&{\bf21.43}&{\bf22.30}&{\bf22.87}&{\bf23.45}&{\bf23.75}&{\bf24.23}&{\bf24.56}&{\bf24.85}&{\bf 24.94}&{\bf 25.17}  \\
               &       & SSIM & 0.5294 & 0.5841 & 0.6236 & 0.6499 & 0.6747  & 0.6877 & 0.7063 & 0.7200 & 0.7298 & 0.7329 & 0.7426 \\
               &       & CPU  & 366.20 & 337.21 & 314.40 & 293.42 & 278.06  & 261.55 & 253.14 & 242.91 & 235.86 & 261.02 & 248.59 \\ \hline
                  &    & const$_1$  &  20    &  30    & 40     & 50     & 60      &  70    &  80    & 90     & 100   & 110  & 120      \\
         \cline{3-14}
									&		& sampling     & \multirow{2}{*}{0.007}      & \multirow{2}{*}{0.011}  & \multirow{2}{*}{0.015} & \multirow{2}{*}{0.018} & \multirow{2}{*}{0.022} & \multirow{2}{*}{0.025} & \multirow{2}{*}{0.029}
                       & \multirow{2}{*}{0.033} & \multirow{2}{*}{0.036} & \multirow{2}{*}{0.040} & \multirow{2}{*}{0.044} \\
                   &    & rate    &       &  &  &  &  &  &
                       &  & & & \\ \hline \hline
 \multirow{21}{*}{Announcer}&
 \multirow{3}{*}{LRTC} & PSNR & 12.72  & 14.03  & 15.28  & 16.42  & 17.76   & 18.83  & 19.88  & 20.74  & 21.46  & 22.06  & 22.58 \\
               &       & SSIM & 0.4727 & 0.4900 & 0.5087 & 0.5336 & 0.5601  & 0.5869 & 0.6120 & 0.6341 & 0.6520 & 0.6710 & 0.6871  \\
               &       & CPU  & 100.75 & 96.25  & 100.67 & 103.05 & 104.02  & 105.00 & 104.68 & 103.02 & 100.29 & 101.23 &100.96 \\ \cline{2-14}
& \multirow{3}{*}{TF}  & PSNR & 4.39   & 9.99   & 11.36  & 15.48  & 24.12   & 25.65  & 26.22  & 26.28  & 27.99  & 28.29  & 28.34 \\
               &       & SSIM & 0.4229 & 0.6176 & 0.6599 & 0.6743 & 0.7076  & 0.7363 & 0.7534 & 0.7544 & 0.8169 & 0.8265 & 0.8300 \\
               &       & CPU  & 887.20 & 862.11 & 872.41 & 874.64 & 869.01  & 864.04 & 822.11 & 835.96 & 858.93 & 862.62 & 839.15 \\ \cline{2-14}
& \multirow{3}{*}{Square Deal}
                       & PSNR & 11.50  & 14.61  & 16.71  & 19.36  & 20.86   & 23.34  & 24.15  & 24.27  & 26.76  & 28.24  & 28.87 \\
               &       & SSIM & 0.1692 & 0.2702 & 0.3780 & 0.5004 & 0.6207  & 0.6950 & 0.7488 & 0.7928 & 0.8394 & 0.8743 & 0.8918 \\
               &       & CPU  & 86.39  & 87.33  & 80.39  & 78.26  & 77.23   & 77.46  & 80.36  & 80.28  & 81.06  & 80.25  & 80.38  \\ \cline{2-14}
& \multirow{3}{*}{GoG} & PSNR & 21.85  & 23.59  & 23.72  & 24.80   & 24.96  & 25.42  & 26.20  & 26.87  & 27.14  & 27.43  & 27.79 \\
               &       & SSIM & 0.5113 & 0.5854 & 0.6144 &  0.6691 & 0.6775 & 0.6976 & 0.7183 & 0.7566 & 0.7977 & 0.8236 & 0.8395 \\
               &       & CPU  & 1.94e5 & 2.31e5 & 2.60e5 &  3.54e5 & 6.09e5 & 6.36e5 & 5.49e5 & 5.98e5 & 5.90e5 & 5.89e5 & 6.01e5 \\ \cline{2-14}
& \multirow{3}{*}{t-SVD (FFT)}
                       & PSNR & 11.01  & 16.64  & 27.03  & 27.98  & 28.63   & 29.11  & 29.46  & 29.83  & 30.12  & 30.35  & 30.62 \\
               &       & SSIM & 0.2606 & 0.5339 & 0.8125 & 0.8380 & 0.8564  & 0.8686 & 0.8776 & 0.8865 & 0.8922 & 0.8987 & 0.9022 \\
               &       & CPU  & 61.87  & 97.82  & 153.63 & 144.82 & 132.18  & 125.67 & 118.75 & 106.63 & 103.44 & 113.25 & 108.63 \\ \cline{2-14}
& \multirow{3}{*}{t-SVD (DCT)}
                       & PSNR & 26.06  & 27.12  & 27.92  & 28.58  & 29.16   & 29.58  & 29.95  & 30.32  & 30.69  & 30.93  & 31.17 \\
               &       & SSIM & 0.7645 & 0.8078 & 0.8353 & 0.8875 & 0.8703  & 0.8798 & 0.8893 & 0.8985 & 0.9046 & 0.9099 & 0.9140 \\
               &       & CPU  & 537.80 & 466.05 & 432.45 & 407.57 & 377.23  & 350.59 & 329.40 & 298.93 & 287.79 & 298.46 & 284.32 \\ \cline{2-14}
& \multirow{3}{*}{t-SVD (data)}
        & PSNR &{\bf26.08}&{\bf27.25}&{\bf28.12}&{\bf28.76}&{\bf29.41}&{\bf29.88}&{\bf30.24}&{\bf30.71}&{\bf31.14}&{\bf 31.29}&{\bf 31.58}  \\
               &       & SSIM & 0.7649 & 0.8113 & 0.8399 & 0.8581 & 0.8741  & 0.8838 & 0.8941 & 0.9037 & 0.9106 & 0.9150 & 0.9198 \\
               &       & CPU  & 608.38 & 531.39 & 472.35 & 460.30 & 425.59  & 395.42 & 353.94 & 341.43 & 330.40 & 365.25 & 341.29 \\ \hline
    \end{tabular}
  \end{center}
\end{table}

Similar to Section \ref{HPDS}, the video data sets are not exactly low-rank.
First, we show the distributions of the transformed multi-ranks with different transformations
and truncations $\varpi$ in Figure \ref{rankAK}.
It can be observed that the $\sum_{i=1}^{n_3} r_i(\varpi)$ obtained by t-SVD (data)
is much smaller than that obtained by t-SVD (FFT) and t-SVD (DCT) for different $\varpi$.
Therefore, the number of samples required by t-SVD (data) would be smaller than that
required by t-SVD (FFT) and t-SVD (DCT) for the same performance.
The distributions of transformed multi-ranks obtained by t-SVD (FFT)
are symmetric for different $\varpi$ since the FFT has symmetric property.

\begin{figure}[htbp]
	\centering
	\subfigure[20th frame]{
		\begin{minipage}[b]{0.2\textwidth}
			\centerline{\scriptsize } \vspace{1pt}
			\includegraphics[width=1.2in,height=0.76in]{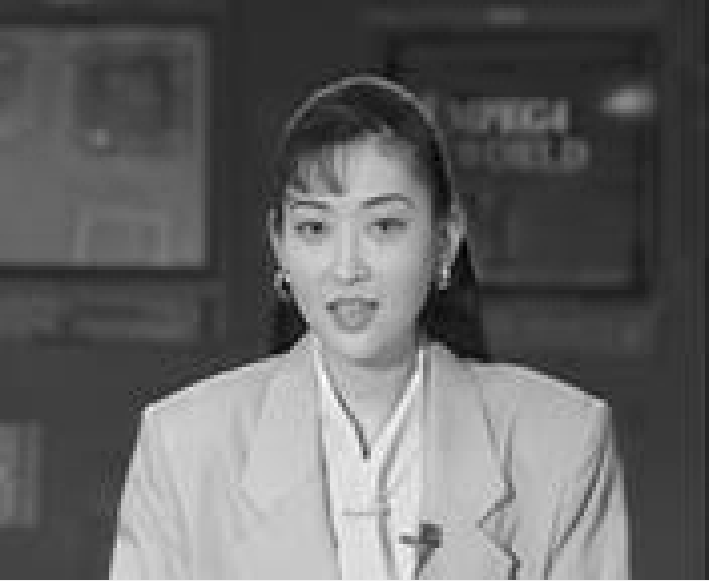}
			\centerline{\scriptsize }\\ \vfill \vspace{-23pt}
			\includegraphics[width=1.2in,height=0.76in]{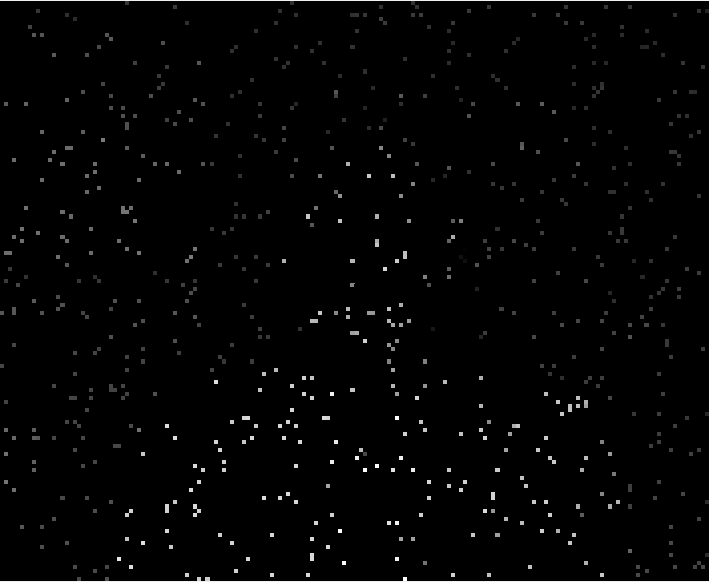}
			\centerline{\scriptsize }\\ \vfill \vspace{-23pt}
			\includegraphics[width=1.2in,height=0.76in]{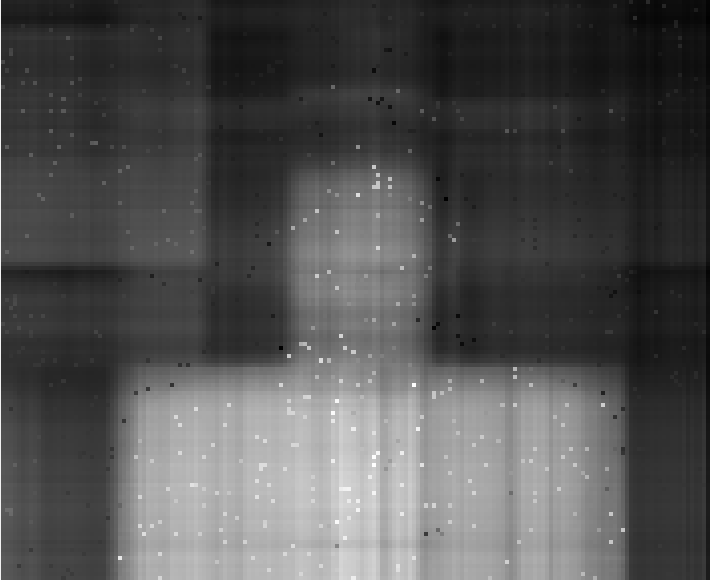}
			\centerline{\scriptsize }\\ \vfill \vspace{-23pt}
			\includegraphics[width=1.2in,height=0.76in]{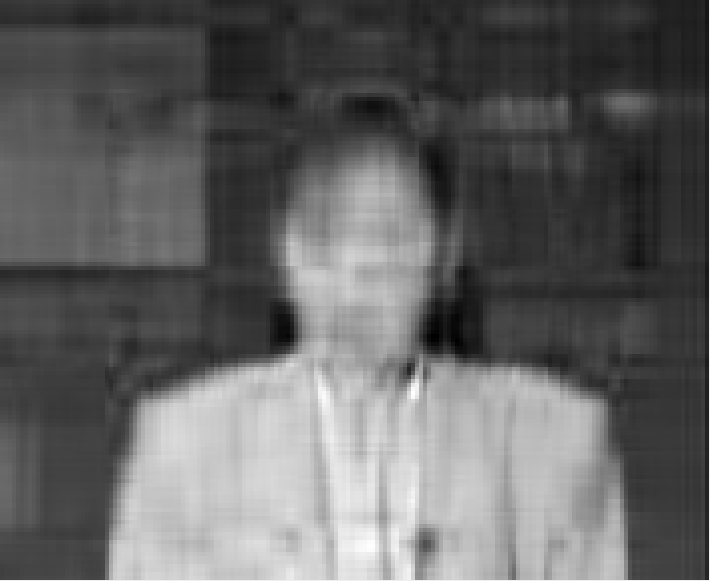}
			\centerline{\scriptsize }\\ \vfill \vspace{-23pt}
			\includegraphics[width=1.2in,height=0.76in]{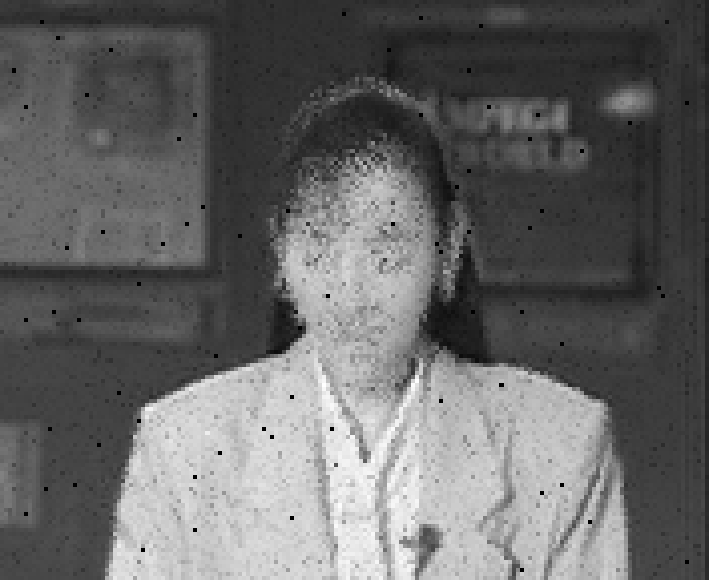}
			\centerline{\scriptsize }\\ \vfill \vspace{-23pt}
			\includegraphics[width=1.2in,height=0.76in]{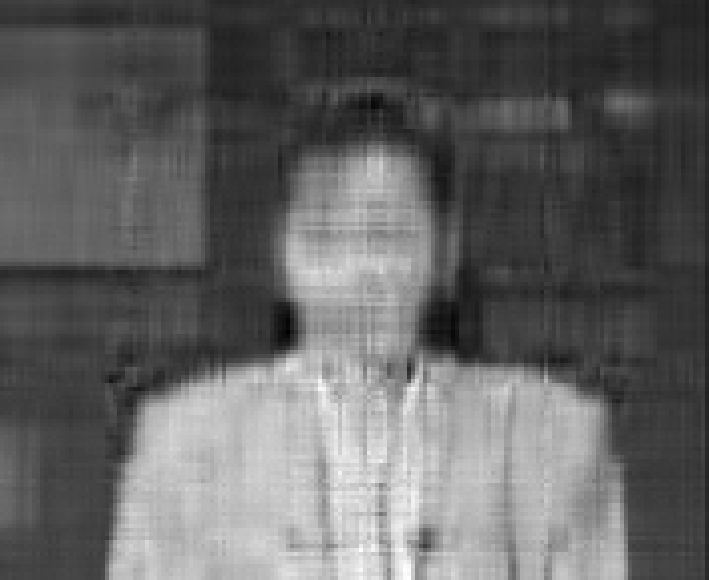}
			\centerline{\scriptsize }\\ \vfill \vspace{-23pt}
			\includegraphics[width=1.2in,height=0.76in]{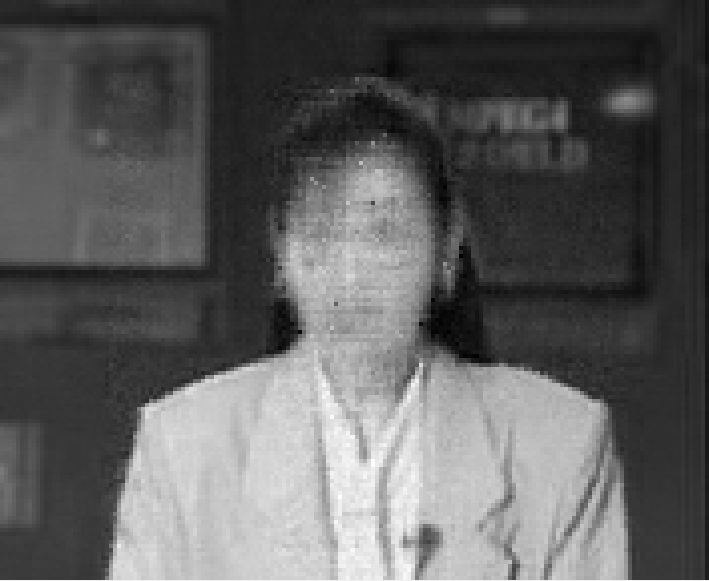}
			\centerline{\scriptsize }\\ \vfill \vspace{-23pt}
			\includegraphics[width=1.2in,height=0.76in]{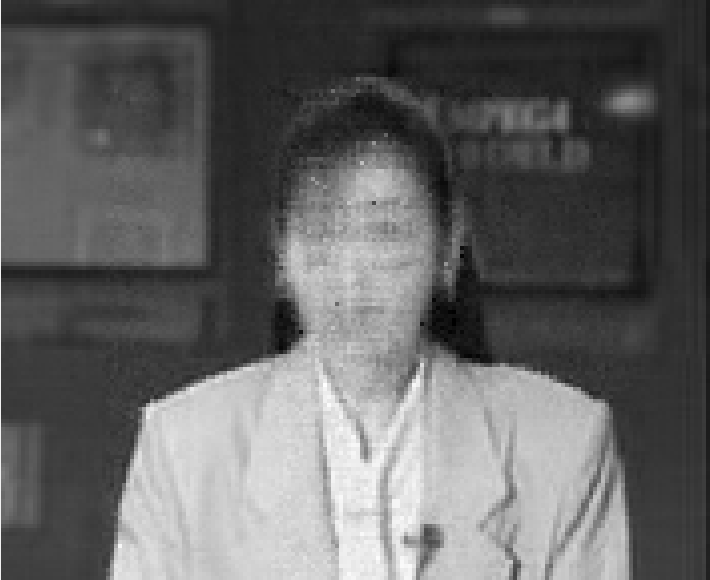}
			\centerline{\scriptsize }\\ \vfill \vspace{-23pt}
			\includegraphics[width=1.2in,height=0.76in]{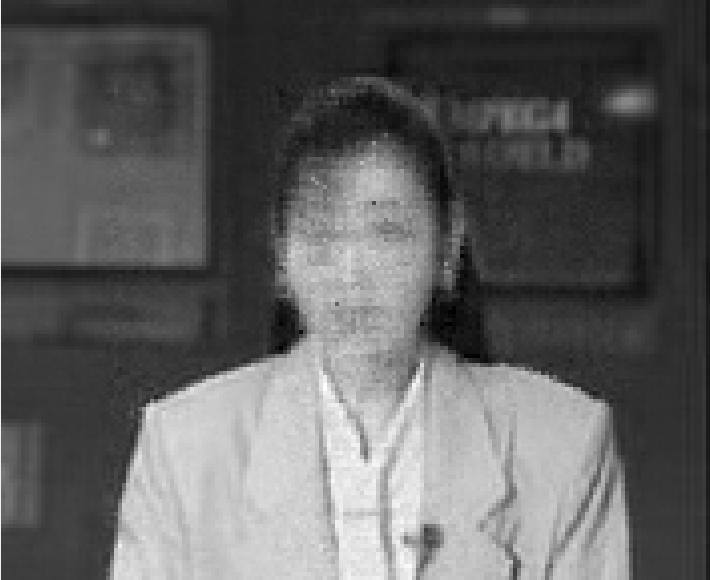}
		\end{minipage}
	}
	\subfigure[80th frame]{
		\begin{minipage}[b]{0.2\textwidth}
			\centerline{\scriptsize } \vspace{1pt}
			\includegraphics[width=1.2in,height=0.76in]{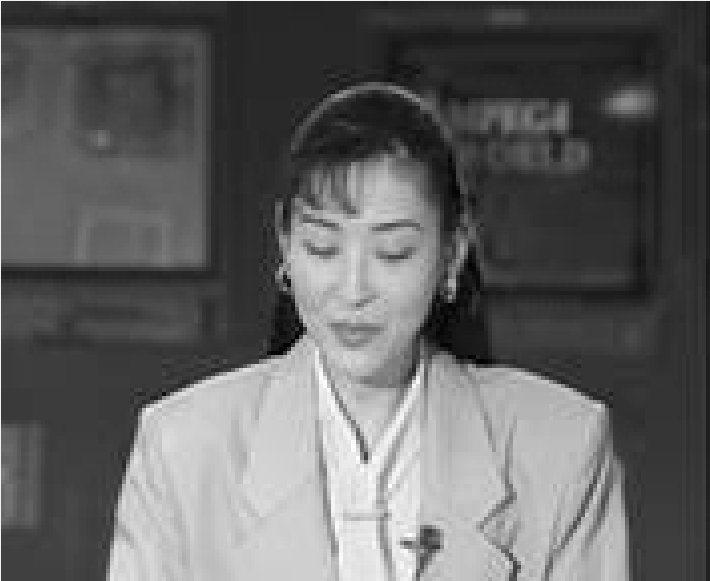}
			\centerline{\scriptsize }\\ \vfill \vspace{-23pt}
			\includegraphics[width=1.2in,height=0.76in]{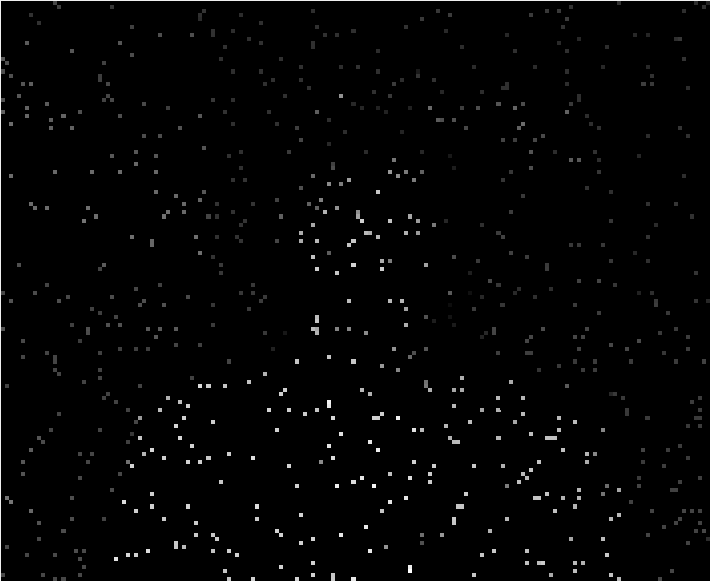}
			\centerline{\scriptsize }\\ \vfill \vspace{-23pt}
			\includegraphics[width=1.2in,height=0.76in]{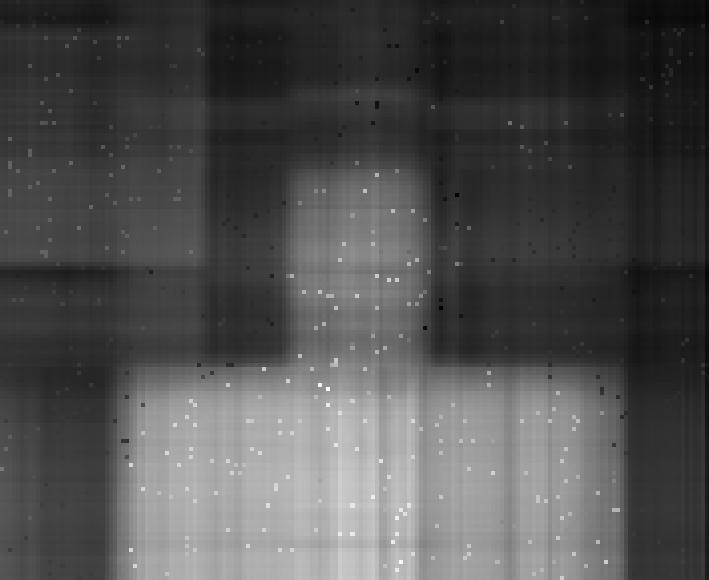}
			\centerline{\scriptsize }\\ \vfill \vspace{-23pt}
			\includegraphics[width=1.2in,height=0.76in]{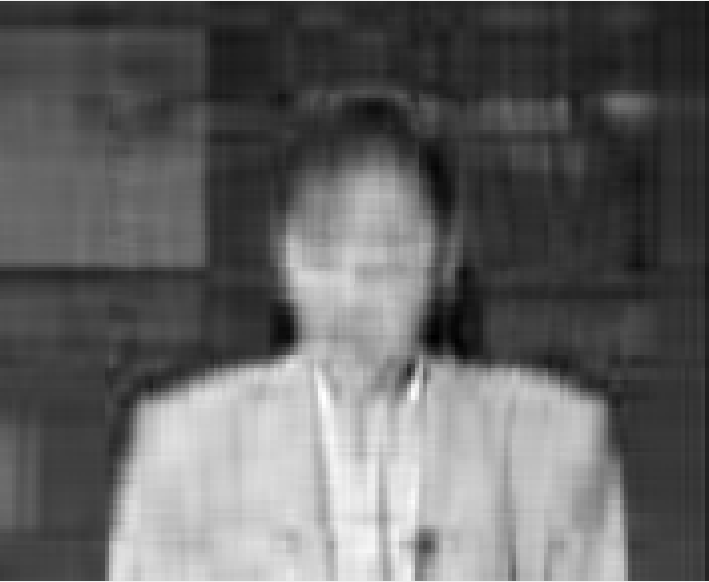}
			\centerline{\scriptsize }\\ \vfill \vspace{-23pt}
			\includegraphics[width=1.2in,height=0.76in]{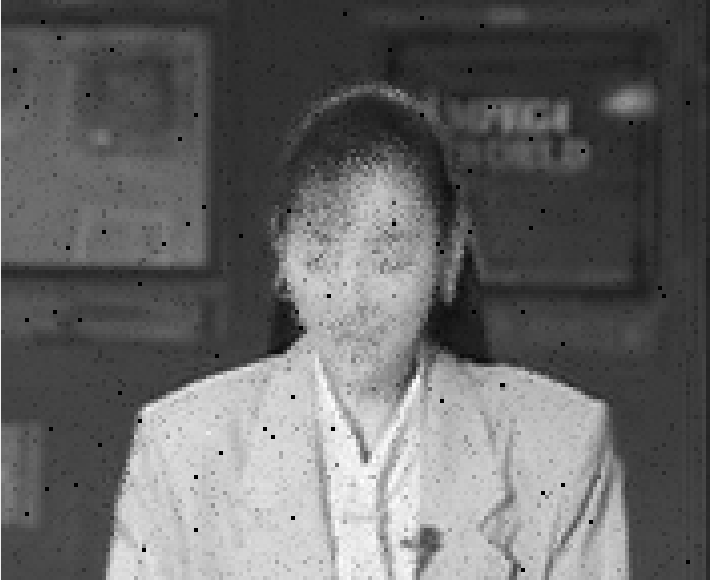}
			\centerline{\scriptsize }\\ \vfill \vspace{-23pt}
			\includegraphics[width=1.2in,height=0.76in]{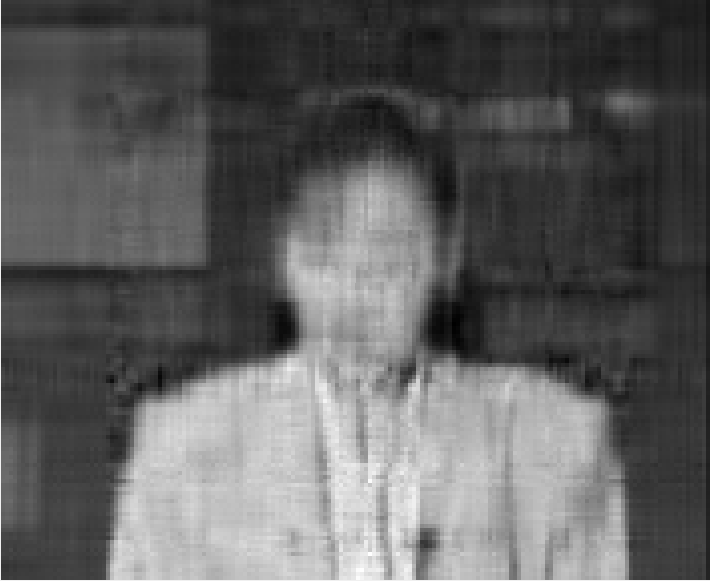}
			\centerline{\scriptsize }\\ \vfill \vspace{-23pt}
			\includegraphics[width=1.2in,height=0.76in]{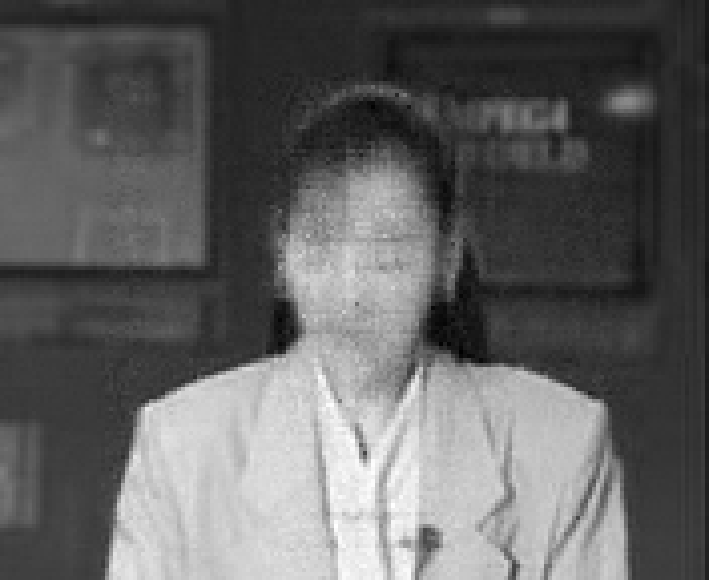}
			\centerline{\scriptsize }\\ \vfill \vspace{-23pt}
			\includegraphics[width=1.2in,height=0.76in]{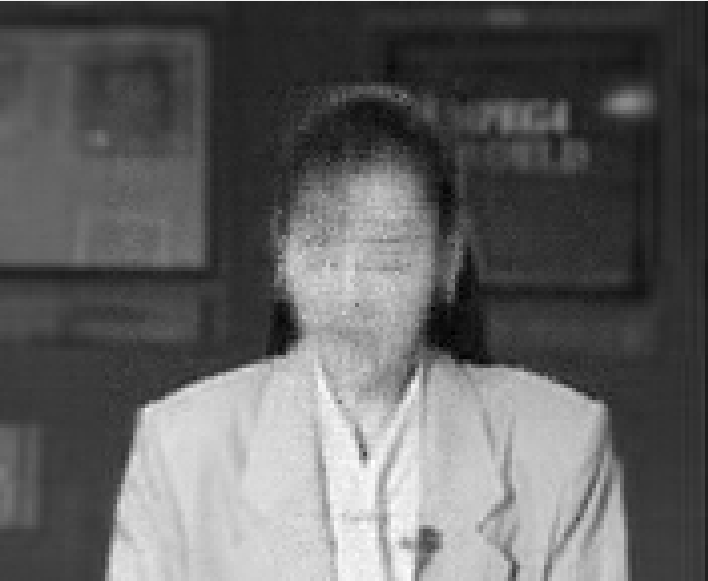}
			\centerline{\scriptsize }\\ \vfill \vspace{-23pt}
			\includegraphics[width=1.2in,height=0.76in]{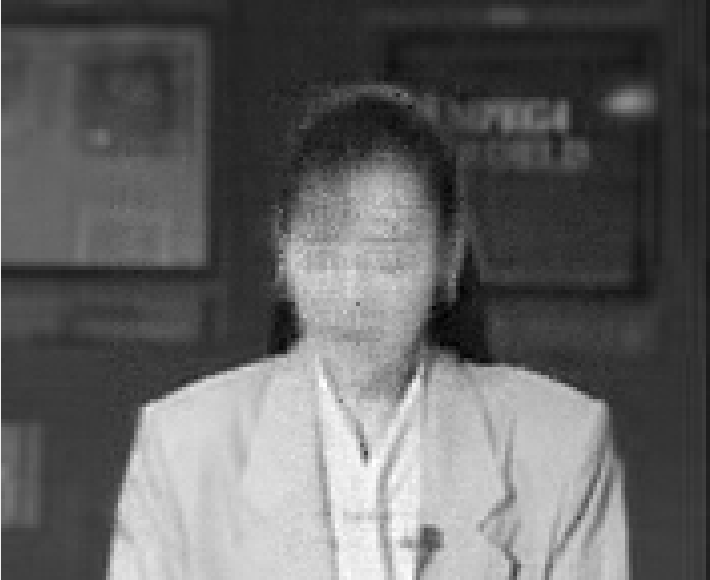}
		\end{minipage}
	}
	\subfigure[120th frame]{
		\begin{minipage}[b]{0.2\textwidth}
			\centerline{\scriptsize } \vspace{1pt}
			\includegraphics[width=1.2in,height=0.76in]{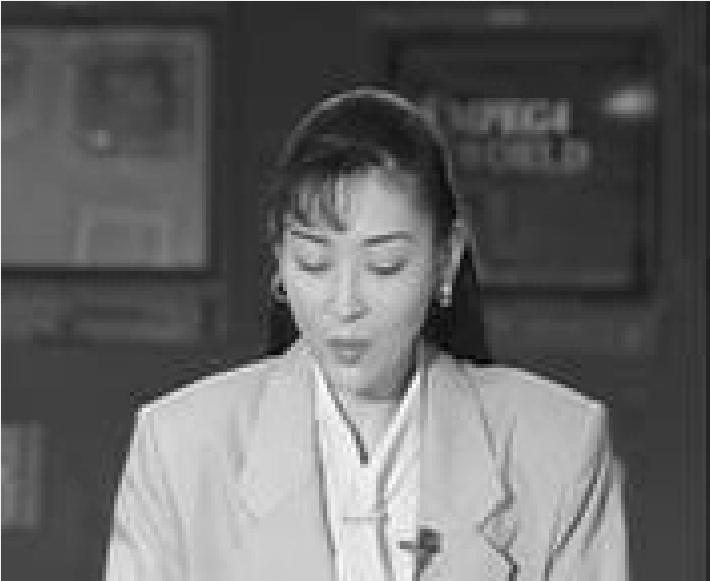}
			\centerline{\scriptsize }\\ \vfill \vspace{-23pt}
			\includegraphics[width=1.2in,height=0.76in]{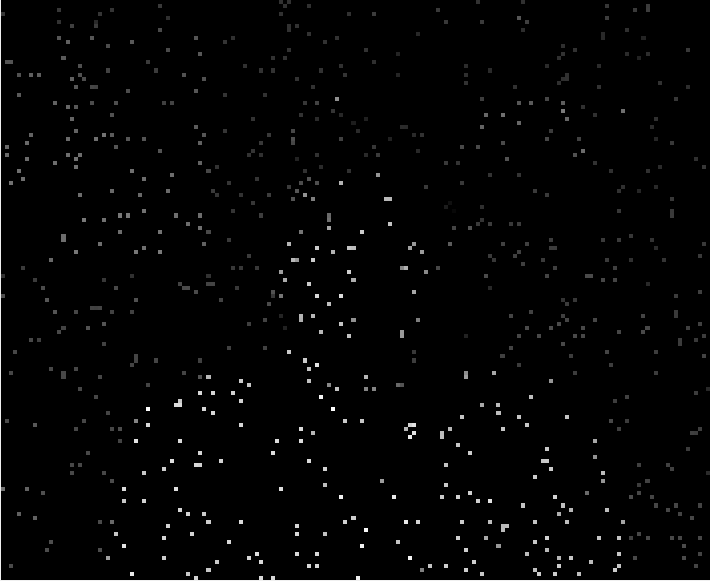}
			\centerline{\scriptsize }\\ \vfill \vspace{-23pt}
			\includegraphics[width=1.2in,height=0.76in]{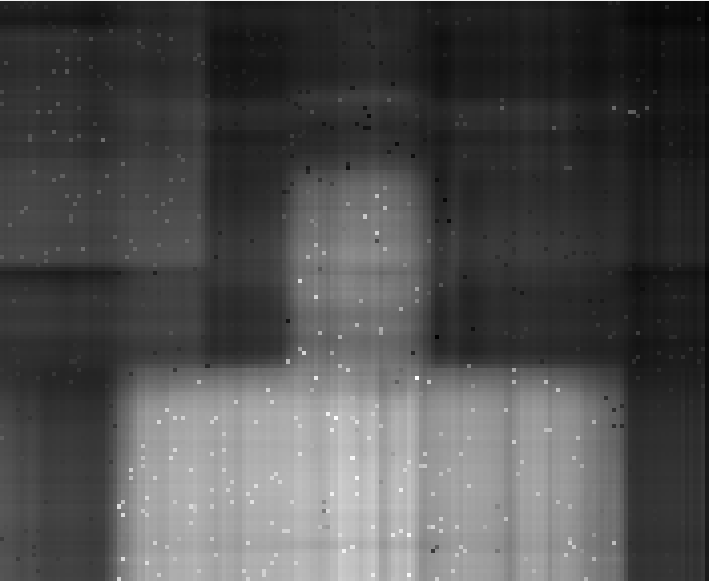}
			\centerline{\scriptsize }\\ \vfill \vspace{-23pt}
			\includegraphics[width=1.2in,height=0.76in]{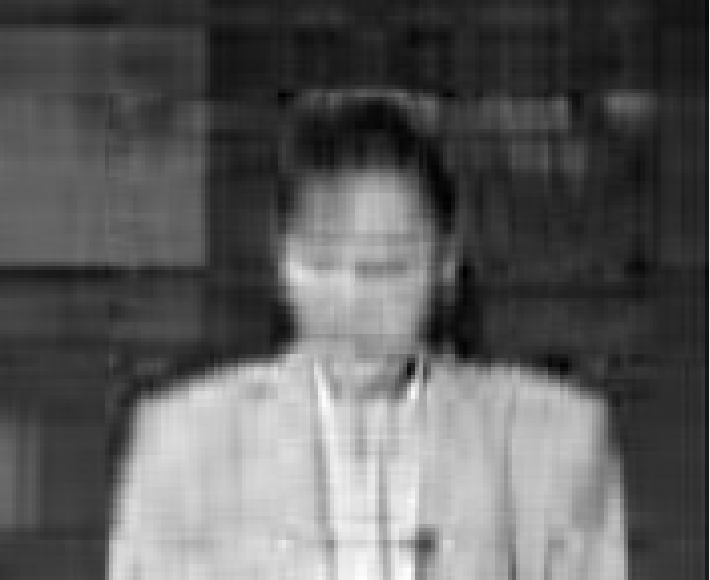}
			\centerline{\scriptsize }\\ \vfill \vspace{-23pt}
			\includegraphics[width=1.2in,height=0.76in]{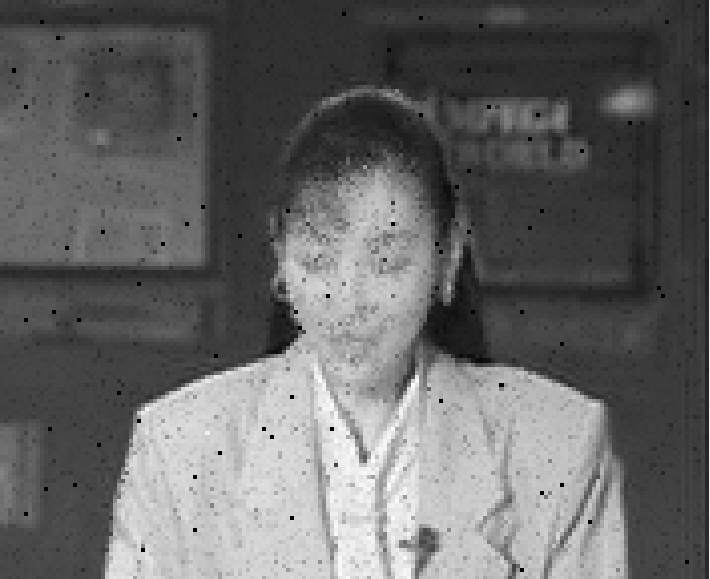}
			\centerline{\scriptsize }\\ \vfill \vspace{-23pt}
			\includegraphics[width=1.2in,height=0.76in]{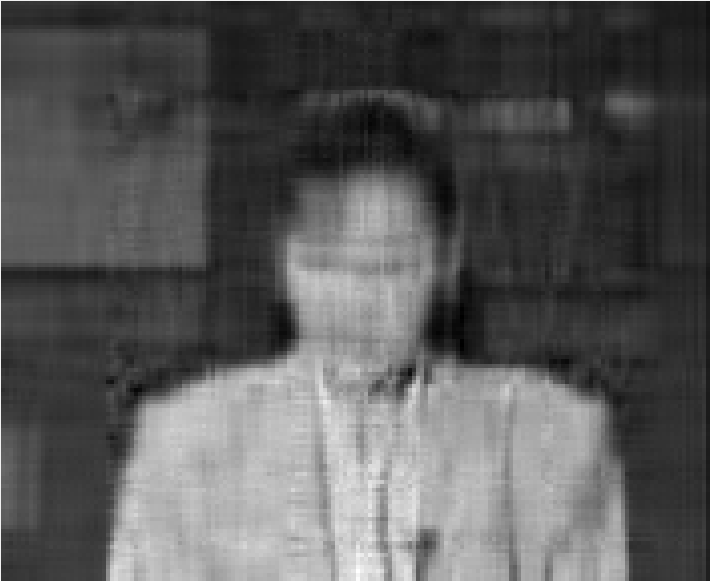}
			\centerline{\scriptsize }\\ \vfill \vspace{-23pt}
			\includegraphics[width=1.2in,height=0.76in]{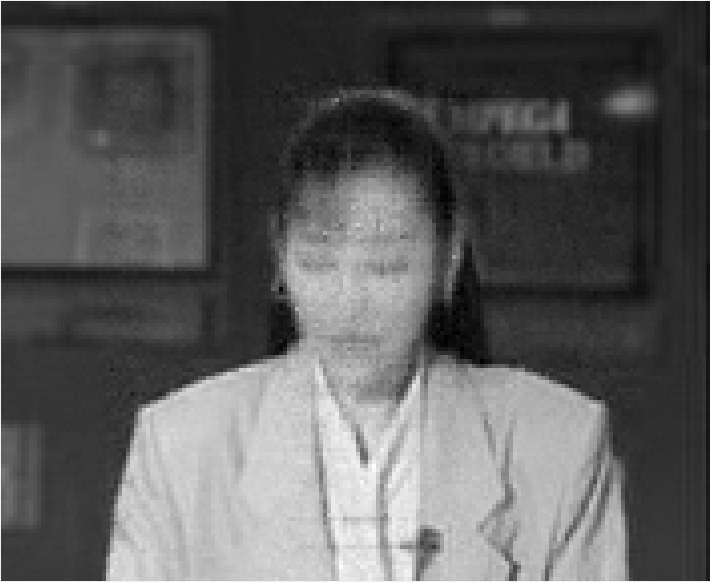}
			\centerline{\scriptsize }\\ \vfill \vspace{-23pt}
			\includegraphics[width=1.2in,height=0.76in]{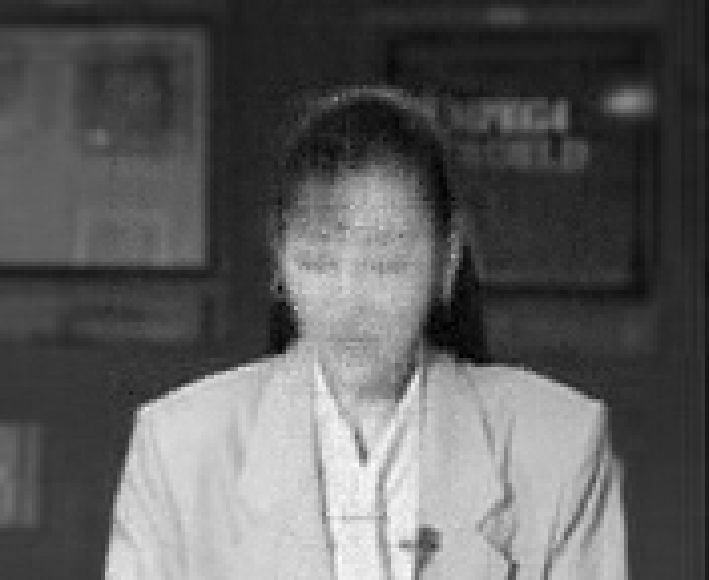}
			\centerline{\scriptsize }\\ \vfill \vspace{-23pt}
			\includegraphics[width=1.2in,height=0.76in]{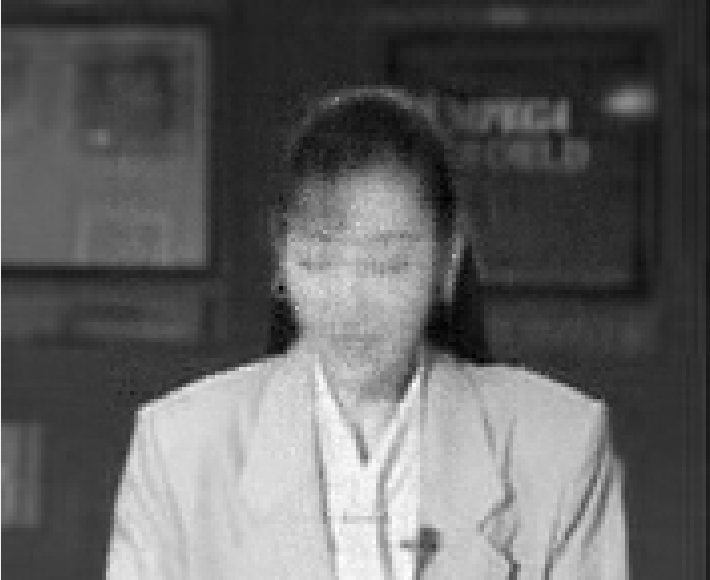}
		\end{minipage}
	}
	\subfigure[180th frame]{
		\begin{minipage}[b]{0.2\textwidth}
			\centerline{\scriptsize } \vspace{1pt}
			\includegraphics[width=1.2in,height=0.76in]{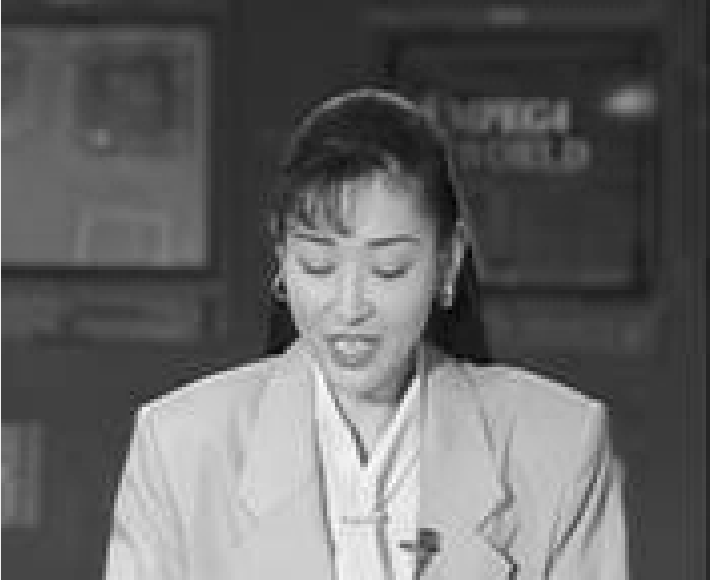}
			\centerline{\scriptsize }\\ \vfill \vspace{-23pt}
			\includegraphics[width=1.2in,height=0.76in]{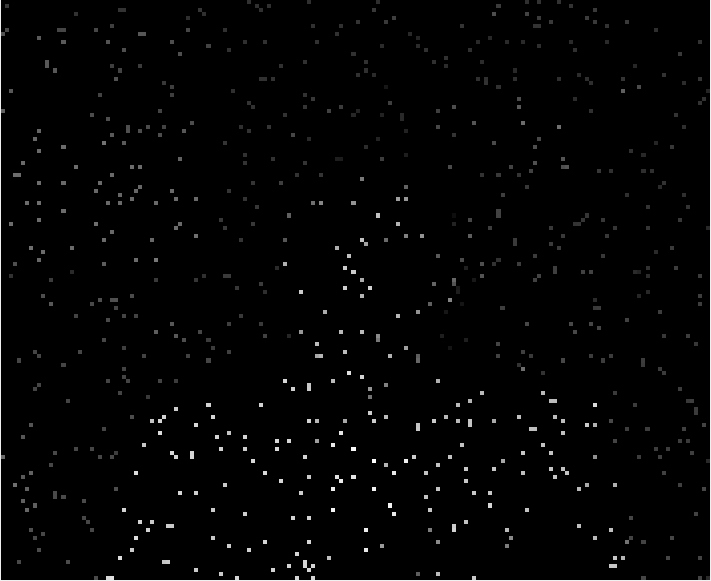}
			\centerline{\scriptsize }\\ \vfill \vspace{-23pt}
			\includegraphics[width=1.2in,height=0.76in]{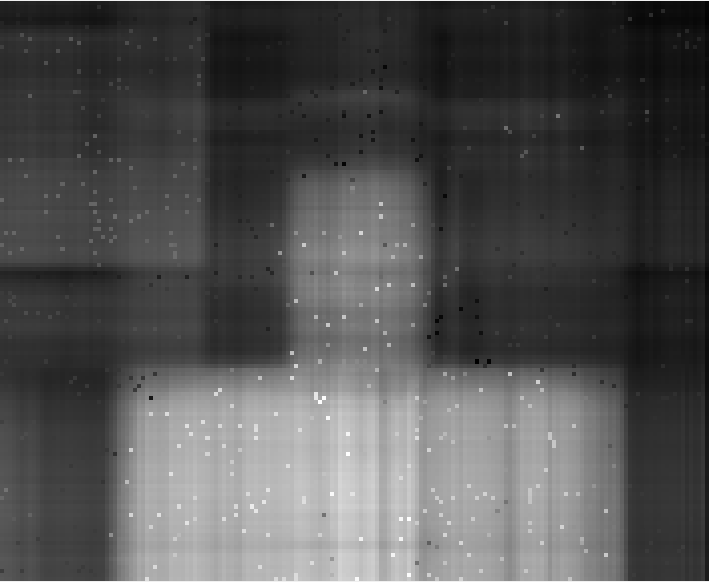}
			\centerline{\scriptsize }\\ \vfill \vspace{-23pt}
			\includegraphics[width=1.2in,height=0.76in]{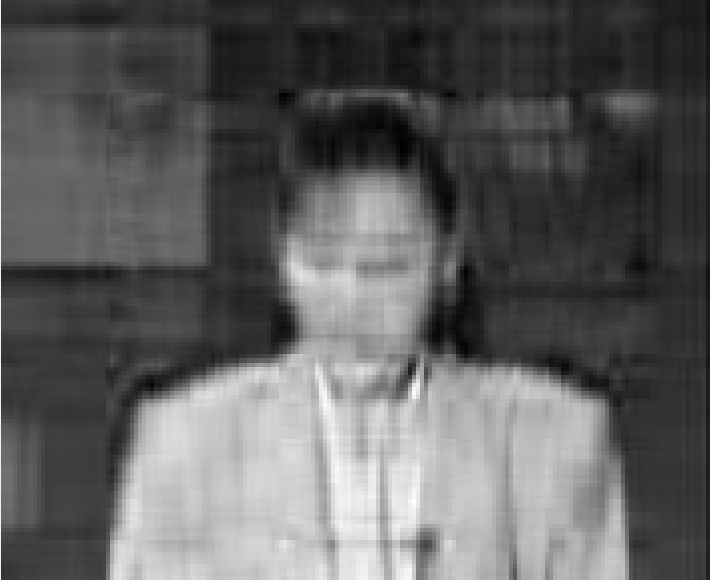}
			\centerline{\scriptsize }\\ \vfill \vspace{-23pt}
			\includegraphics[width=1.2in,height=0.76in]{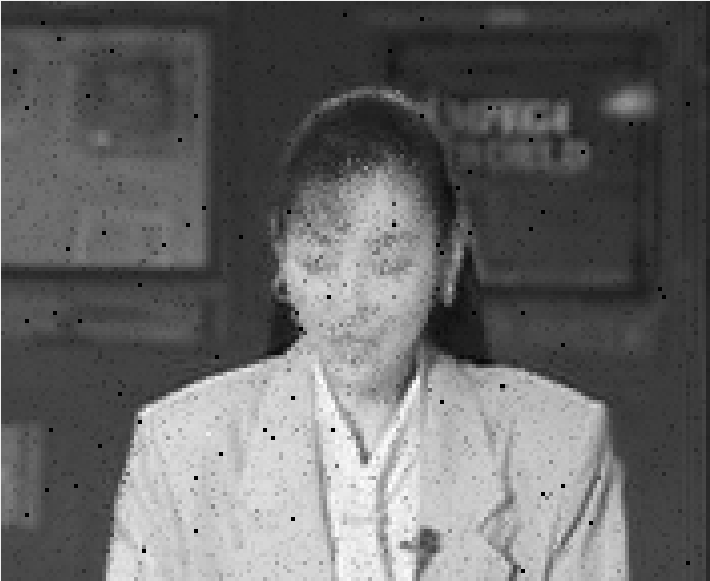}
			\centerline{\scriptsize }\\ \vfill \vspace{-23pt}
			\includegraphics[width=1.2in,height=0.76in]{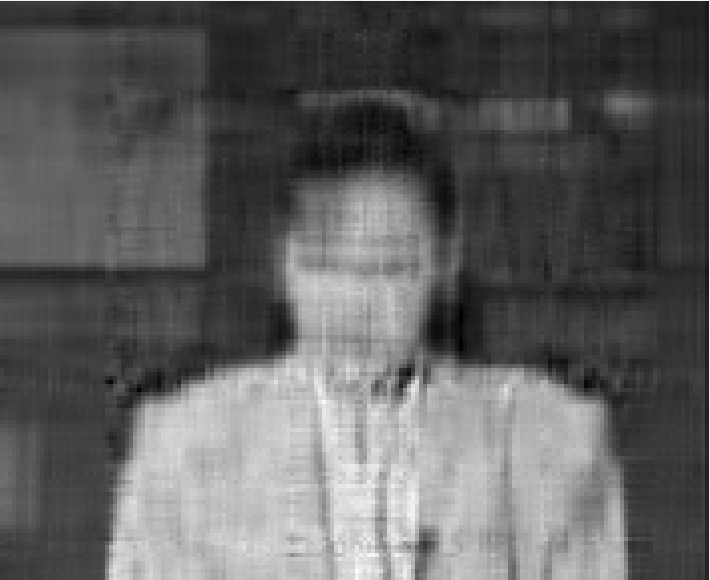}
			\centerline{\scriptsize }\\ \vfill \vspace{-23pt}
			\includegraphics[width=1.2in,height=0.76in]{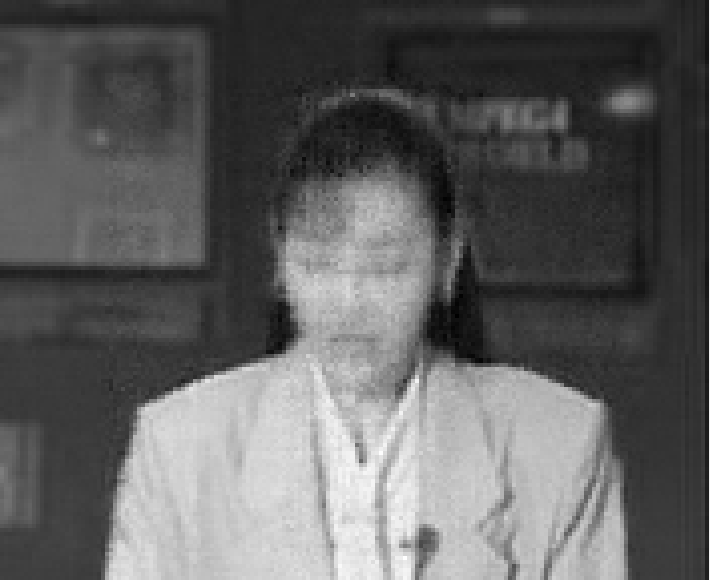}
			\centerline{\scriptsize }\\ \vfill \vspace{-23pt}
			\includegraphics[width=1.2in,height=0.76in]{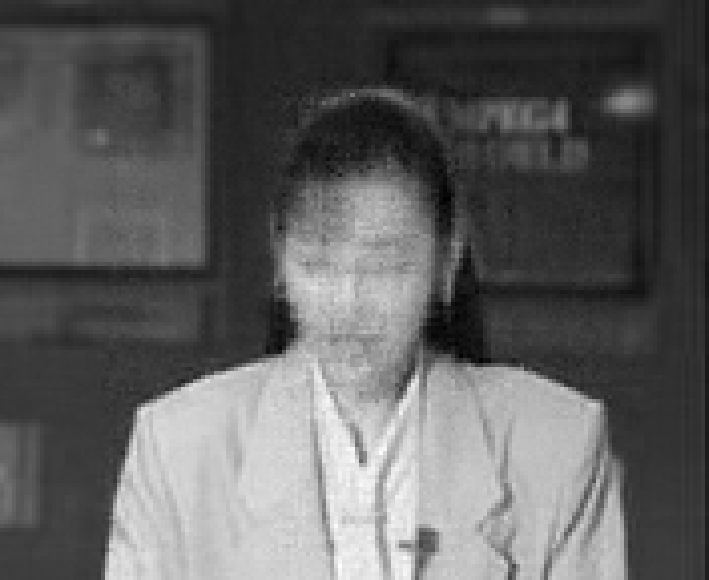}
			\centerline{\scriptsize }\\ \vfill \vspace{-23pt}
			\includegraphics[width=1.2in,height=0.76in]{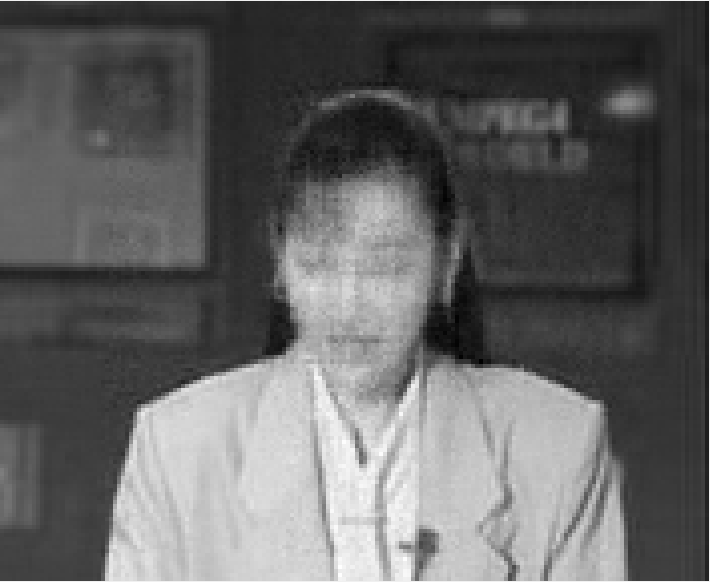}
		\end{minipage}
	}
	\caption{\small Recovery results by different methods for the Announcer data with const$_1=80$.
		First row: Original images.
		Second row: Observed images.
		Third row: Recovered images  by LRTC.
		Fourth row: Recovered images by TF.
		Fifth row: Recovered images by Square Deal.
		Sixth row: Recovered images by GoG.
		Seventh row: Recovered images by t-SVD (FFT).
		Eighth row: Recovered images by t-SVD (DCT).
		Ninth row: Recovered images by t-SVD (data).}\label{VideoVisc}
\end{figure}

We use const$_1 n_{(1)}\log(n_{(1)}n_3)$
number of samples based on Theorem \ref{Theorem1}. The
range const$_1$ is from 20 to 120 with increment
step-size being 10.
For example, when $\varpi=70\%$,
the $\sum_{i=1}^{n_3}r_i(\varpi)$ of Announcer obtained by t-SVD (FFT), t-SVD (DCT), t-SVD (data)
are $1886$, $1649$, $1502$, respectively. When const$_1$ is equal to 100,
the required sample size would be enough for successful recovery by Theorem \ref{Theorem1}.
 When const$_1 >100$, e.g., const$_1= 110,120$, the improvements of PSNR values by the three t-SVD methods are very small.
In Table \ref{tab5}, we show the PSNR, SSIM values, and CPU time (in seconds)
of different methods for the testing video data sets with different const$_1$.
It can be seen that the performance of t-SVD (data) is better
than that of LRTC, TF, Square Deal, GoG, t-SVD (FFT), and t-SVD (DCT) in terms of PSNR and SSIM values.
The PSNR and SSIM values obtained by t-SVD (DCT) are higher than those obtained by t-SVD (FFT).
Hence, the number of samples required by t-SVD (DCT) and  t-SVD (data) is
smaller than that required by LRTC, TF, Square Deal, GoG, and t-SVD (FFT) for the same recovery performance,
which demonstrates the conclusion of Theorem \ref{Theorem1}.
Moreover, the CPU time (in seconds) required by GoG is much more than that required by other methods.

Figure \ref{VideoVisc} shows the visual quality of the 20th, 80th, 120th, 180th frames of the recovered images by LRTC, TF, Square Deal, GoG, and three t-SVD methods, where const$_1 = 80$.
It can be seen that the three t-SVD methods outperform LRTC, TF, Square Deal,
and GoG for different frames in terms of visual quality,
where the recovered images by the t-SVD methods are more clear.

\section{Concluding Remarks}\label{Sect5}

In this paper, we have established the sample size requirement
for exact recovery in the tensor completion problem
by using transformed tensor SVD.
We have shown that for any $\mathcal{Z}\in\mathbb{C}^{n_1\times n_2\times n_3}$
with transformed multi-rank $(r_1,r_2,\ldots, r_{n_3})$,
one can recover the tensor exactly with high probability under some incoherence conditions
if the sample size of observations is of the order
$O(\sum_{i=1}^{n_3}r_i\max \{n_1, n_2 \} \log ( \max \{ n_1, n_2 \} n_3))$
under uniformly sampled entries.
The sample size requirement of our theory for exact recovery is smaller
than that of existing methods for tensor completion.
Moreover, several numerical experiments on both synthetic data and real-world data sets are
presented to show the superior performance of our methods in comparison with other state-of-the-art methods.

In further work, it would be of great interest to extend the transformed tensor SVD and
tensor completion results to higher-order tensors (cf. \cite{martin2013order}).
 It would be also of great interest to extend the result of the unitary transformation
to any invertible linear transformation for tensor completion (cf. \cite{kernfel}).

\section*{Acknowledgments}
The authors are grateful to  Dr. Cun Mu at Walmart Labs
 and Dr. Dong Xia at The Hong Kong University of Science and Technology
for sharing the codes of the Square Deal \cite{mu2014square} and GoG methods \cite{Xia2019On}, respectively.
 The authors are also grateful to the anonymous referees for
 their constructive suggestions and comments to improve the presentation of the paper.

\section*{Appendix A.} \label{Attechment}

We first list the following lemma, which is the main tool to prove our conclusions.
\begin{lemma}(\cite[Theorem 4]{Recht2011})\label{lem5}
Let $X_{1},\ldots,X_{L}\in \R^{n\times n}$ be independent zero mean random matrices of dimension $d_{1}\times d_{2}$. Suppose
$\rho_{k}^2=\max\left\{\|\mathbb{E} [X_{k}X_{k}^{T}]\|,\|\mathbb{E} [X_{k}^{T}X_{k}]\|\right\}$
and $\|X_{k}\|\leq B$ almost surely for all $k=1,\ldots,L.$ Then for any $\tau>0,$
\begin{equation*}
\PP\left[\Biggl\|\sum_{k=1}^{L}X_{k}\Biggl\|>\tau\right]\leq(d_{1}+d_{2})\exp\left(\frac{-\tau^2/2}{\sum_{k=1}^{L}\rho_{k}^2+B\tau/3}\right).
\end{equation*}
Moreover, if
$
\max\left\{\big\|\sum_{k=1}^{L}X_{k}X_{k}^{T}\big\|,\big\|\sum_{k=1}^{L}X_{k}^{T}X_{k}\big\|\right\}\leq \sigma^2,
$
then for any $c>1,$ we have
\begin{align*}
\left\|\sum_{k=1}^{L}X_{k}\right\|\leq \sqrt{4c\sigma^2\log(d_{1}+d_{2})}+cB\log(d_{1}+d_{2})
\end{align*}
holds with probability at least $1-(d_{1}+d_{2})^{-(c-1)}.$
\end{lemma}

\vspace{2mm}
\noindent
\subsection*{Proof of Lemma \ref{le1}}
Let $\mathcal{E}_{ijk}$  be a unit tensor whose $(i,j,k)$-th
entry is 1 and others are 0. Then for an arbitrary tensor $ \mathcal{Z}\in \mathbb{C}^{n_1 \times n_2 \times n_3}$, we have $\mathcal{Z} = \sum_{i,j,k} \<\mathcal{E}_{ijk},
\mathcal{Z}\>\mathcal{E}_{ijk}.$ Recall Definition \ref{defn},  $\mathcal{E}_{ijk}$ can be expressed as $\mathcal{E}_{ijk}=  \tc{e}_{ik}\diamond_{\bf \Phi}
\ddot{\bm{e}}_k \diamond_{\bf \Phi} \tc{e}_{jk}^H,$ then  $\PT(\mathcal{Z})$ can also be decomposed as
\begin{align*}
\PT(\mathcal{Z}) =  \sum_{i,j,k} \<\PT(\mathcal{Z}), \eijk\> \eijk,
\end{align*}
where $T$ is defined as \eqref{e1}. Similarly,
\begin{align}
\rho^{-1}\PT\PO\PT(\mathcal{Z})  = \sum_{i,j,k}
\rho^{-1}\delta_{ijk}\<\mathcal{Z}, \PT(\eijk)\>\PT(\eijk). \nonumber
\end{align}
Define the operator $\mathcal{T}^{ijk}$ as:
\begin{align*}
\mathcal{T}^{ijk}(\mathcal{Z})=\rho^{-1}\delta_{ijk}\<\mathcal{Z}, \PT(\eijk)\>\PT(\eijk).
\end{align*}
Then by the definition of tensor operator norm,  we can get
\begin{align*}
\|\mathcal{T}^{ijk}\|_{\textup{op}} =\frac{1}{\rho}\|\PT(\eijk)\|^2_F, ~\text{and}~\|\PT\|_{\textup{op}} \leq 1.
\end{align*}
Note the fact that for any two positive semidefinite matrices ${A},{B}\in\mathbb{C}^{n\times n}$, we have $\|{A} -
{B}\| \leq \max\{\|{A}\|, \|{B}\|\}.$
Therefore, by the inequality given in Proposition \ref{pro1}, we have
\begin{equation}
\Big\|\mathcal{T}^{ijk} -
\frac{1}{n_1n_2n_3}\PT\Big\|_{\textup{op}} \leq
\max\left\{\frac{1}{\rho}\|\PT(\eijk)\|^2_F,
\frac{1}{n_1n_2n_3}\right\} \leq  \frac{2\mu \sum{r_{i}}}{n_{(2)}n_{3}\rho}.
\nonumber
\end{equation}
In addition, one has
\begin{align}
&~\Big\|\E\Big(\mathcal{T}^{ijk}- \frac{1}{n_1 n_2 n_3}\PT\Big)^2\Big\|_\textup{op}\nonumber\\
\leq &~\left\|\E\left(\frac{1}{\rho}\|\PT(\eijk)\|^2_F\mathcal{T}^{ijk}\right) - \frac{2}{n_1 n_2 n_3}\PT\E(\mathcal{T}^{ijk}) + \frac{1}{n^2_1n^2_2n^2_3}\PT\right\| \nonumber\\
=& ~ \left\|\frac{1}{\rho}\|\PT(\eijk)\|^2_F\frac{1}{n_1n_2n_3}\PT - \frac{1}{n^2_1 n^2_2 n^2_3}\PT\right\|
\leq \frac{2\mu \sum{r_{i}}}{n_{(1)}n^2_{(2)}n^2_3\rho}. \nonumber
\end{align}
Setting $\tau = \sqrt{\frac{14 \mu \beta \sum{r_{i}}
\log(n_{(1)}n_3)}{3n_{(2)}n_{3}\rho}} \leq
\frac{1}{2}$ with any $\beta>1$ and using Lemma~\ref{lem5}, we have
\begin{align}
&\PP[\|\rho^{-1}\PT\PO\PT - \PT\|_\textup{op} > \tau]
 = \PP\left[\left\|\sum_{i,j,k} \left(\mathcal{T}^{ijk} - \frac{1}{n_1n_2n_3}\PT\right)\right\|_{\textup{op}}> \tau \right] \nonumber\\
\leq~ & 2n_{(1)}n_3\exp\Bigg(\frac{\frac{-7 \mu \beta\sum{r_{i}}
\log(n_{(1)}n_3)}{3n_{(2)}n_{3}\rho}}{\frac{2\mu \sum{r_{i}}}{n_{(2)}n_{3}\rho} + \frac{2\mu \sum{r_{i}}}{6n_{(2)}n_{3}\rho}}\Bigg)
=  2n_{(1)}n_3\exp({-\beta \log n_{(1)}n_{3}})= 2(n_{(1)}n_3)^{1-\beta},
\nonumber
\end{align}
which implies that
\begin{align}
\PP\left[\|\rho^{-1}\PT\PO\PT - \PT\|_\textup{op} \leq \epsilon\right] \geq  1 - 2(n_{(1)}
n_3)^{1-\beta}.\nonumber
\end{align}
This completes the proof.

\vspace{2mm}
\noindent
\section*{Appendix B. Proof of Lemma \ref{lemma2}}
 Denote
$$
\rho^{-1}\mathcal{P}_{\Omega}(\mathcal{Z})-\mathcal{Z}= \sum_{i,j,k}\mathcal{G}^{ijk}=\sum_{i,j,k}\Big(\frac{1}{\rho}\delta_{ijk}-1\Big)\mathcal{Z}_{ijk}\eijk.
$$
Then by the independence of $\delta_{ijk}$, we have $\E[\mathcal{G}^{ijk}]=\bf{0}$ and $\|\mathcal{G}^{ijk}\|\leq \frac{1}{\rho}\|\mathcal{Z}\|_{\infty}.$
 Moreover,
\begin{align*}
\left\|\E\left[\sum_{i,j,k}(\mathcal{G}^{ijk})^H\diamond_{\bf \Phi}\mathcal{G}^{ijk}\right]\right\|
& =\left\|\sum_{i,j,k}|\mathcal{Z}_{ijk}|^2\tc{e}_{jk}\diamond_{\bf \Phi}\tc{e}_{jk}^H\E\left(\frac{1}{\rho}\delta_{ijk}-1\right)^2\right\| \\
& =\left\|\frac{1-\rho}{\rho}\sum_{i,j,k}|\mathcal{Z}_{ijk}|^2\tc{e}_{ik}\diamond_{\bf \Phi}\tc{e}_{ik}^H\right\|.
\end{align*}
Recall the definition of tensor basis, we can get that
${\bf \Phi}[\tc{e}_{jk}\diamond_{\bf \Phi}\tc{e}_{jk}^H]$ is a tensor except the $(j,j,t)$-th tube entries equaling to $({\bf \Phi}[\tub{e}_{k}])_{t}^2=\alpha^2_{t}, t=1,\ldots,n_3 $ with $\sum_{t=1}^{n_{3}}\alpha_{t}^2=1,$
and 0 otherwise. Hence, we get that
\begin{align*}
\left\|\E\left[\sum_{i,j,k}(\mathcal{G}^{ijk})^H\diamond_{\bf \Phi}\mathcal{G}^{ijk}\right]\right\|
&=\frac{1-\rho}{\rho}\left\|\sum_{i,j,k}|\mathcal{Z}_{ijk}|^2\tc{e}_{jk}\diamond_{\bf \Phi}\tc{e}_{jk}^H\right\| \\
& =\frac{1-\rho}{\rho}\max_{j}\left\|\sum_{i,j,k}|\mathcal{Z}_{ijk}|^2\tc{e}_{jk}\diamond_{\bf \Phi}\tc{e}_{jk}^H\right\|\\
&=\frac{1-\rho}{\rho}\max_{j}\left\|\sum_{i,k}|\mathcal{Z}_{ijk}|^2(\overline{\tc{e}}_{jk})_{\bf \Phi}\cdot(\overline{\tc{e}}_{jk})_{\bf \Phi}^H\right\|
\leq \frac{1}{\rho}\|\mathcal{Z}\|^2_{\infty,w}.
\end{align*}
Moreover, $\left\|\E\left[\sum_{i,j,k}\mathcal{G}^{ijk}\diamond_{\bf \Phi}(\mathcal{G}^{ijk})^H\right]\right\|$ can be also bounded similarly.
Then by Lemma \ref{lem5}, we can get that
\begin{align*}
\|\rho^{-1}\mathcal{P}_{\Omega}(\mathcal{Z})-\mathcal{Z}\|_\textup{op}\leq
c\left(\frac{\log(n_{(1)}n_{3})}{\rho}\|\mathcal{Z}\|_{\infty}+\sqrt{\frac{\log (n_{(1)}n_{3})}{\rho}}\|\mathcal{Z}\|_{\infty,w}\right)
\end{align*}
holds with high probability provided that
$m\geq C_0 \epsilon^{-2} \mu \sum{r_{i}}n_{(1)}\log(n_{(1)}n_3).$

\vspace{2mm}
\noindent
\section*{Appendix C. Proof of Lemma \ref{le3}}
 Denote the weighted $b$-th  lateral slice of $(\rho^{-1}\mathcal{P}_{T}\mathcal{P}_{\Omega}-\mathcal{P}_{T})\mathcal{Z}$ as
\begin{align*}
\sum_{i,j,k}\mathcal{F}^{ijk}:&=(\rho^{-1}\mathcal{P}_{T}\mathcal{P}_{\Omega}-\mathcal{P}_{T})\mathcal{Z}\diamond_{\bf \Phi}\tc{e}_{bk} \\
&=\sum_{i,j,k}\Big(\frac{1}{\rho}\delta_{ijk}-1\Big)\mathcal{Z}_{ijk}\mathcal{P}_{T}(\eijk)\diamond_{\bf \Phi}\tc{e}_{bk},
\end{align*}
where $\mathcal{F}^{ijk}\in \C^{n_1 \times 1 \times n_3}$  are zero-mean independent lateral slices. By the incoherence conditions given in Proposition \ref{pro1}, we have
\begin{align*}
\|\mathcal{F}^{ijk}\|_{{F}}=\left\|\left(\frac{1}{\rho}\delta_{ijk}-1\right)\mathcal{Z}_{ijk}\mathcal{P}_{T}(\eijk)\diamond_{\bf \Phi}\tc{e}_{bk}\right\|_{{F}}
\leq \frac{1}{\rho}\sqrt{\frac{2\mu \sum_{i=1}^{n_{3}}r_{i}}{n_{1}n_3}}\|\mathcal{Z}\|_{\infty}.
\end{align*}
Furthermore,
\begin{align*}
\left\|\E\left[\sum_{i,j,k}(\mathcal{F}^{ijk})^H\diamond_{\bf\Phi}\mathcal{F}^{ijk}\right]\right\|_{{F}}
=\frac{1-\rho}{\rho}\sum_{i,j,k}|\mathcal{Z}_{ijk}|^2\|\mathcal{P}_{T}(\eijk)\diamond_{\bf\Phi}\tc{e}_{bk}\|^2_{{F}}.
\end{align*}
Then by the definition of $\mathcal{P}_{T}$, we can get
\begin{align*}
&~\|\mathcal{P}_{T}(\eijk)\diamond_{\bf \Phi}\tc{e}_{bk}\|^2_{F}\\
=&~\|\mathcal{U}\diamond_{\bf \Phi}\mathcal{U}^H\diamond_{\bf\Phi}\eijk\diamond_{\bf \Phi}\tc{e}_{bk} \\
&~~~~+(\mathcal{I}_{\bf \Phi}-\mathcal{U}\diamond_{\bf\Phi}\mathcal{U}^H)\diamond_{\bf\Phi}\eijk\diamond_{\bf\Phi}\mathcal{V}\diamond_{\bf\Phi}
\mathcal{V}^H\diamond_{\bf \Phi}\tc{e}_{bk}\|^2_{{F}}\\
\leq&~\frac{\mu \sum_{i=1}^{n_{3}}r_{i}}{n_{1}n_3}\|\ddot{\bm{e}}_k\diamond_{\bf \Phi}\tc{e}_{jk}\diamond_{\bf \Phi}\tc{e}_{bk}\|^2_{{F}}
+\|\tc{e}_{jk}^H\diamond_{\bf\Phi}\mathcal{V}\diamond_{\bf\Phi}
\mathcal{V}^H\diamond_{\bf \Phi}\tc{e}_{bk}\|^2_{{F}}.
\end{align*}
Therefore, we obtain
\begin{align*}
~& \left\|\E\left[\sum_{i,j,k}(\mathcal{F}^{ijk})^H\diamond_{\bf\Phi}\mathcal{F}^{ijk}\right]\right\|_{F} \\
\leq ~&
\frac{1}{\rho}\sum_{i,j,k}|\mathcal{Z}_{ijk}|^2\|\mathcal{P}_{T}(\eijk)\diamond_{\bf\Phi}\tc{e}_{bk}\|^2_{{F}}\\
\leq ~& \frac{1}{\rho}\sum_{i,j,k}|\mathcal{Z}_{ijk}|^2\frac{\mu \sum_{i=1}^{n_{3}}r_{i}}{n_{1}n_3}\|\ddot{\bm{e}}_k\diamond_{\bf \Phi}\tc{e}_{jk}\diamond_{\bf \Phi}\tc{e}_{bk}\|^2_{{F}}
+\frac{1}{\rho}\sum_{i,j,k}|\mathcal{Z}_{ijk}|^2\|\tc{e}_{jk}^H\diamond_{\bf\Phi}\mathcal{V}\diamond_{\bf\Phi}
\mathcal{V}^H\diamond_{\bf \Phi}\tc{e}_{bk}\|^2_{{F}}\\
\leq ~& \frac{\mu \sum_{i=1}^{n_{3}}r_{i}}{\rho n_{1}n_3}\|\mathcal{Z}\|^2_{\infty,w}+\frac{1}{\rho}\sum_{i,k}|\mathcal{Z}_{ijk}|^2\|\tc{e}_{jk}^H\diamond_{\bf\Phi}\mathcal{V}\diamond_{\bf\Phi}
\mathcal{V}^H\diamond_{\bf \Phi}\tc{e}_{jk}\|^2_{{F}}\\
\leq ~& \frac{2\mu \sum_{i=1}^{n_{3}}r_{i}}{\rho n_{(2)}n_3}\|\mathcal{Z}\|_{\infty,w}^2,
\end{align*}
where the third inequality can be derived by $\tc{e}_{jk}^H\diamond_{\bf\Phi}\tc{e}_{bk}=\bf{0}$ if $j\neq b.$ By the same argument, $\left\|\E[\sum_{ijk}\mathcal{F}^{ijk}\diamond_{\bf\Phi}(\mathcal{F}^{ijk})^H]\right\|_{F}$ can be bounded by the same quantity. Therefore, by Lemma \ref{lem5}, we  get that
\begin{align*}
\|(\rho^{-1}\mathcal{P}_{T}\mathcal{P}_{\Omega}-\mathcal{P}_{T})\mathcal{Z}\diamond_{\bf \Phi}\tc{e}_{bk}\|_{{F}}\leq \frac{1}{2}\|\mathcal{Z}\|_{\infty,w}+\frac{1}{2}\sqrt{\frac{n_{(1)}n_{3}}{\mu \sum_{i=1}^{n_{3}}r_{i}}}\|\mathcal{Z}\|_{\infty}
\end{align*}
holds with high probability.
We can also get the same results with respect to $\tc{e}^H_{ak}\diamond_{\bf \Phi}(\rho^{-1}\mathcal{P}_{T}\mathcal{P}_{\Omega}-\mathcal{P}_{T})\mathcal{Z}.$
 Then Lemma  \ref{le3} follows from using a union bound over
all the tensor columns and rows, and the desired results hold with high probability.

\bibliographystyle{abbrv}

\bibliography{ref}

\end{document}